\documentclass[acmtog,nonacm]{acmart}

\usepackage{booktabs} 
\usepackage[caption=false,font=normalsize,labelfont=sf,textfont=sf]{subfig}
\usepackage{ulem}
\usepackage{color}
\usepackage[ruled]{algorithm2e} 

\SetAlFnt{\small}
\SetAlCapFnt{\small}
\SetAlCapNameFnt{\small}
\SetAlCapHSkip{0pt}

\begin{document}
\title{NeuralRoom: Geometry-Constrained Neural Implicit Surfaces for Indoor Scene Reconstruction}

\author{Yusen Wang}
\affiliation{%
 \institution{School of Computer Science, Wuhan University}
 \city{Wuhan}
 \country{China}}
\email{wangyusen@whu.edu.cn}
\author{Zongcheng Li}
\affiliation{%
 \institution{School of Computer Science, Wuhan University}
 \city{Wuhan}
 \country{China}}
\author{Yu Jiang}
\affiliation{%
 \institution{School of Computer Science, Wuhan University}
 \city{Wuhan}
 \country{China}}
\author{Kaixuan Zhou}
\affiliation{%
 \institution{Riemann Lab, Huawei}
 \city{Wuhan}
 \country{China}}
\author{Tuo Cao}
\affiliation{%
 \institution{School of Computer Science, Wuhan University}
 \city{Wuhan}
 \country{China}}
\author{Yanping Fu}
\affiliation{%
 \institution{School of Computer Science and Technology, Anhui University}
 \city{Hefei}
 \country{China}}
\author{Chunxia Xiao}
\authornote{Corresponding author}
\affiliation{
 \institution{School of Computer Science, Wuhan University}
 \city{Wuhan}
 \country{China}}



\begin{abstract}
We present a novel neural surface reconstruction method called NeuralRoom for reconstructing room-sized indoor scenes directly from a set of 2D images. Recently, implicit neural representations have become a promising way to reconstruct surfaces from multiview images due to their high-quality results and simplicity. However, implicit neural representations usually cannot reconstruct indoor scenes well because they suffer severe shape-radiance ambiguity. We assume that the indoor scene consists of texture-rich and flat texture-less regions. In texture-rich regions, the multiview stereo can obtain accurate results. In the flat area, normal estimation networks usually obtain a good normal estimation. Based on the above observations, we reduce the possible spatial variation range of implicit neural surfaces by reliable geometric priors to alleviate shape-radiance ambiguity. Specifically, we use multiview stereo results to limit the NeuralRoom optimization space and then use reliable geometric priors to guide NeuralRoom training. Then the NeuralRoom would produce a neural scene representation that can render an image consistent with the input training images. In addition, we propose a smoothing method called perturbation-residual restrictions to improve the accuracy and completeness of the flat region, which assumes that the sampling points in a local surface should have the same normal and similar distance to the observation center. Experiments on the ScanNet dataset show that our method can reconstruct the texture-less area of indoor scenes while maintaining the accuracy of detail. We also apply NeuralRoom to more advanced multiview reconstruction algorithms and significantly improve their reconstruction quality.

\end{abstract}

%
%

\ccsdesc[500]{Computing methodologies~Reconstruction}

%
%

\keywords{Neural Implicit Representation, Indoor Scene Reconstruction, Multiview Reconstruction}

\begin{teaserfigure}
\includegraphics[width=\textwidth]{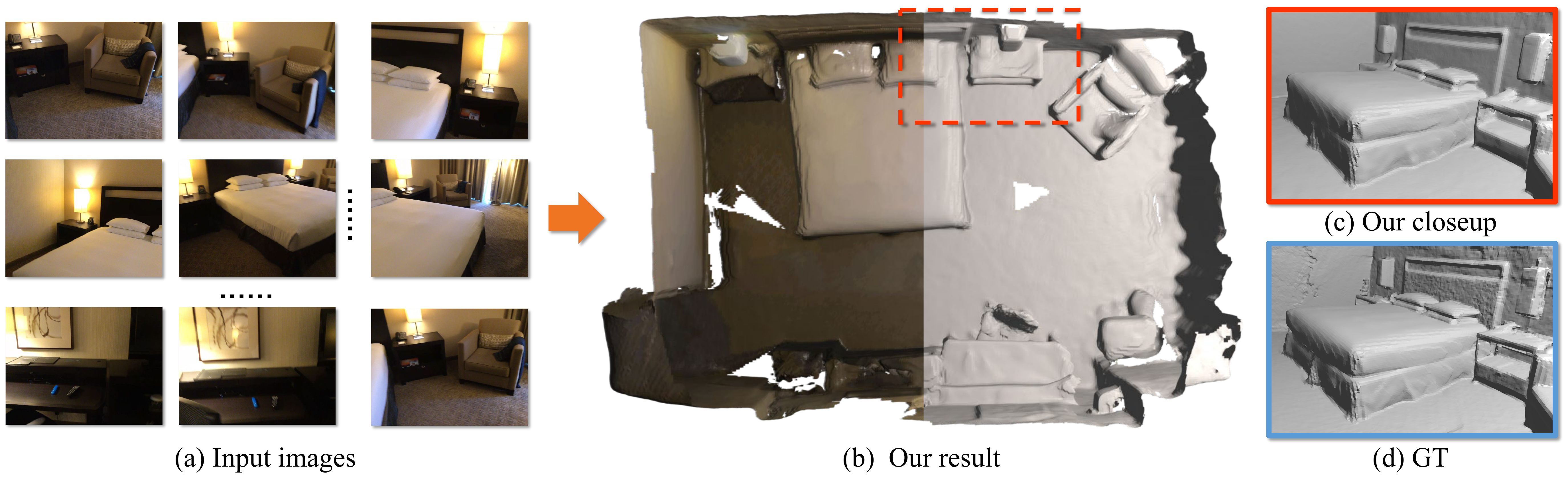}
\caption{We present a system called NeuralRoom for reconstructing a room-sized indoor scene from 2D images. There are many texture-less regions in indoor scenes, making conventional multiview stereo methods fail in reconstruction. The implicit neural representation method has recently become a promising reconstruction method due to its simplicity and high reconstruction quality. However, shape-radiance ambiguity makes it unable to reconstruct indoor scenes well. NeuralRoom effectively integrates normal and depth information to overcome ambiguity, which guarantees reconstruction details and completeness.}
\label{fig:teaser}
\end{teaserfigure}

\maketitle

\begin{figure*}[th]
  \centering 
  \newcommand{\myvspace}{1.0 pt} 
  \newcommand{\widthOfFullPage}{0.1895} 
  \newcommand{\widthOfMiniPage}{1}
  \subfloat[NeuS]{
    \begin{minipage}[b]{\widthOfFullPage\linewidth}
      \centering
            \includegraphics[width=\widthOfMiniPage\linewidth]{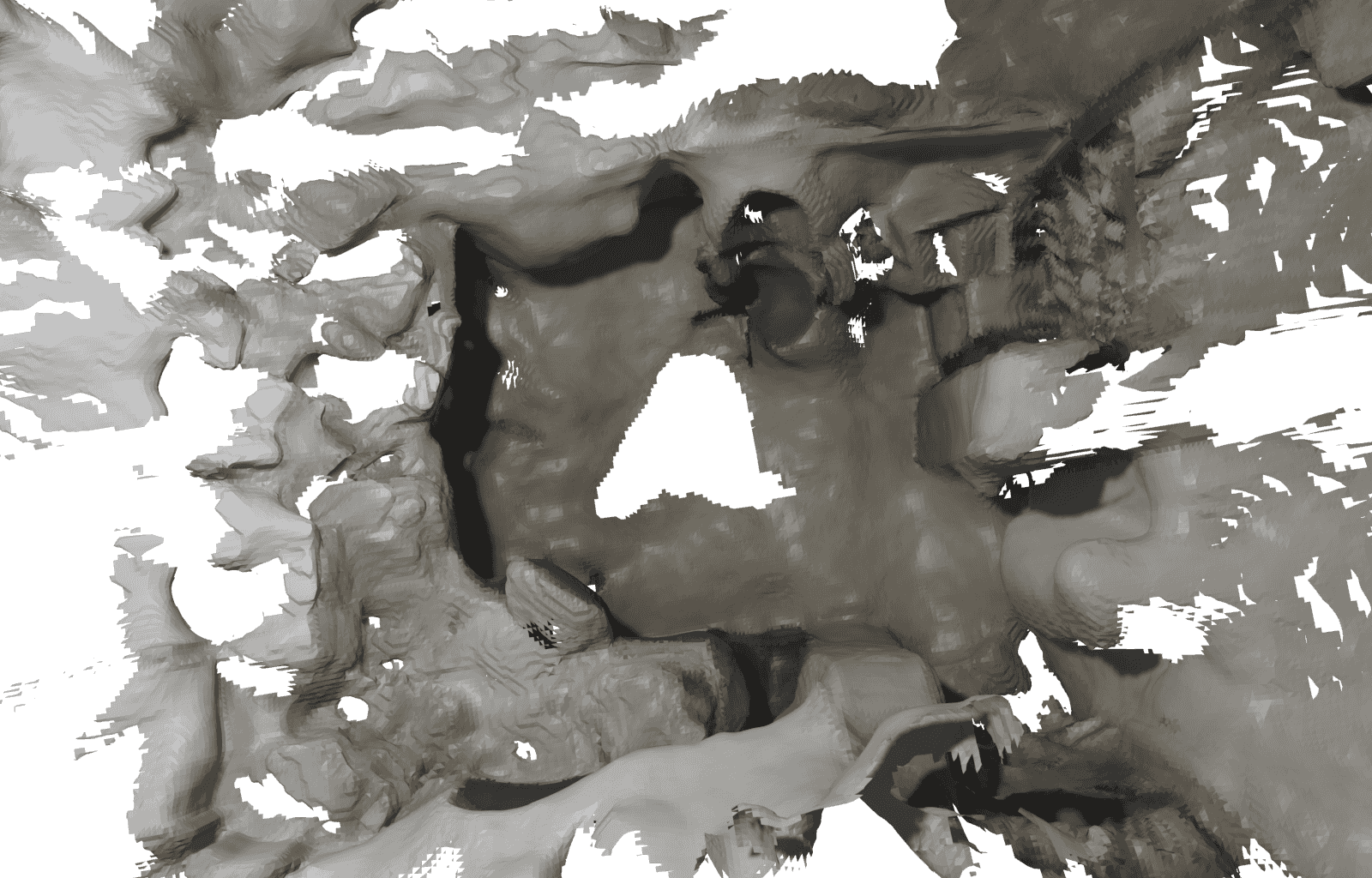}\vspace{\myvspace}
                        \includegraphics[width=\widthOfMiniPage\linewidth]{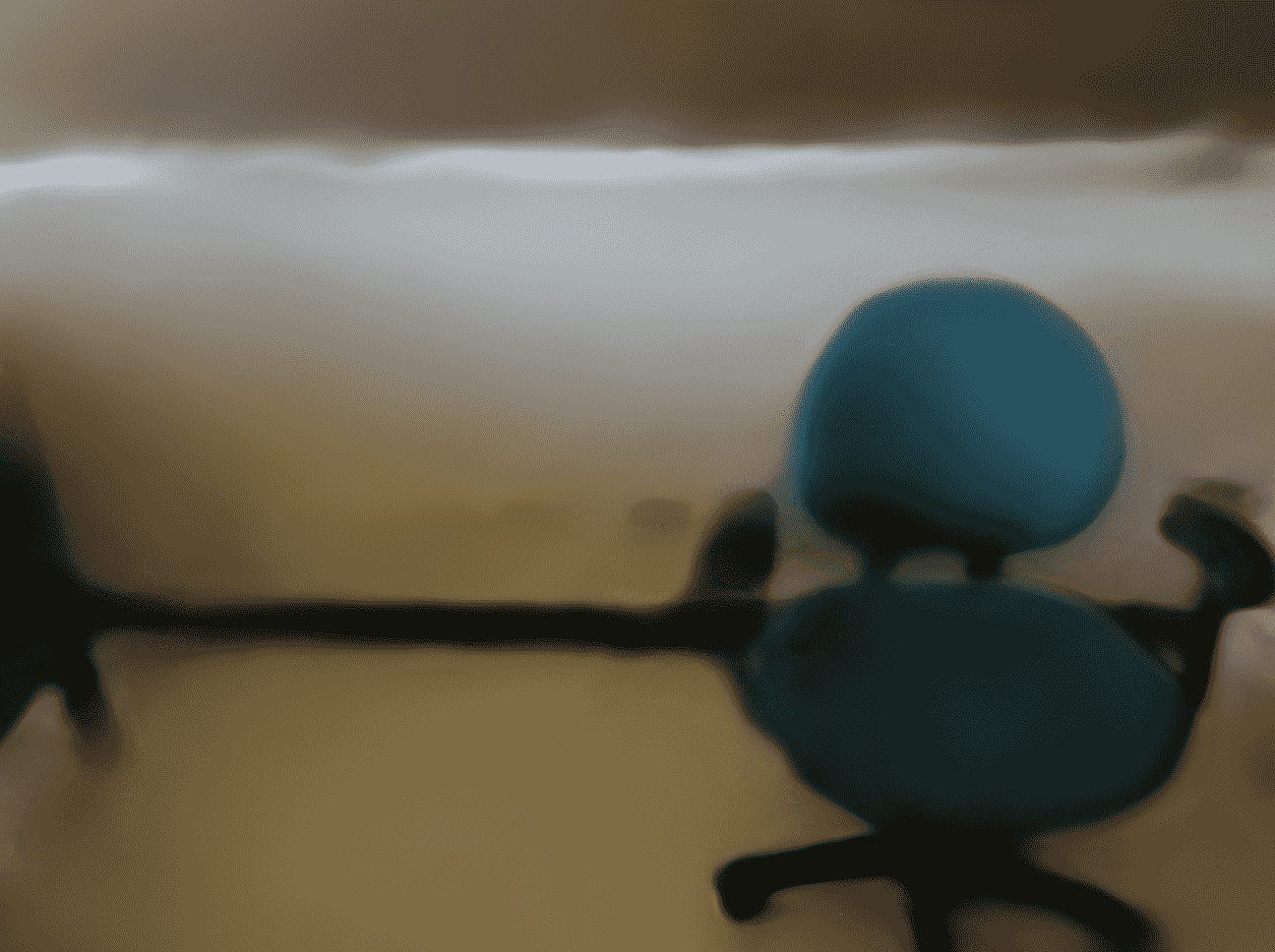}\vspace{\myvspace}
      \includegraphics[width=\widthOfMiniPage\linewidth]{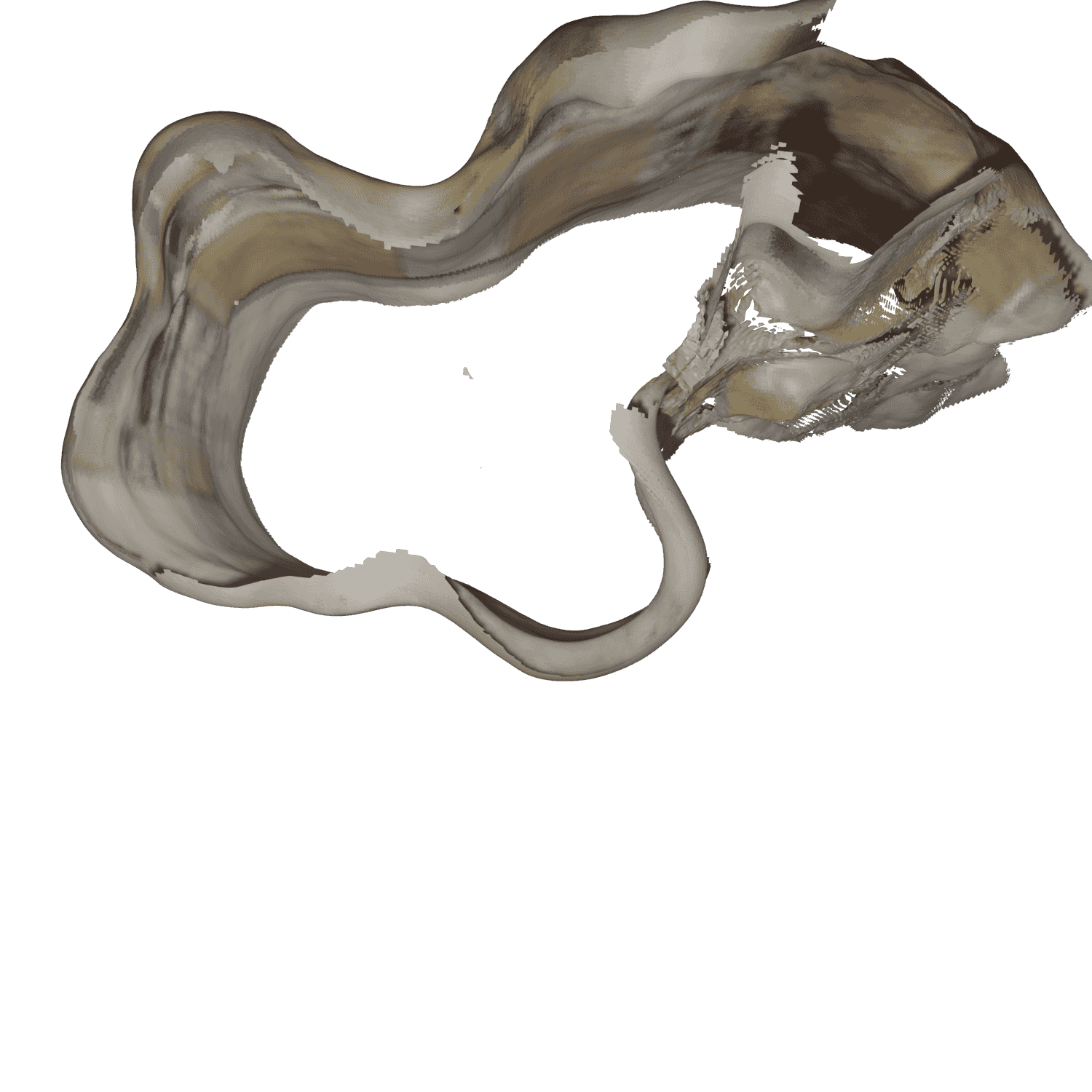}\vspace{\myvspace}
                  \includegraphics[width=\widthOfMiniPage\linewidth]{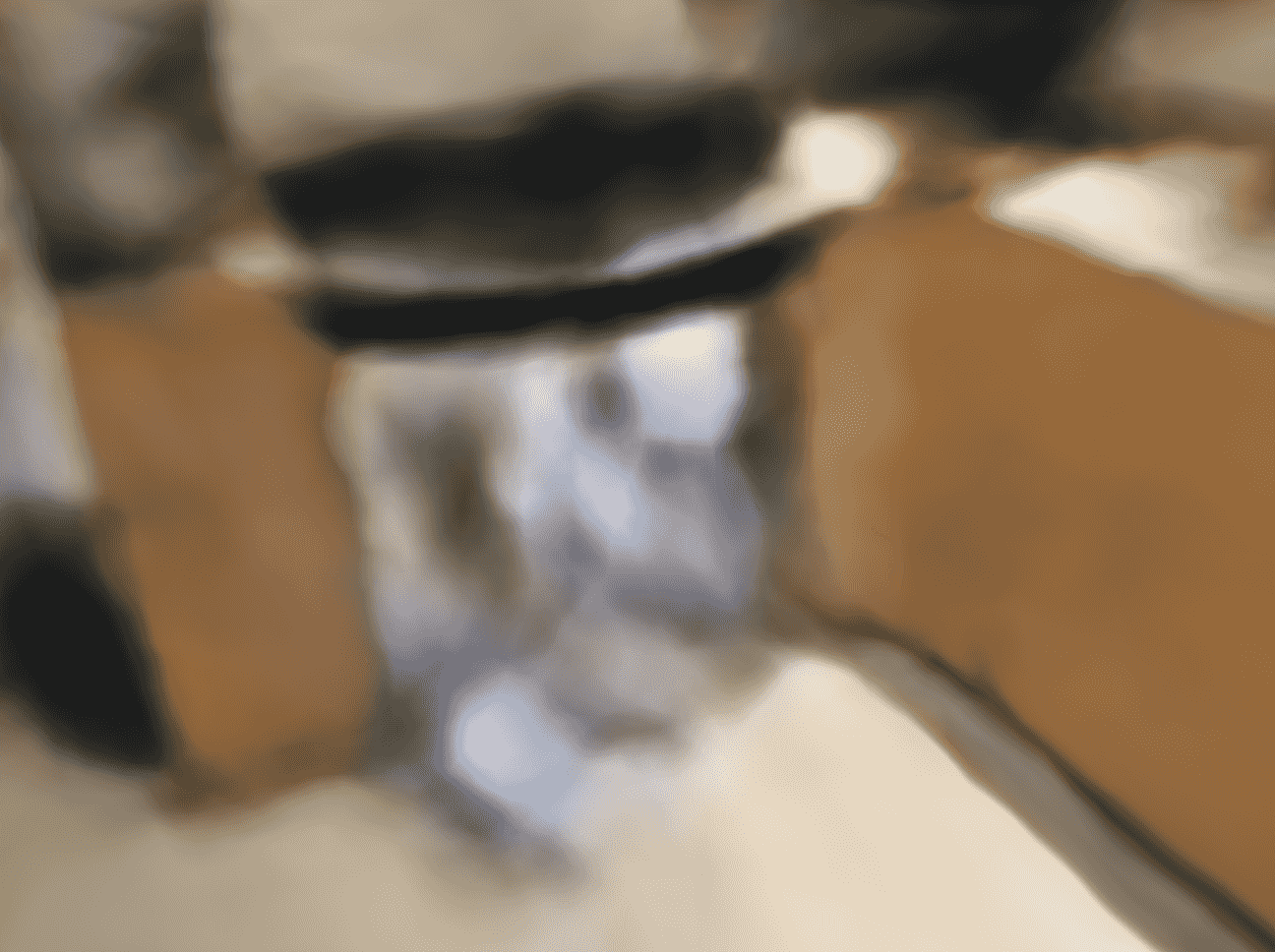}\vspace{\myvspace}

    \end{minipage}
  }
  \subfloat[Unisurf]{
    \begin{minipage}[b]{\widthOfFullPage\linewidth}
      \centering
            \includegraphics[width=\widthOfMiniPage\linewidth]{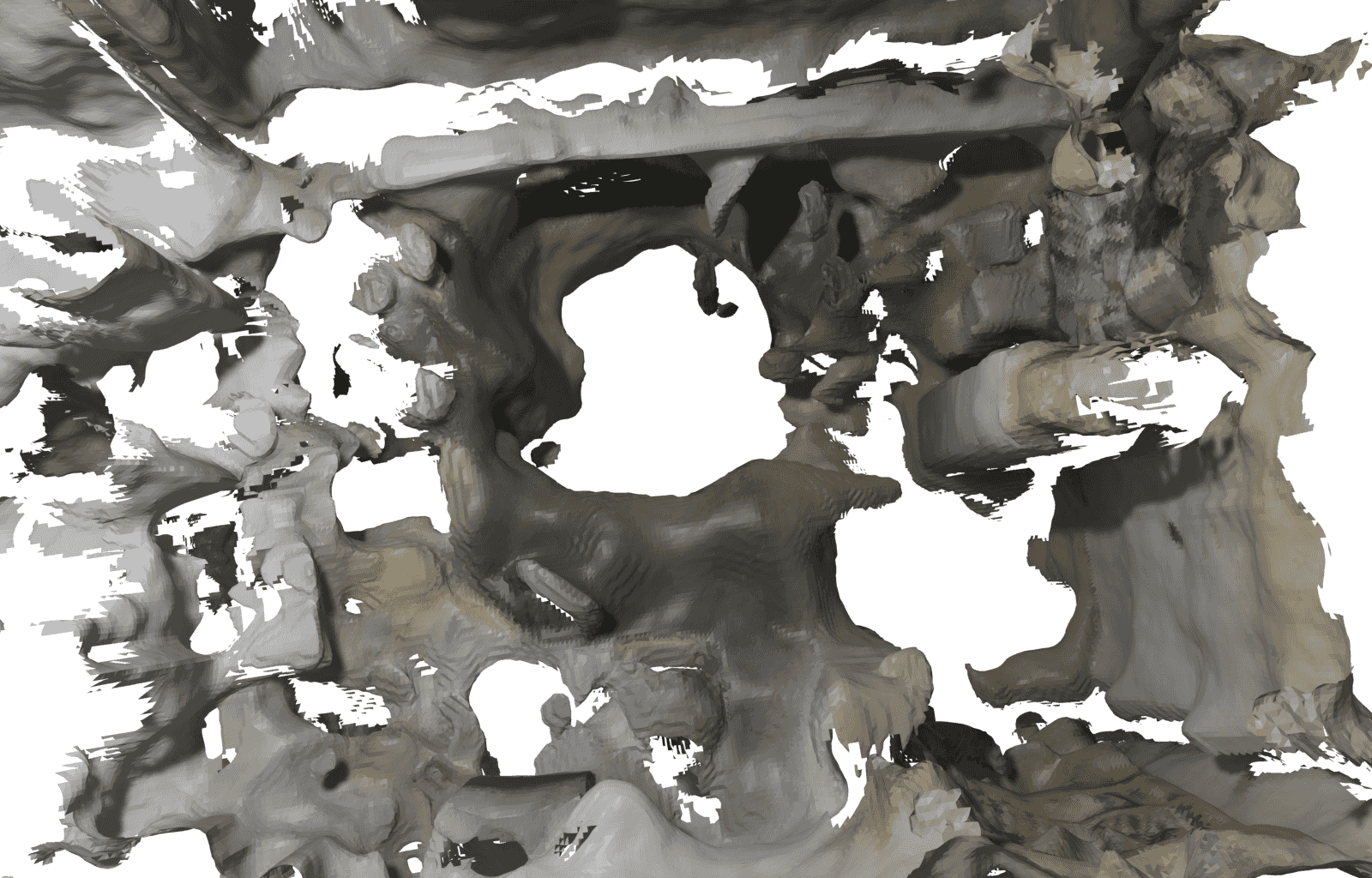}\vspace{\myvspace}
            \includegraphics[width=\widthOfMiniPage\linewidth]{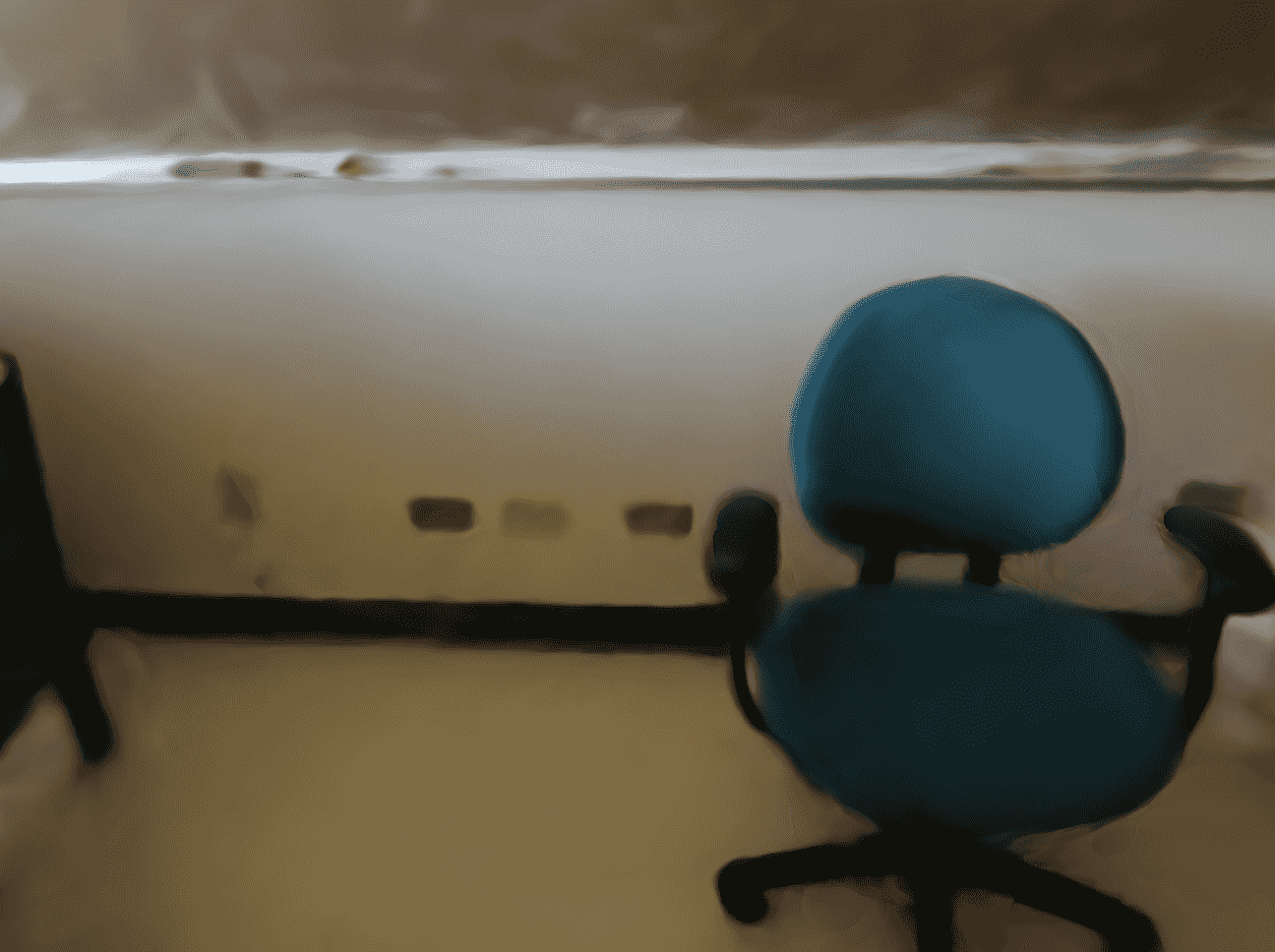}\vspace{\myvspace}
            \includegraphics[width=\widthOfMiniPage\linewidth]{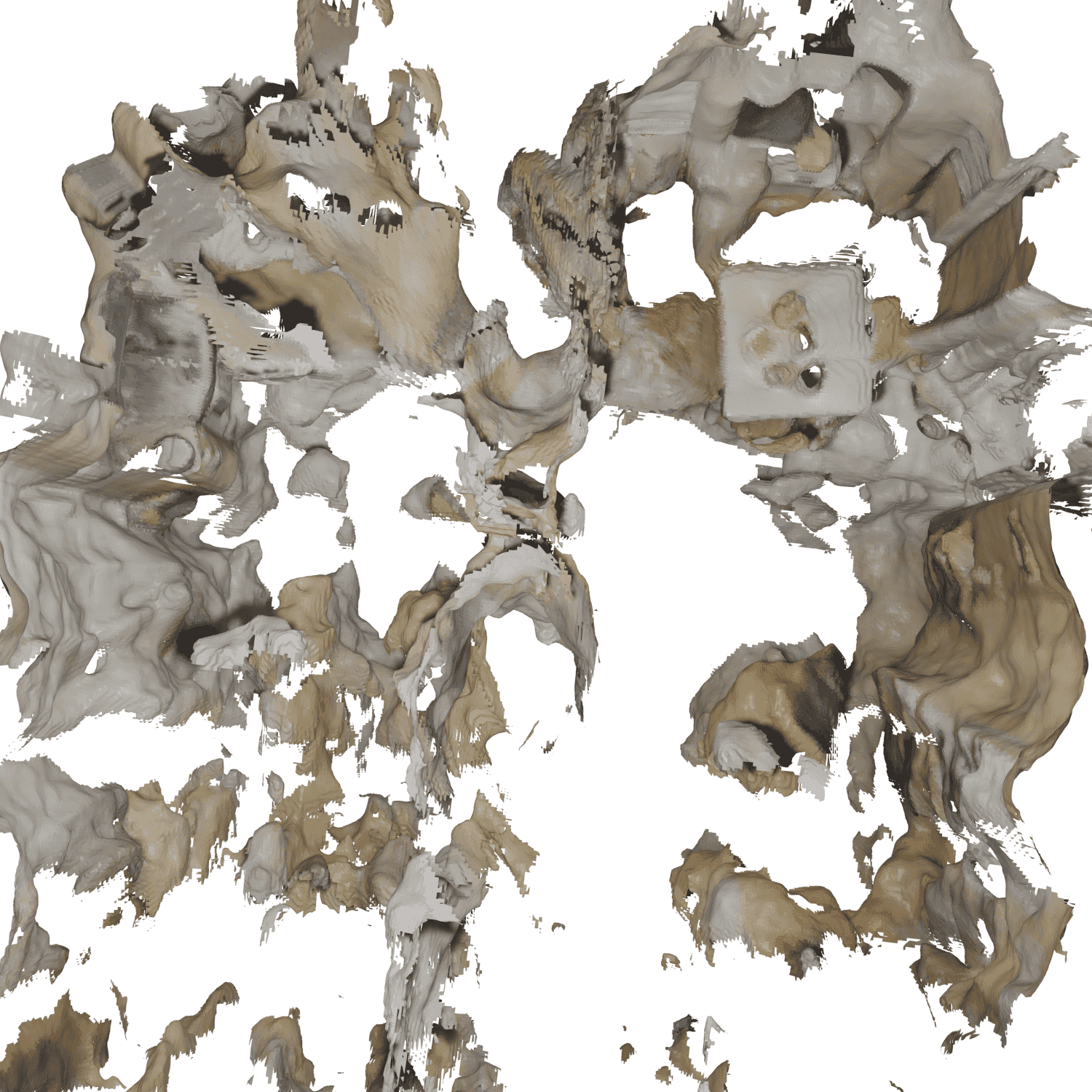}\vspace{\myvspace}
             \includegraphics[width=\widthOfMiniPage\linewidth]{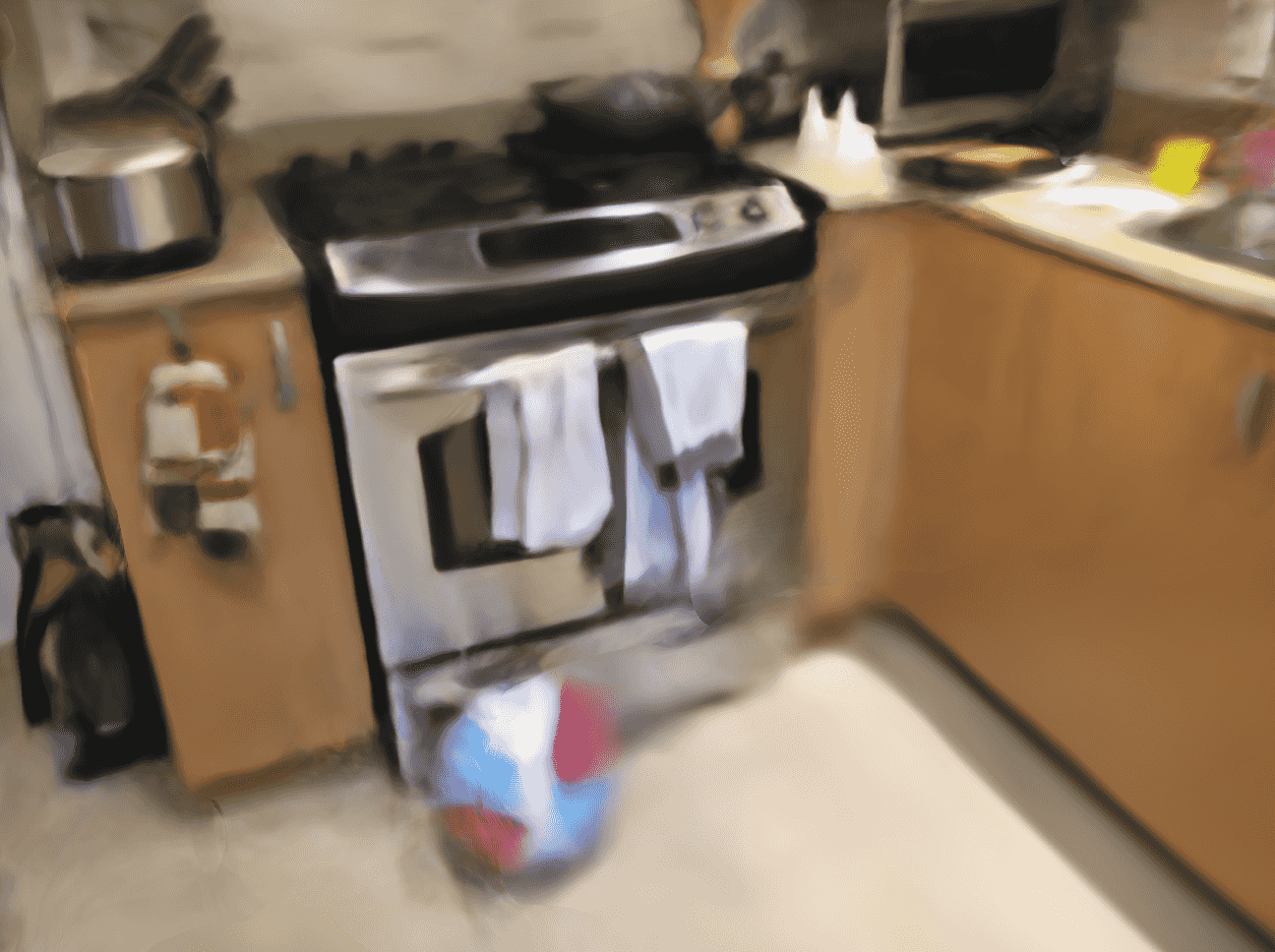}\vspace{\myvspace}
    \end{minipage}
  }
  \subfloat[VolSDF]{
    \begin{minipage}[b]{\widthOfFullPage\linewidth}
      \centering
            \includegraphics[width=\widthOfMiniPage\linewidth]{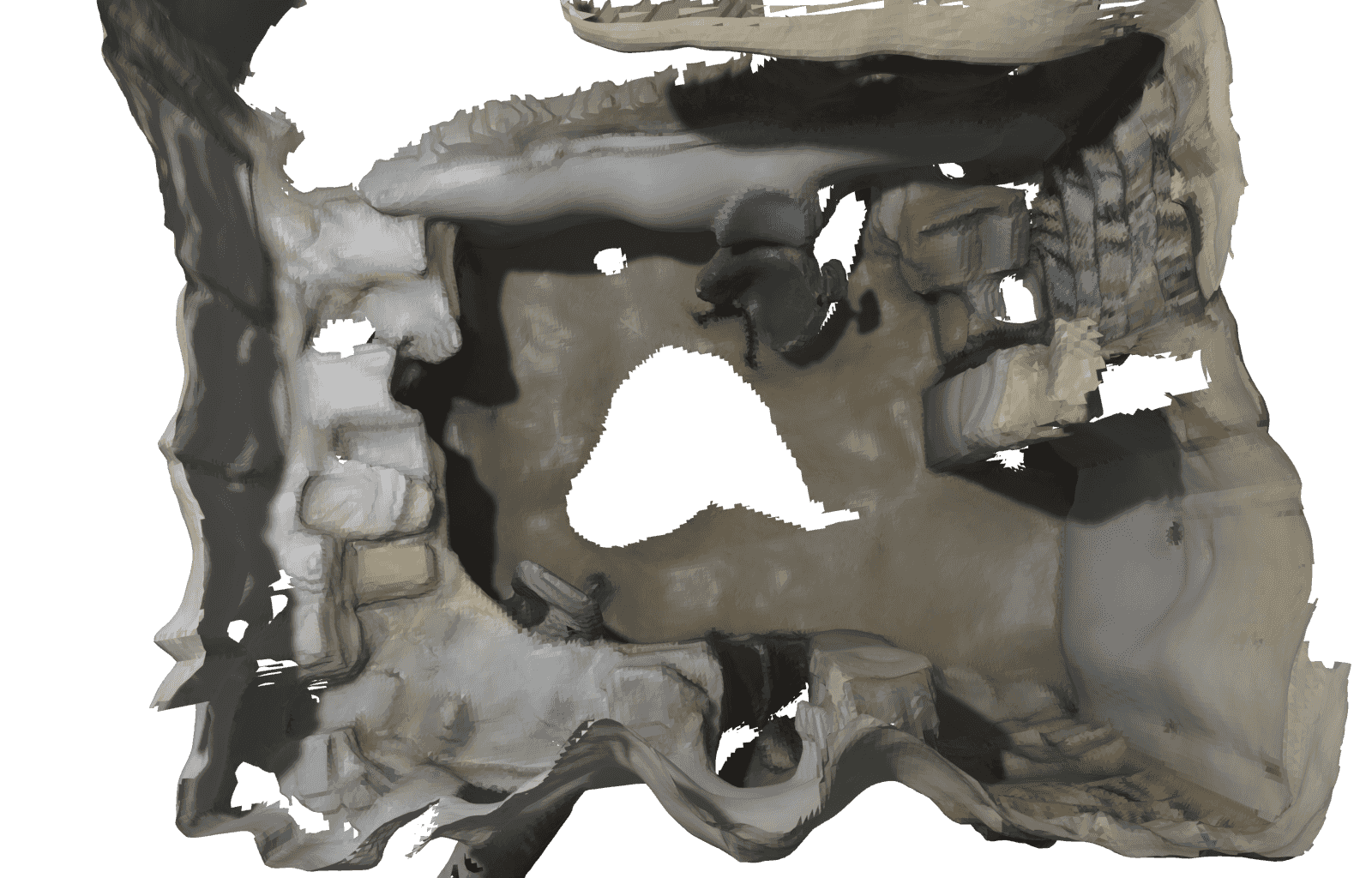}\vspace{\myvspace}
            \includegraphics[width=\widthOfMiniPage\linewidth]{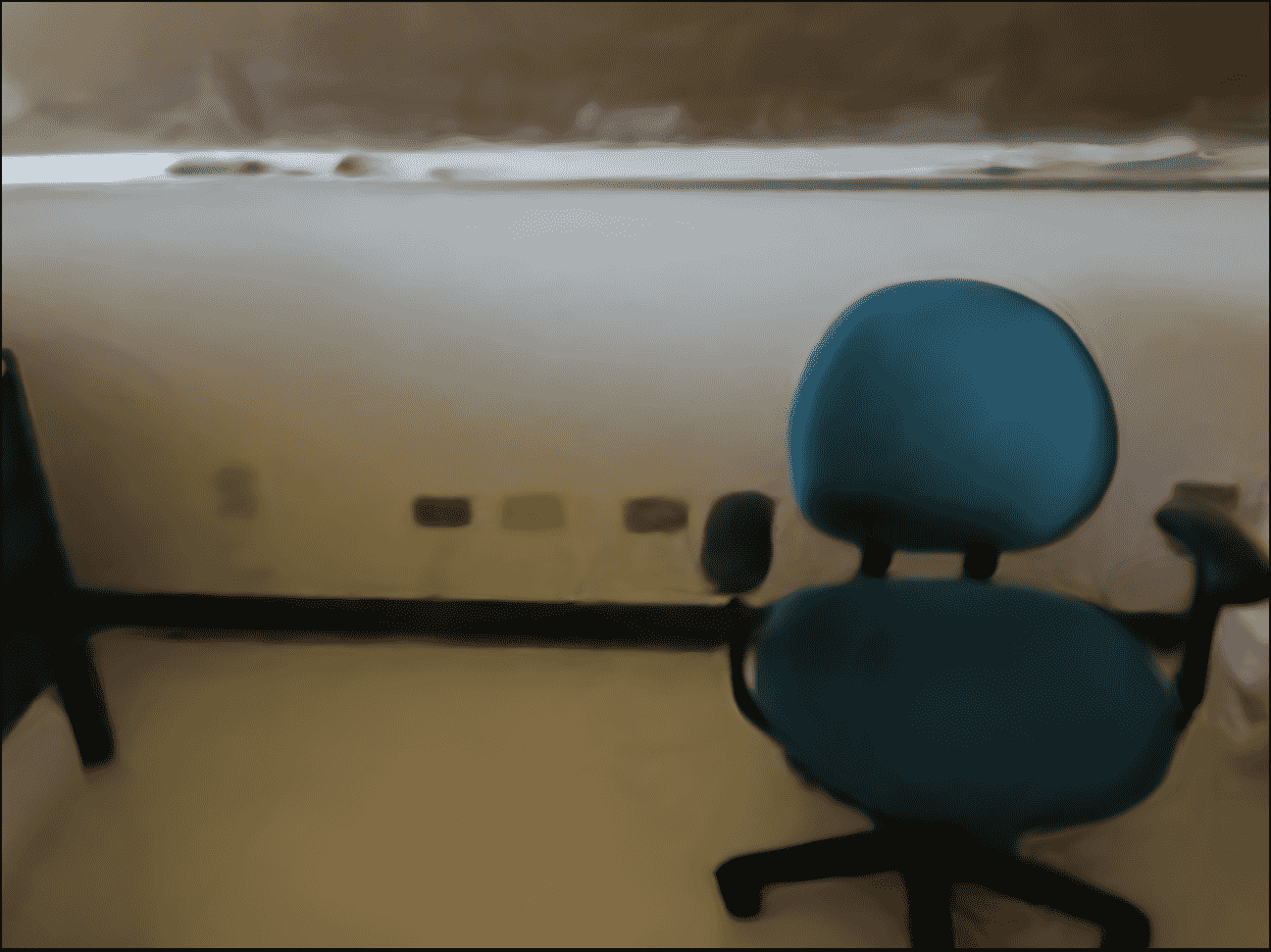}\vspace{\myvspace}
            \includegraphics[width=\widthOfMiniPage\linewidth]{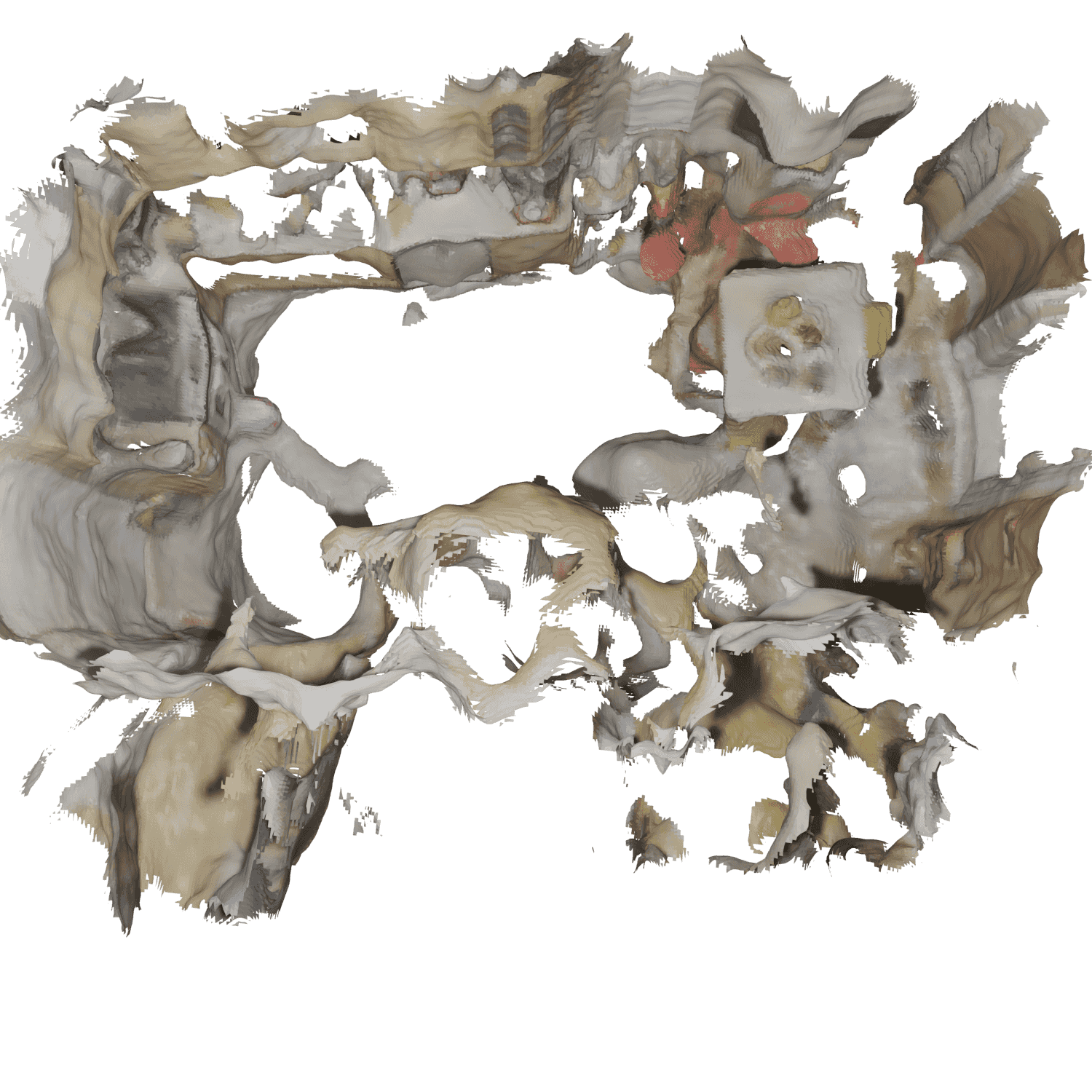}\vspace{\myvspace}
             \includegraphics[width=\widthOfMiniPage\linewidth]{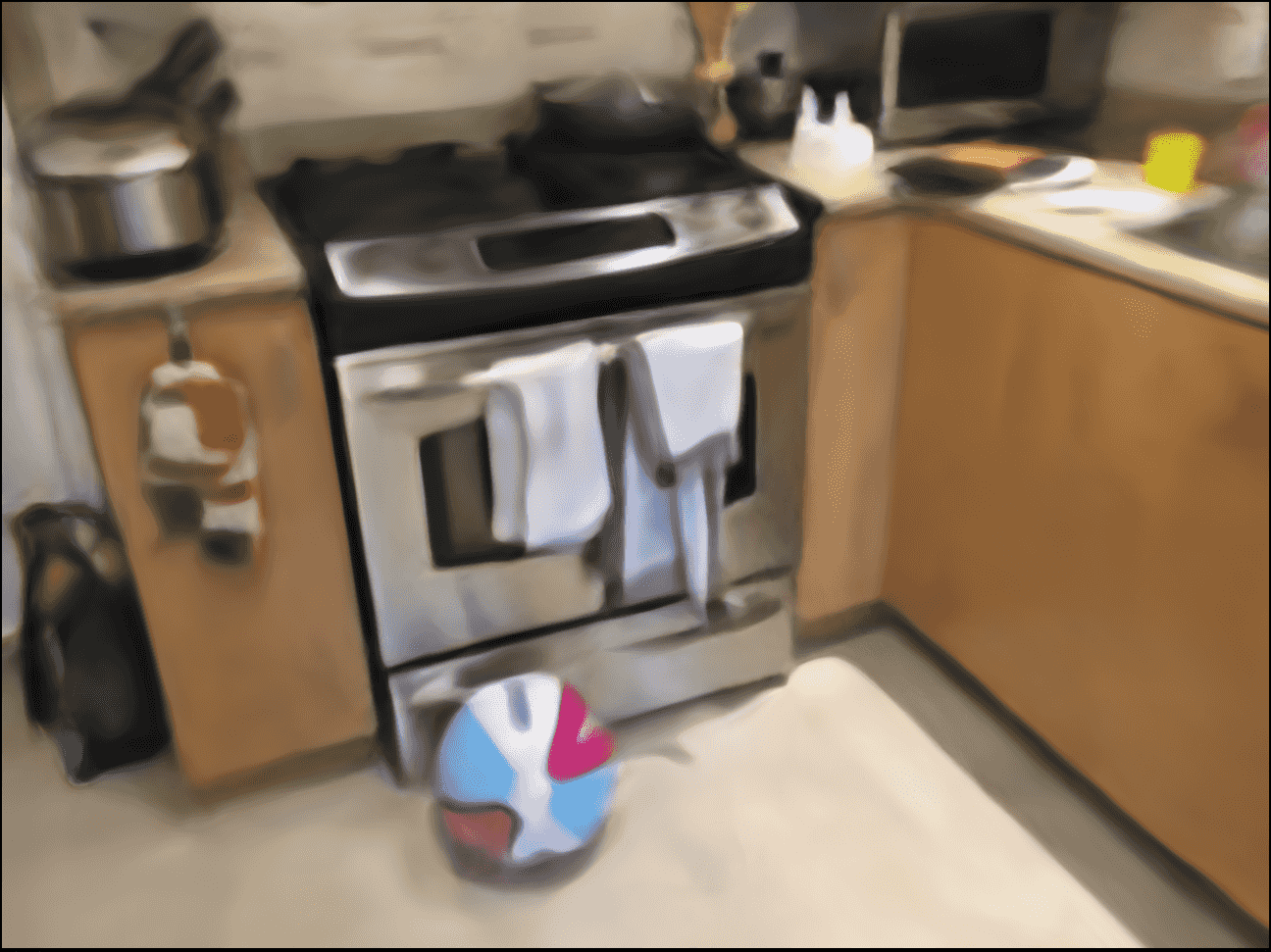}\vspace{\myvspace}
    \end{minipage}
  }
  \subfloat[Ours]{
    \begin{minipage}[b]{\widthOfFullPage\linewidth}
      \centering
            \includegraphics[width=\widthOfMiniPage\linewidth]{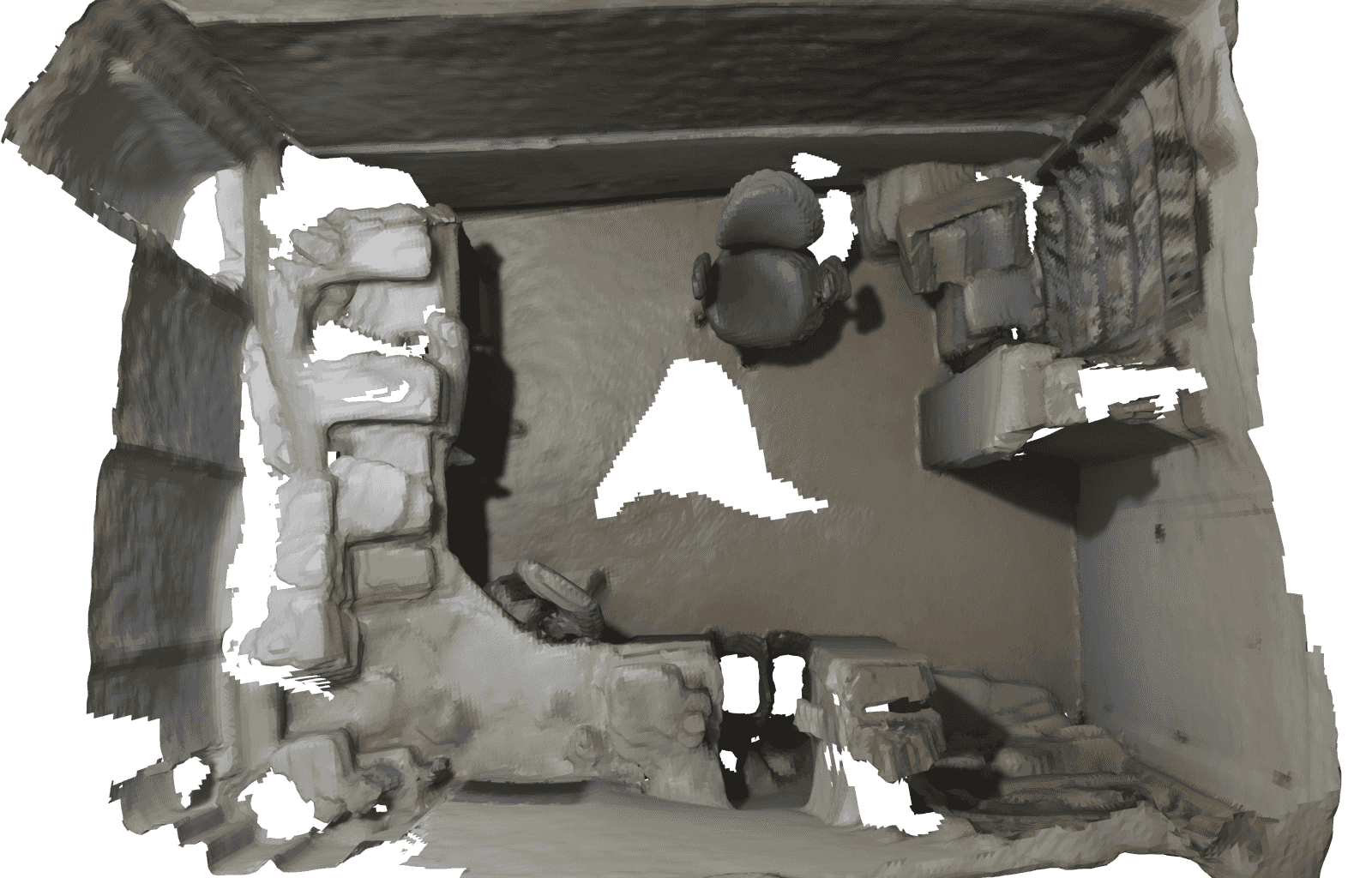}\vspace{\myvspace}
            \includegraphics[width=\widthOfMiniPage\linewidth]{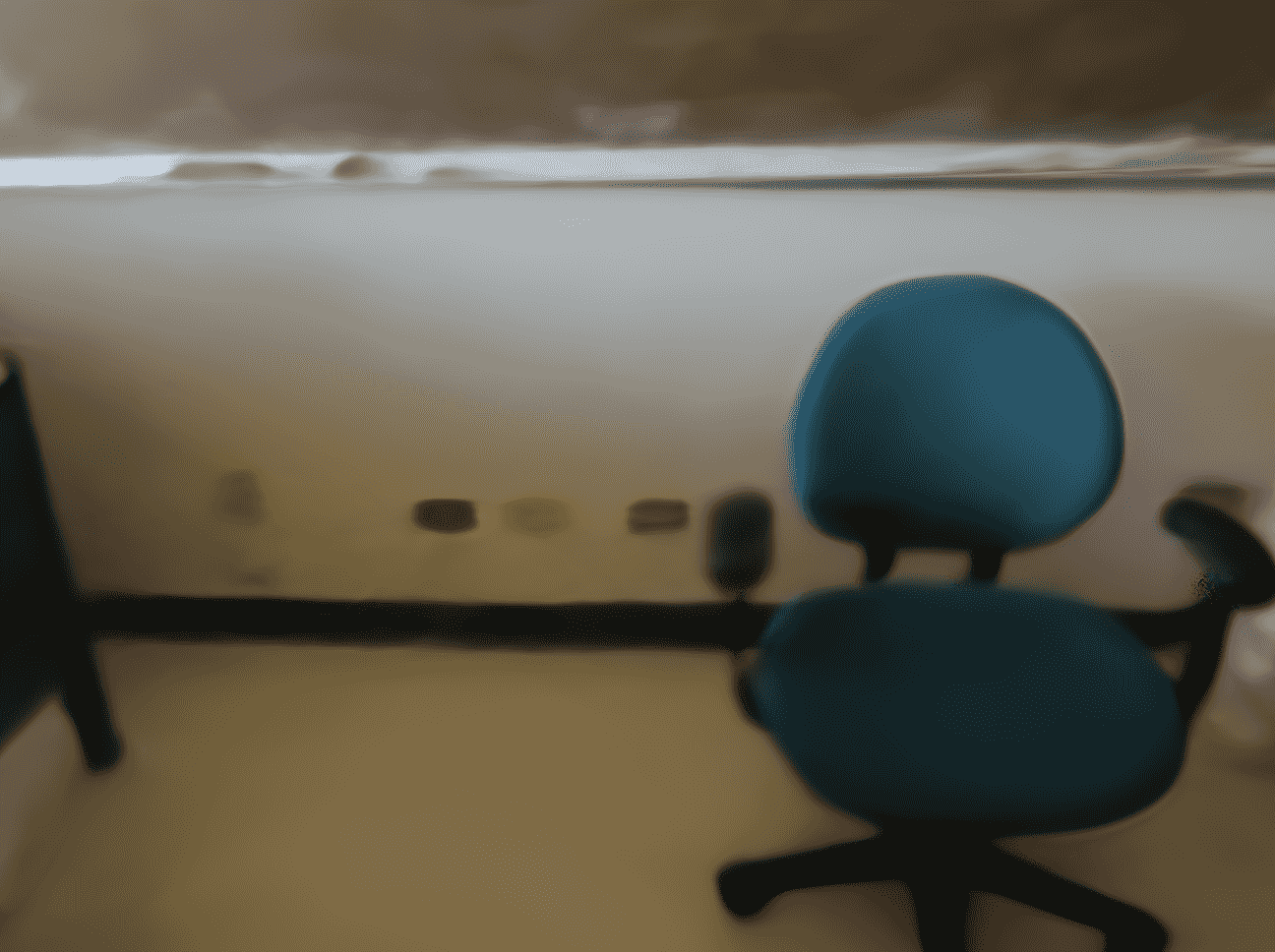}\vspace{\myvspace}
      \includegraphics[width=\widthOfMiniPage\linewidth]{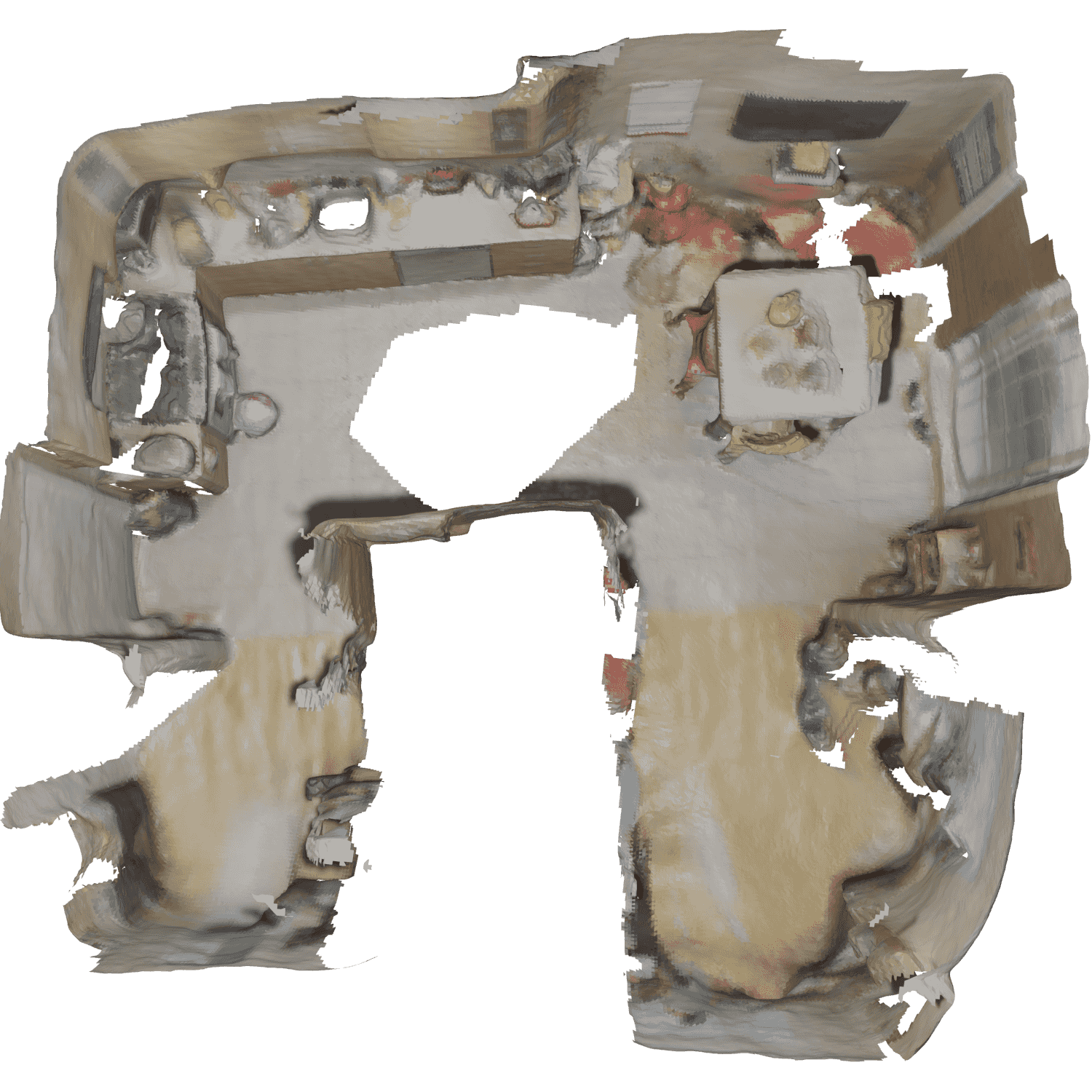}\vspace{\myvspace}
                        \includegraphics[width=\widthOfMiniPage\linewidth]{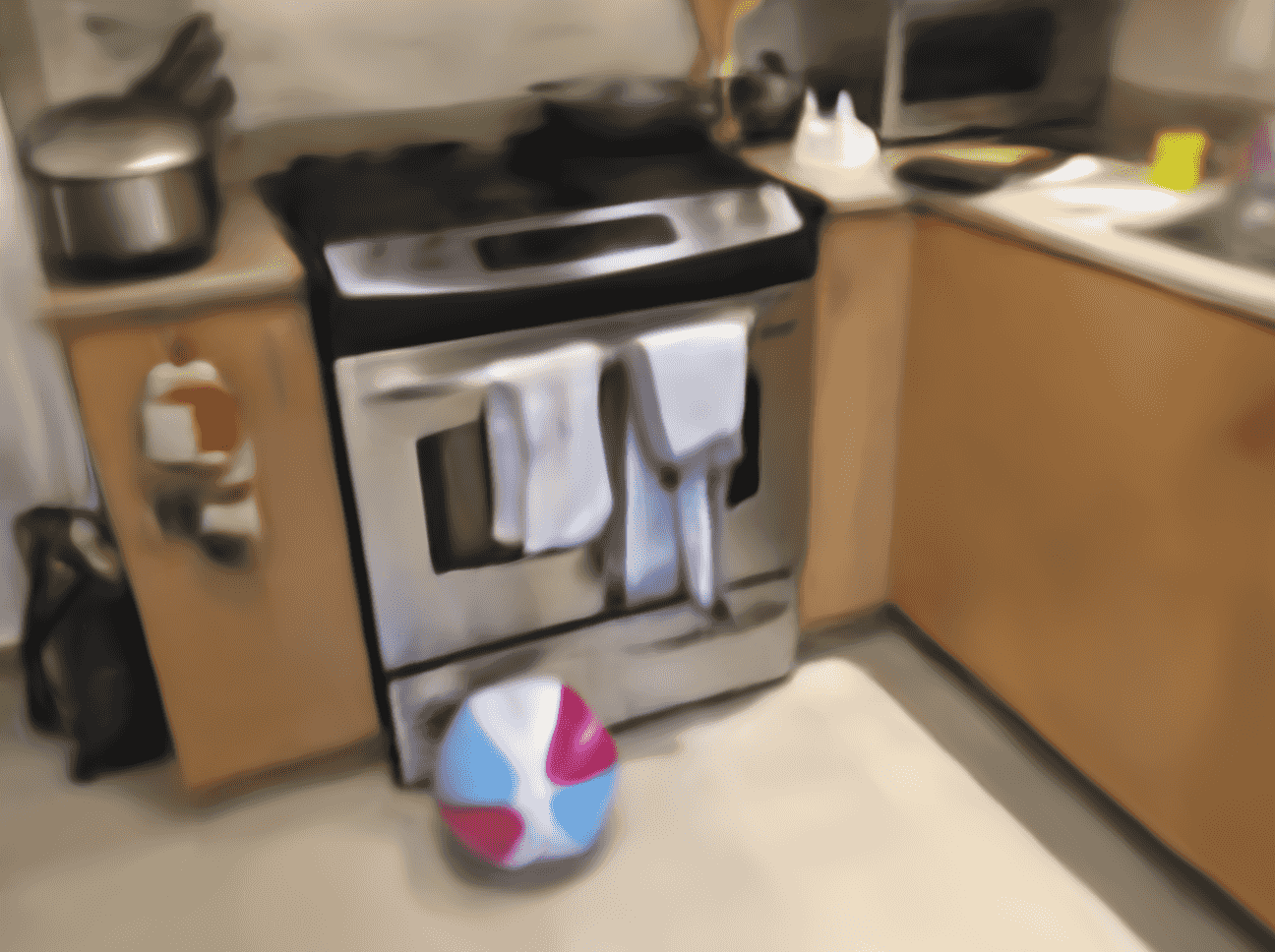}\vspace{\myvspace}

    \end{minipage}
  }
  \subfloat[GT]{
    \begin{minipage}[b]{\widthOfFullPage\linewidth}
      \centering
            \includegraphics[width=\widthOfMiniPage\linewidth]{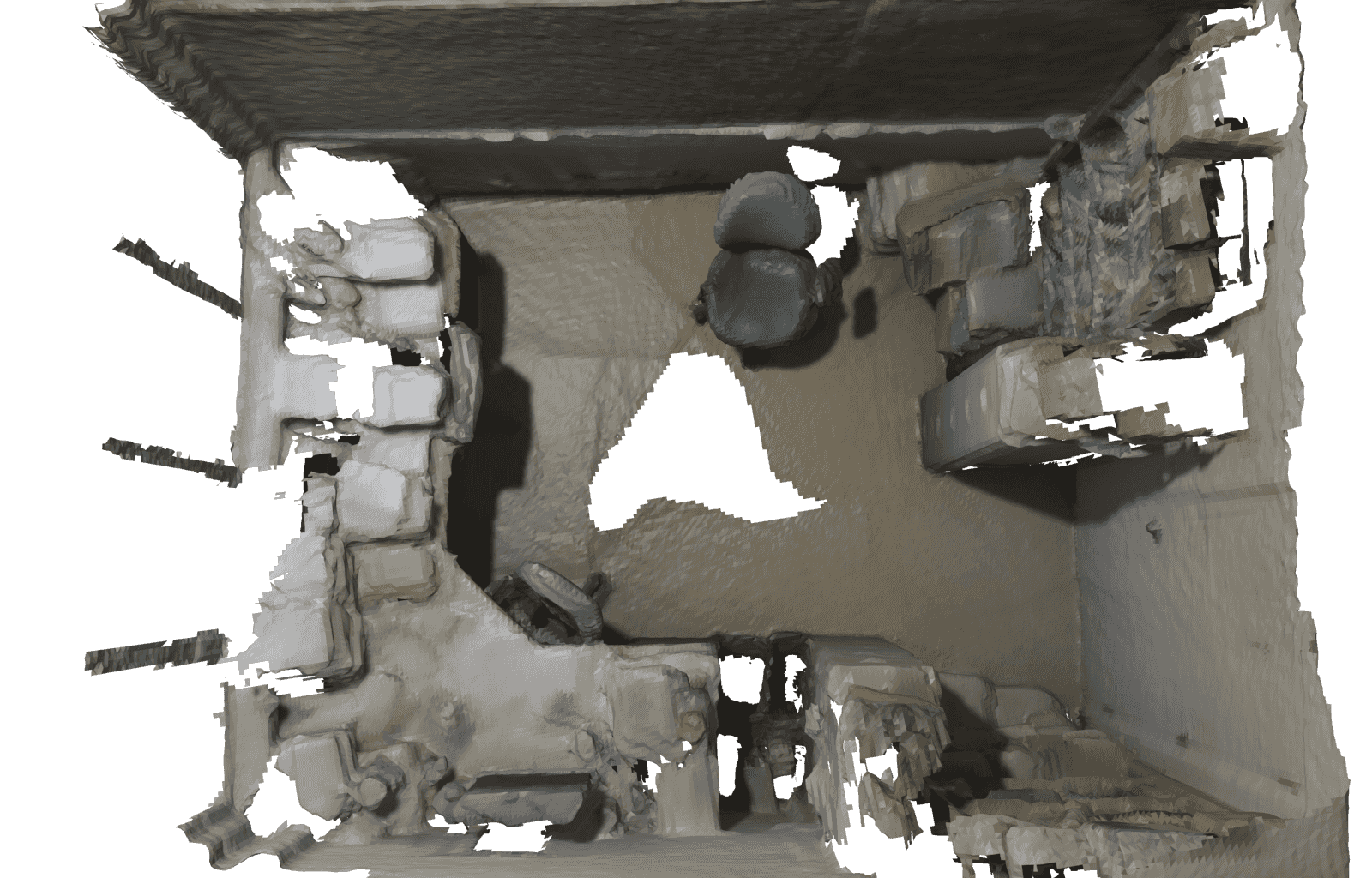}\vspace{\myvspace}
                        \includegraphics[width=\widthOfMiniPage\linewidth]{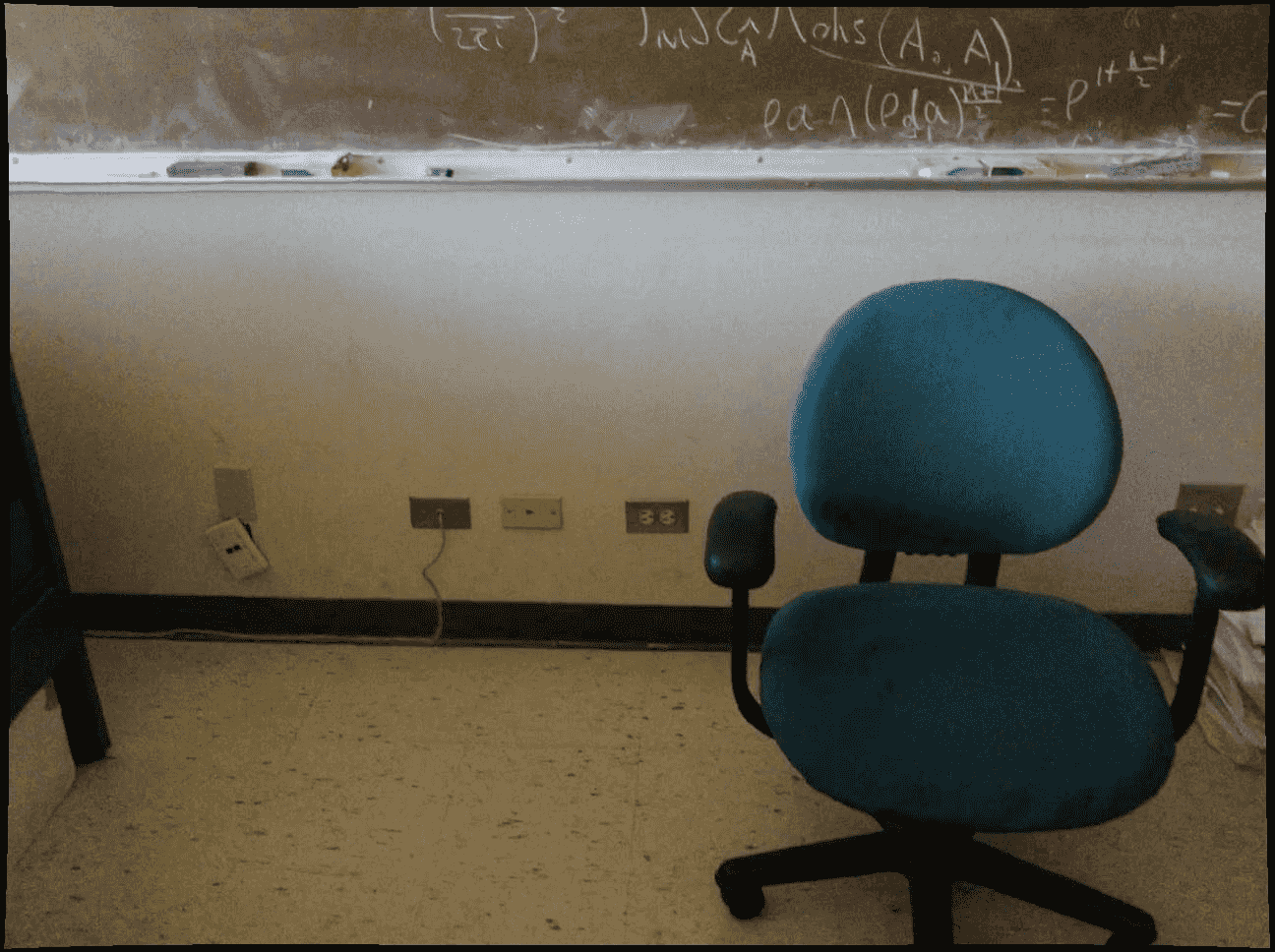}\vspace{\myvspace}
      \includegraphics[width=\widthOfMiniPage\linewidth]{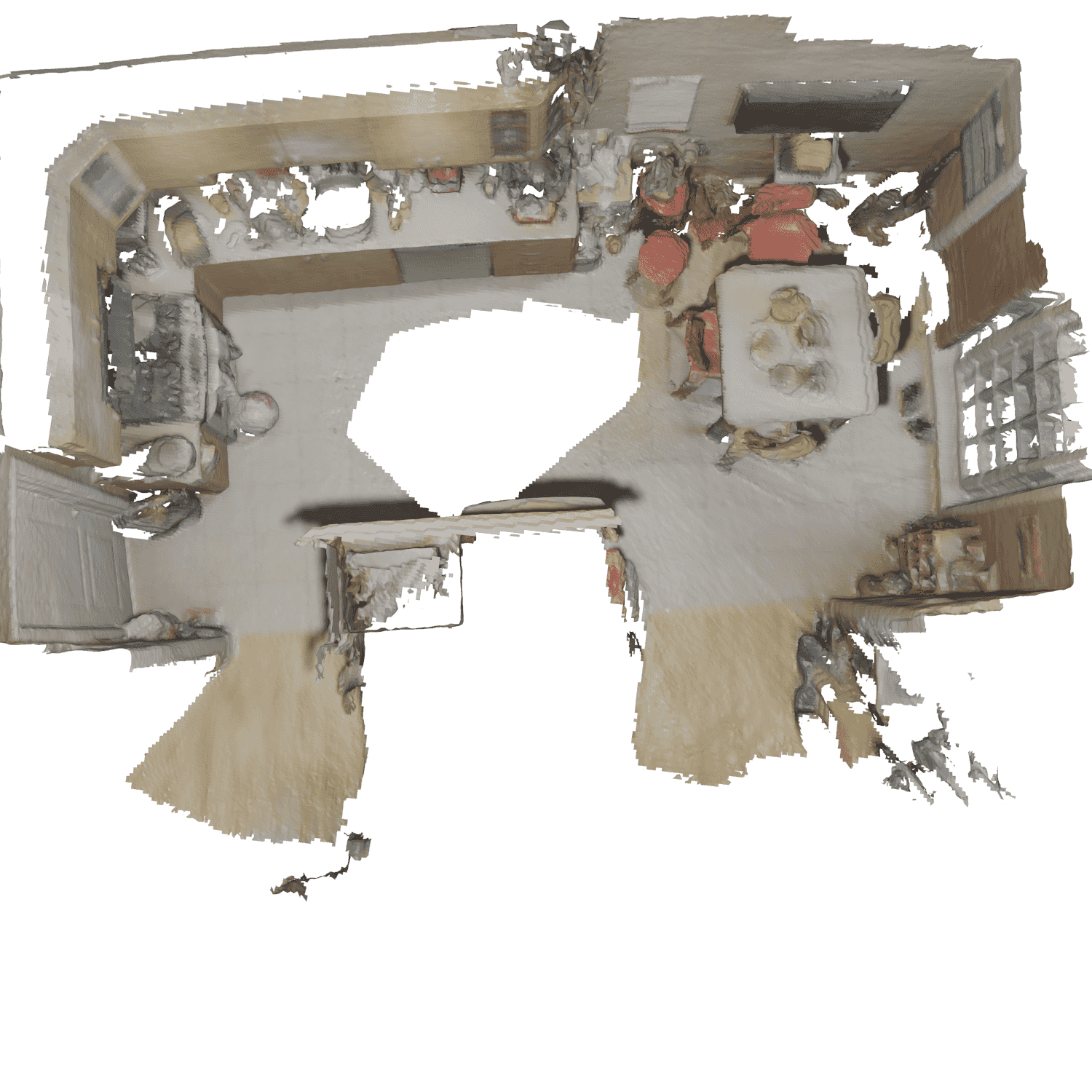}\vspace{\myvspace}
                        \includegraphics[width=\widthOfMiniPage\linewidth]{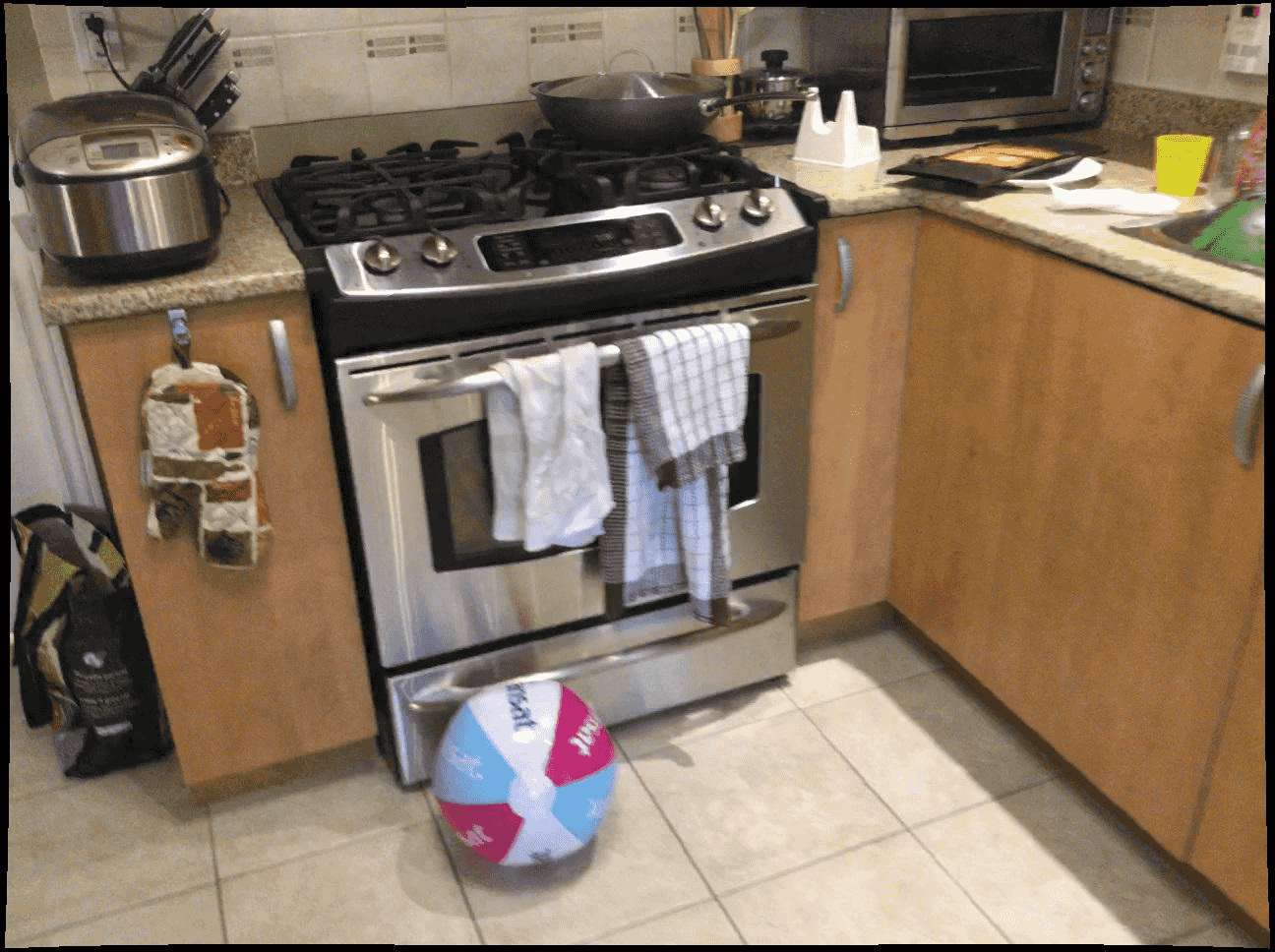}\vspace{\myvspace}

    \end{minipage}
  }
  \caption{Inherent shape-radiance ambiguity. For each set of results, the first row shows the reconstruction results. The second row shows rendering results from the implicit neural representation. State-of-the-art methods ~\cite{wang2021neus,oechsle2021unisurf,volsdf2021} can yield impressive high-quality reconstruction of a single object, while they often yield unsatisfactory reconstruction and rendering results for indoor scenes. Optimization of implicit neural representations easily falls into a local optimum, which will result in an incorrect reconstruction or even fail in reconstruction. }
 \label{shape-radiance-ambiguity}
\end{figure*}

\section{Introduction}

Reconstructing 3D geometries from multiview images is a fundamental topic in computer vision and graphics. The reconstructed model can be applied to VR/AR, video games, 3D printing and CAD manufacturing. Reconstruction of outdoor scenes and single objects with rich textures has been widely studied. However, few studies on indoor multiview reconstruction are directly based on color images. Objects in indoor scenes are usually of a single color, such as walls, furniture, and floors, which cannot be well restricted by photometric consistency.

MVS reconstruction algorithms have difficulty performing feature matching on texture-less areas, which leads to incompleteness and a large number of outliers.

Learning-based per-view depth estimation methods can obtain a good depth estimation for a single image. However, although it has good performance in quantitative evaluation, the fused 3D model is usually inaccurate due to the lack of consistency constraints between views. In addition, the areas with sudden depth changes are usually oversmoothed.

Another kind of learning-based reconstruction method directly extracts a 3D mesh from a volumetric representation.~\cite{murez2020atlas,sun2021neuralrecon} obtained effective reconstruction of indoor scenes. The reconstruction results of these methods generally have better completeness, but the scene details need to be further improved.

Recently, neural surface reconstruction methods have significantly promoted the development of 3D reconstruction, which achieves better reconstruction quality than conventional reconstruction approaches. In addition, these methods have the potential to reconstruct objects with non-Lambertian, less observed regions and complex geometry.

NeRF-based methods~\cite{mildenhall2020nerf} learn a function that maps 3D coordinates and 2D viewing directions to opacity and color values. However, since additional viewing directions need to be an additional input, the solution is not unique to explaining the input training images when lacking explicit or implicit regularization. Such a situation is called shape-radiance ambiguity~\cite{zhang2020nerf++,chai2000plenoptic,debevec1996modeling,buehler2001unstructured}, which means incorrect geometry can render an image consistent with the input training image. Shape-radiance ambiguity becomes the bottleneck in the reconstruction of indoor scenes for rendering-based surface reconstruction methods (Figure~\ref{shape-radiance-ambiguity}), although these methods ~\cite{wang2021neus,oechsle2021unisurf,volsdf2021} yield impressive reconstruction results on a single object.

\begin{figure}[h]
  \centering 
  \newcommand{\myvspace}{2pt} 
  \newcommand{\widthOfFullPage}{0.32} 
  \newcommand{\widthOfMiniPage}{0.98}
  \subfloat[Normal prior area]{
    \begin{minipage}[b]{\widthOfFullPage\linewidth}
      \centering
      \includegraphics[width=\widthOfMiniPage\linewidth]{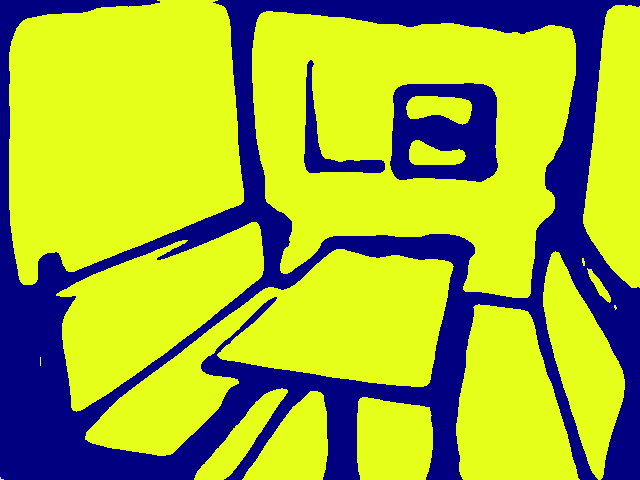}\vspace{\myvspace}
      \includegraphics[width=\widthOfMiniPage\linewidth]{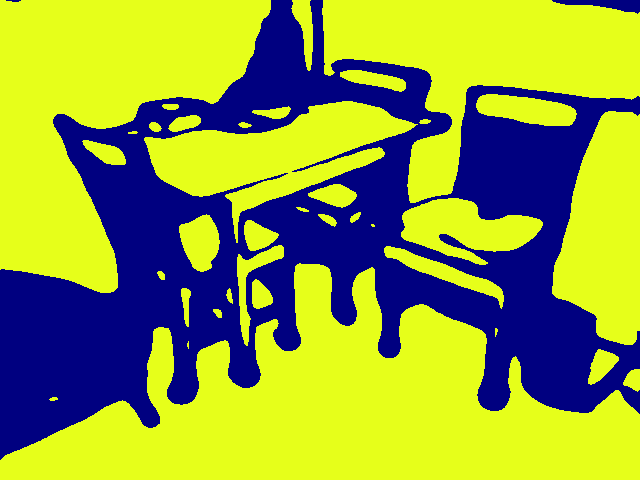}\vspace{\myvspace}
    \end{minipage}
  }
  \subfloat[Distance prior area]{
    \begin{minipage}[b]{\widthOfFullPage\linewidth}
      \centering
      \includegraphics[width=\widthOfMiniPage\linewidth]{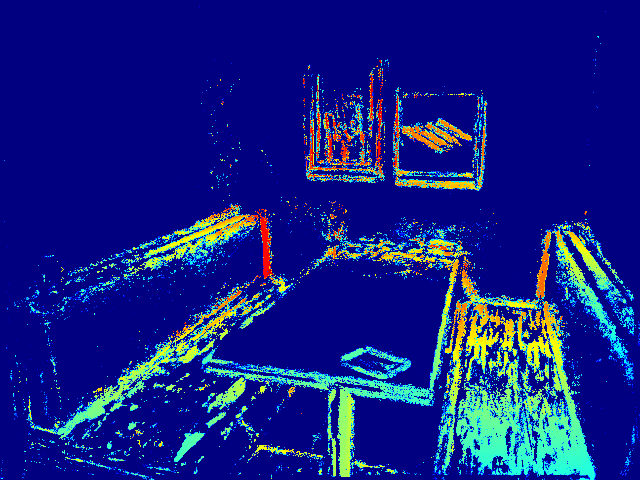}\vspace{\myvspace}
      \includegraphics[width=\widthOfMiniPage\linewidth]{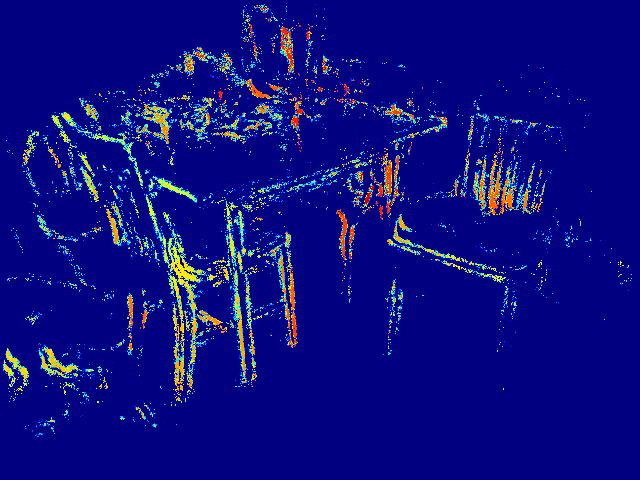}\vspace{\myvspace}
    \end{minipage}
  }
  \subfloat[Priors overlap]{
    \begin{minipage}[b]{\widthOfFullPage\linewidth}
      \centering
      \includegraphics[width=\widthOfMiniPage\linewidth]{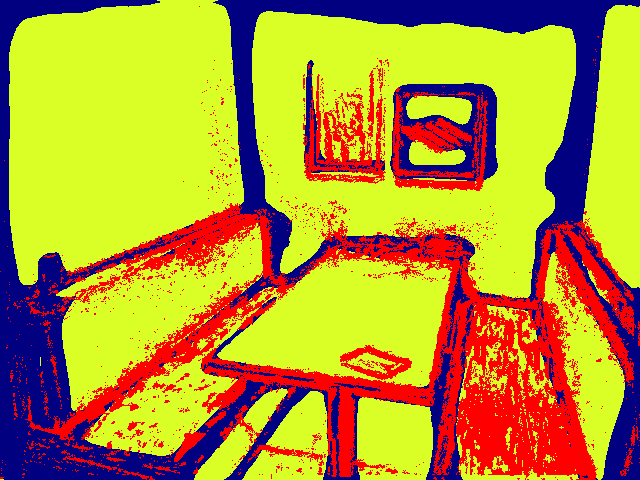}\vspace{\myvspace}
      \includegraphics[width=\widthOfMiniPage\linewidth]{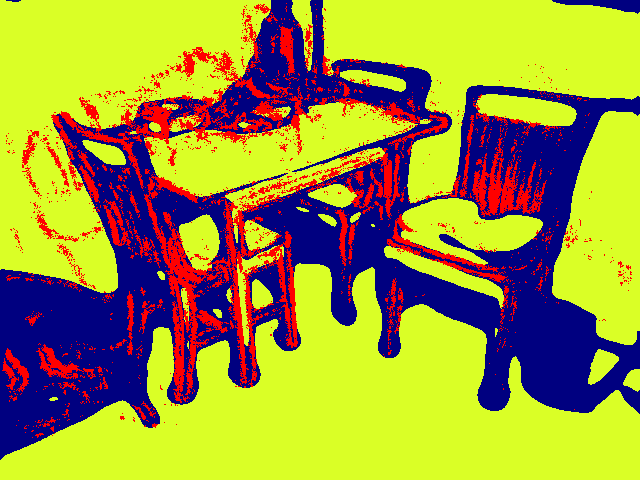}\vspace{\myvspace}
    \end{minipage}
  }
  \caption{Different areas in which priors take effect on. (a) shows the flat regions constrained by a reliable normal prior filtered by uncertainty, which is provided by ~\cite{bae2021estimating}. (b) shows the rich texture and edge regions constrained by the distance prior provided by COLMAP~\cite{schonberger2016structure}. (c) shows the overlapping area of (a) and (b). The rich texture regions and edges can have a good depth estimation. The texture-less regions can have an accurate normal estimation. The reliable depth and normal information are used to guide the optimization of neural implicit representation.}
  \label{hint}
\end{figure}

In this work, we propose a novel neural surface reconstruction method called NeuralRoom to assign appropriate global and local constraints between sampling rays to overcome shape-radiance ambiguity. We assume that the indoor scene is composed of two parts. The first part contains rich textured areas and edges such as object edges and various color decorations. The other part contains texture-less regions, such as walls and ground. We found that MVS can obtain high-precision estimates in textured areas but obtain relatively inaccurate results in texture-less areas. In contrast, neural network-based normal estimation methods~\cite{huang2019framenet,do2020surface,wang2020vplnet,bae2021estimating} always obtain a good estimation on flat areas but obtain inaccurate results on edges and rich textured areas (Figure~\ref{hint}). In our method, we combine their advances and propose a new smoothing term to alleviate shape-radiance ambiguity.

Specifically, our system consists of two main stages. The first stage is geometry prior acquisition (Figure~\ref{inputpriors}). We use multiview stereo to calculate the depth map and bounding box based on the depth fusion result. Then, we generate the distance prior from the depth map of each view. In addition, we use a learning-based normal estimation network to predict the normal map and filter it with uncertainty. The distance prior and normal prior are used to guide the next stage NeuralRoom differentiable renderer. 

The NeuralRoom renderer takes the above priors to optimize the implicit neural surface. The distance prior ensures the accuracy of reconstructed details and the normal prior limits the geometry feature of the texture-less region. To further improve the reconstruction quality of the flat regions, we conduct a smoothing method called perturbation-residual restriction for our NeuralRoom renderer. We assume that the sampling points on the local surface should be close to each other. If the local area is flat, the normal of sampling points should be the same. 

Our main contributions are the following:

\begin{itemize}
\item Porpose NeuralRoom, a novel neural surface reconstruction pipeline, for reconstructing indoor scenes. Researchers can use their own more advanced estimation methods which can yield more accurate priors and differentiable renderers to achieve better reconstruction results.
\item Introduce several efficient geometric priors for overcoming shape-radiance ambiguity. The distance prior acquired by MVS helps improve the detail accuracy. The normal prior helps enhance the completeness and accuracy of the texture-less region. 
\item Develop perturbation-residual restrictions working as smoothing terms to improve the accuracy and completeness of the flat and texture-less regions. There are some noise data in the priors, so this smoothing term can further enhance the reconstruction quality. 
\end{itemize}

Experimental results on the ScanNet dataset show that we have successfully applied multiview neural surface reconstruction for indoor scene reconstruction. Our complete framework has significantly improved state-of-the-art multiview reconstruction results on the tested indoor scenes. In addition, we use mobile phones to take photos of real-world scenes and successfully use our proposed system for reconstruction.

\section{Related work}

\subsection{Multiview Reconstruction}

Multiview reconstruction aims to reconstruct the three-dimensional geometric model of the scene from a set of images with or without calibrated camera poses. The key point of the image-based approach is photometric consistency assumptions. Among all kinds of methods, depth-map merging-based methods are the most widely used.

Multiview stereo has been widely studied~\cite{seitz2006comparison}. Traditional MVS methods~\cite{galliani2015massively,schonberger2016structure,xu2019multi} estimate the corresponding depth map for each input high-resolution image offline and then fuse it into the final three-dimensional model~\cite{merrell2007real,kazhdan2006poisson,bernardini1999ball}. These MVS methods often use the idea of sampling and propagation in PatchMatch to make depth estimation more effective. The learning based MVS approaches~\cite{yao2018mvsnet,ding2021transmvsnet,cheng2020deep,gu2020cascade,yang2020cost,wang2021patchmatchnet,yao2019recurrent,wei2021aa,yan2020dense,zhang2020visibility,kuhn2020deepc,chen2019point,xu2021self} have become popular in recent years and have shown some advantages in terms of accuracy and completeness in specific datasets. Most of the deep learning methods take MVSNet~\cite{yao2018mvsnet} as their skeleton and are based on plane-sweep stereo~\cite{collins1996space}. They use a convolutional neural network to extract higher dimensional 2D features in the image, usually following a 3D convolution to regress the depth for per pixel.
The 3D CNN is time- and memory-consuming due to the large number of parameters. The following works adopt a pyramid structure~\cite{gu2020cascade,cheng2020deep,yang2020cost,liao2021adaptive} or recurrent network~\cite{yao2019recurrent,yan2020dense,wei2021aa} to reduce the memory consumption. Some other work integrates PatchMatch ~\cite{wang2021patchmatchnet}, uncertainty~\cite{zhang2020visibility}, attention mechanism~\cite{ding2021transmvsnet}, semantic segmentation~\cite{xu2021self} and other mechanisms~\cite{kuhn2020deepc,chen2019point} into MVS approaches.

For indoor scenes, because there are a large number of weak texture areas in the scene, such as walls, floors and solid color furniture, the correspondence matching quality of MVS is usually poor, which leads to missing parts and outliers in the reconstruction. Previous works have attempted to reconstruct texture-less regions with different methods, such as multiresolution~\cite{xu2019multi}, planar priors~\cite{xu2020planar,sun2021phi}, and depth map completion~\cite{liu2020depth,kuhn2020deepc}.
These methods can improve the reconstruction results of the texture-less region to a certain extent and are better than the conventional methods in quantitative evaluation. However, they still cannot obtain a satisfactory result in the indoor scene.

Some learning-based multiview depth estimation and SLAM methods~\cite{long2021multi,long2021adaptive,rich20213dvnet,jiang2021plnet,teed2018deepv2d,wang2018mvdepthnet,hou2019multi} use data-driven methods to reduce the over-reliance on the photometric consistency assumption. They can quickly produce a low-resolution dense depth map. These methods can assign depth values to texture-less areas, but the areas with sudden depth changes are often oversmoothed. The 3D model acquired by TSDF fusion can show the general appearance of the scene, while the accuracy is poor. In addition, after obtaining the raw reconstruction through MVS, object or scene completion algorithms~\cite{sg-nn,dai2021spsg,zhang2022point,bokhovkin2022neural} can be used to complete and optimize the scene.

\subsection{Implicit Neural Representation}
Implicit neural representations represent scenes as a continuous implicit function, which can represent high-resolution geometries in finite memory. This implicit representation has been successfully used in novel view synthesis~\cite{muller2022instant,yu2021plenoxels,xu2022point,mildenhall2020nerf,roessle2021dense,liu2020neural,barron2021mip,tancik2022block,xiangli2021citynerf,niemeyer2021regnerf}, shape representation~\cite{atzmon2020sal,atzmon2019controlling,genova2019learning,mescheder2019occupancy,michalkiewicz2019implicit,park2019deepsdf,peng2020convolutional,muller2022instant}, human reconstruction~\cite{saito2019pifu,saito2020pifuhd}, relighting~\cite{philip2021free}, and multiview 3D reconstruction~\cite{wang2021neus,oechsle2021unisurf,niemeyer2020differentiable,yariv2020multiview,xu2022hybrid}.

~\cite{ji2017surfacenet,ji2020surfacenet+,choy20163d,kar2017learning,murez2020atlas,sun2021neuralrecon} presented end-to-end 3D reconstruction methods that use a global volumetric representation to assemble features from all views and then predict the 3D model directly from the feature volume.  ~\cite{liu2020neural,murez2020atlas} presented learning-based dense reconstruction methods for indoor scenes that can obtain better reconstruction than the previous methods in terms of quality and quantitative comparison. However, the details of the reconstruction results need to be improved.

Recently, differentiable rendering multiview 3D reconstruction methods have significantly advanced the development of 3D reconstruction and achieve better reconstruction quality than other conventional reconstruction methods on a single object. There are two types of rendering methods: surface rendering methods~\cite{yariv2020multiview,kellnhofer2021neural,niemeyer2020differentiable} and volume rendering methods~\cite{wang2021neus,yu2021plenoxels,muller2022instant,mildenhall2020nerf,oechsle2021unisurf}. Surface rendering-based methods assume that the radiance of a ray required for rendering is only related to the intersection of the ray and the geometric surface. This makes the gradient only backpropagated to the local area at the intersection. These methods usually cannot deal with complex geometries well and rely heavily on the object's mask. When the mask is missing, reconstruction usually fails. Volume rendering-based methods assume that the rendering color is related to the radiance and the corresponding alpha weight at all spatial locations through which the ray passes. Therefore, the gradient can be backpropagated to all sites involved in rendering. These methods can reconstruct complex scenes without a mask and deal with some scenes with sudden depth changes. However, it is usually unable to obtain high-precision geometric surfaces, especially for the texture-less region, due to the lack of geometric surface constraints. In addition, the reconstruction may contain conspicuous noises.


The rendering-based implicit neural representation shows the potential to replace traditional MVS reconstruction. However, its performance is not as good as that of traditional reconstruction methods in indoor scenes. NeRF-based methods~\cite{mildenhall2020nerf} map a 5D input (3D coordinates plus the 2D viewing direction) to opacity and color values, which easily suffers from inherent shape-radiance ambiguity due to the additional input dimensions and the lack of implicit or explicit regularization in the implicit representation.

Many works use the depth value as a constraint to alleviate this ambiguity~\cite{roessle2021dense,wei2021nerfingmvs,xu2022point,chen2021mvsnerf} in their respective fields. ~\cite{roessle2021dense,wei2021nerfingmvs} used the monocular depth estimation network optimized by MVS results to provide a distance prior for each pixel to reduce shape-radiance ambiguity and improve the effect of indoor depth estimation and novel view synthesis. However, the consistency between views of depth estimation is far from sufficient to obtain a satisfactory 3D model.

The concurrent works~\cite{monosdf,manhattanindoor2022,wang2022neuris} have ideas similar to those of our work. These methods use depth estimation to provide depth cues and use normal information as additional constraints to reconstruct indoor scenes.

\section{Overview}

\begin{figure*}[th]
  \centering
  \includegraphics[width=\textwidth]{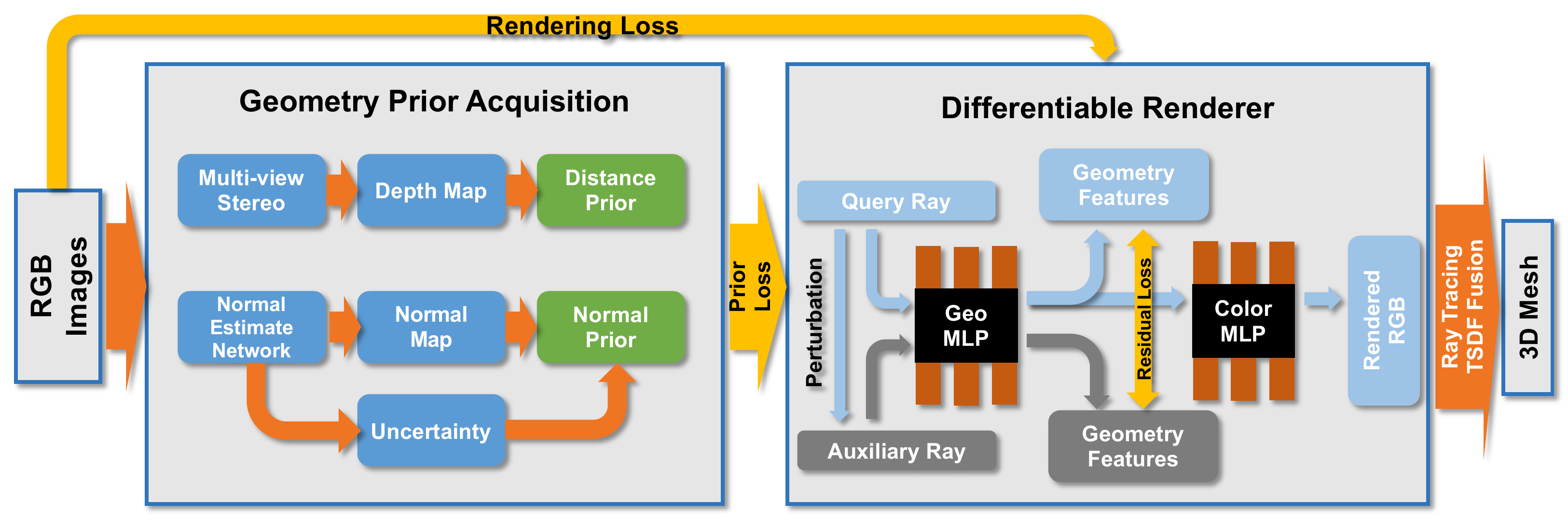}
  \caption{Method overview. The goal of our system is to reconstruct indoor scenes directly from RGB images with known camera parameters. The system consists of two parts. First, we use the multiview stereo method~\cite{schonberger2016structure} and the normal estimate network~\cite{bae2021estimating} to acquire the geometry prior. The distance prior acquired from MVS ensures the accuracy of texture-rich and edge areas, while the normal prior ensures the completeness of the texture-less region. Then, we use these geometry prior and RGB images to guide the optimization of the NeuralRoom module, which is a volume rendering-based neural surface reconstruction method. In addition, in the NeuralRoom module, we propose perturbation-residual restriction to constrain the implicit surface. Finally, we use ray tracing on the reconstructed scene and perform TSDF fusion to obtain the final 3D mesh model.}
 \label{overview}
\end{figure*}

\begin{figure}[h]
  \centering 
  \newcommand{\myvspace}{0.1pt} 
  \newcommand{\widthOfFullPage}{0.32} 
  \newcommand{\widthOfMiniPage}{0.98}
  \subfloat[Input image]{
    \begin{minipage}[b]{\widthOfFullPage\linewidth}
      \centering
      \includegraphics[width=\widthOfMiniPage\linewidth]{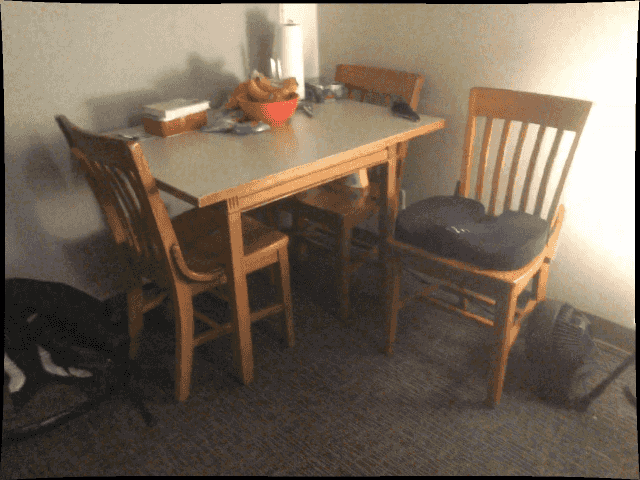}\vspace{\myvspace}
    \end{minipage}
  }
  \subfloat[Estimated depth]{
    \begin{minipage}[b]{\widthOfFullPage\linewidth}
      \centering
      \includegraphics[width=\widthOfMiniPage\linewidth]{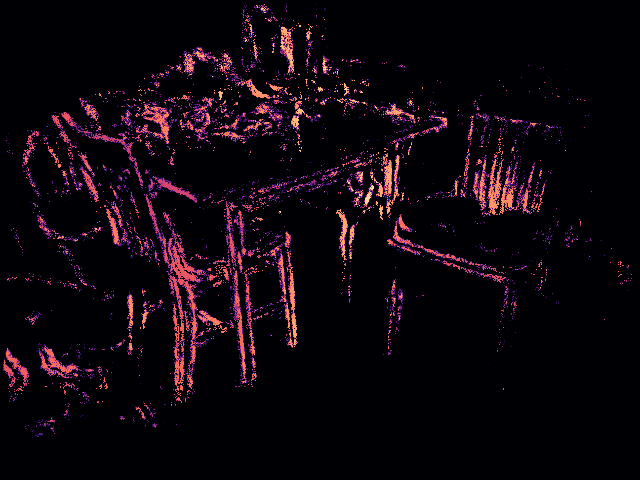}\vspace{\myvspace}
    \end{minipage}
  }
  \subfloat[GT depth]{
    \begin{minipage}[b]{\widthOfFullPage\linewidth}
      \centering
      \includegraphics[width=\widthOfMiniPage\linewidth]{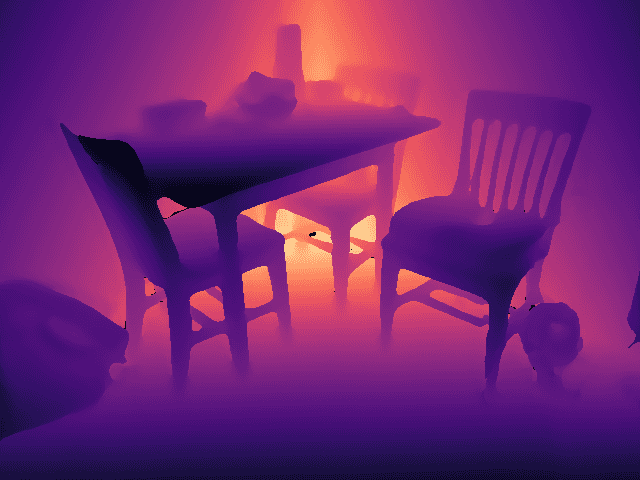}\vspace{\myvspace}
    \end{minipage}
  }
  \qquad 
  \subfloat[Normal uncertainty]{
    \begin{minipage}[b]{\widthOfFullPage\linewidth}
      \centering
      \includegraphics[width=\widthOfMiniPage\linewidth]{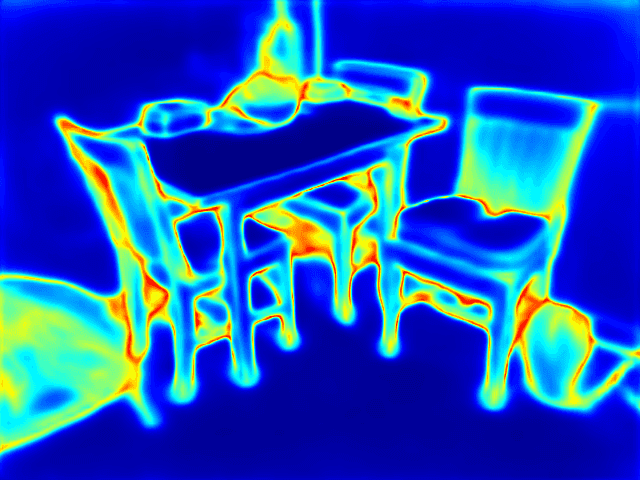}\vspace{\myvspace}
    \end{minipage}
  }
  \subfloat[Estimated normal]{
    \begin{minipage}[b]{\widthOfFullPage\linewidth}
      \centering
      \includegraphics[width=\widthOfMiniPage\linewidth]{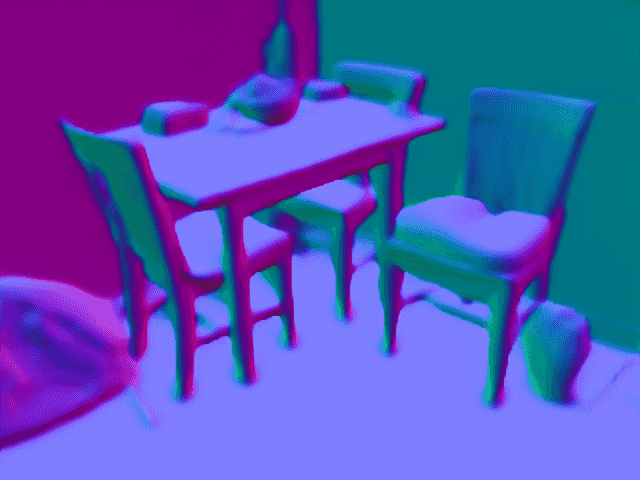}\vspace{\myvspace}
    \end{minipage}
  }
  \subfloat[GT normal]{
    \begin{minipage}[b]{\widthOfFullPage\linewidth}
      \centering
      \includegraphics[width=\widthOfMiniPage\linewidth]{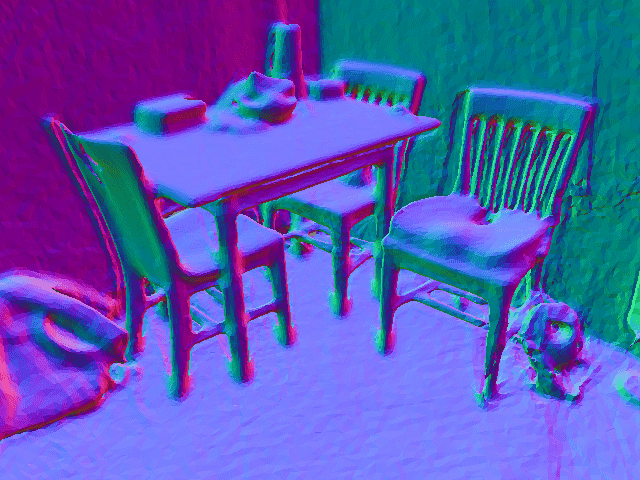}\vspace{\myvspace}
    \end{minipage}
  }
  \caption{Distance and normal priors. (a) Input RGB image. (b) Depth map estimated by COLMAP~\cite{schonberger2016structure}. (c) Ground truth depth map in the ScanNet~\cite{dai2017scannet} dataset. (d) Uncertainty acquired by~\cite{bae2021estimating} indicating the corresponding normal estimation quality. The normal estimation result in the bright colored areas is inaccurate. (e) Normal map estimated by~\cite{do2020surface}. (f) Ground truth normal map acquired by performing ray tracing on the ground truth mesh model. }
  \label{inputpriors}

\end{figure}


Shape-radiance ambiguity~\cite{zhang2020nerf++,chai2000plenoptic,debevec1996modeling,buehler2001unstructured} often exists in the Nerf-based rendering method~\cite{mildenhall2020nerf}. A well-trained implicit neural representation can achieve suitable outgoing 2D radiance at each wrong geometry surface point to perfectly fit a set of training images, as shown in Figure \ref{shape-radiance-ambiguity}. However, for the indoor scene, rectifying this ambiguity faces several challenges:
\begin{itemize}
\item Inaccurate camera pose.
\item Unstable light condition and photo imaging quality.
\item Many texture-less areas.
\item Inside-out shooting which makes some areas only visible from almost the same direction.
\item Large optimization space.
\end{itemize}

We introduce a novel multiview neural surface reconstruction method called NeuralRoom, which aims to handle the shape-radiance ambiguity in indoor scenes. Given a relatively accurate camera pose and varied observation directions, we use appropriate geometry cues to guide the optimization of neural representations and avoid falling into a local optimum. Specifically, we divide the scene into the textured and texture-less regions according to geometric features. For the textured and edge areas, we use the distance prior acquired by COLMAP~\cite{schonberger2016structure}, which is usually accurate. We give a high weight to make the implicit surface consistent with the distance prior. For flat and texture-less regions, we use the normal prior acquired by~\cite{bae2021estimating} to limit the local surface geometry. In addition, we propose smoothing terms called perturbation-residual restrictions to further improve the accuracy and completeness.

The method proceeds in three parts:
\begin{itemize}
\item Geometry prior acquisition. Section~\ref{sec4.1} introduces data preparation and how to acquire the distance and normal prior.
\item NeuralRoom renderer. Section~\ref{sec4.2} and Section~\ref{sec4.3} introduce the NeuralRoom renderer and the perturbation-residual restrictions.
\item Mesh extraction. Section~\ref{sec4.4} introduces our mesh extraction method.
\end{itemize}

\section{Method}

\subsection{Preprocessing}
\label{sec4.1}
\paragraph{Image Processing.}
First, we extract the images from the indoor video sequence. Usually, we would obtain thousands of photos. Because imaging quality will affect the accuracy of MVS, normal estimation, and rendering-based reconstruction, we use a Laplacian filter for blurring detection to determine which images should be used in the experiment. Given a set of images, we divided every ten images into a subgroup. Then we calculate the Laplacian of the source image as a blurring degree. Each subset leaves only one optimal image with the largest Laplacian in the subgroup. All optimal images generate the image set $I$ we used in the experiment.
\paragraph{Distance Prior.}
In the textured areas and edges, MVS can obtain high accuracy results. We hope to ensure the accuracy of these areas in NeuralRoom. We run COLMAP~\cite{schonberger2016structure}, a traditional multiview stereo method, on each selected image $I_i$ with fixed camera intrinsic parameters $K_i$ and extrinsic parameters $T_i=[R_i|t_i]$ to acquire a per-view depth map $D_i^{MVS}$. The depth map $D_i^{MVS}$ has been filtered by geometric consistency and eroded by 3 pixels. We set the values to zero for pixels where the depth is not defined. Therefore, the acquired depth map $D_i^{MVS}$ is sparser but more accurate.

We convert the depth to the distance between the camera center and the corresponding point for more convenient use in the optimization. We reproject the 2D coordinate to a 3D point in the world coordinate using $D_i^{MVS}$ and corresponding parameters $K_i$ and $T_i$:
\begin{equation}
    X_i(p) = T_i^{-1}  K_i^{-1} D_i^{MVS}(p)\ \tilde{p},\  where\ D_i^{MVS}(p)\neq0 ,
\end{equation}
where $p$ is the 2D pixel coordinate in $I_i$, $\tilde{p}$ is the homogeneous augmentation of $p$ , $X_i$ is corresponding 3D point. 

Each scene has a different size and location in world coordinates. Therefore, we normalize the scene into a cube with the center at the origin and a side length of 2. First, we fuse all depth maps into a point cloud and compute an axis-aligned bounding box. Then, we obtain the minimum bounding point $X^{BBox}_{min}$ and the maximum bounding point $X^{BBox}_{max}$. With $X^{BBox}_{min}$ and $X^{BBox}_{max}$, we can define the center of the scene in the real world as the translation $t_{opt}$ to the optimization coordinate. We then further define the scale $s$ which compresses the longest side length of the bounding box by $s$ times, so we ensure that the reconstructed scene falls into the NeuralRoom optimization space. Finally, we define the distance prior $D_i$ as
\begin{equation}
    D_i(p) = \Vert K_i^{-1} D_i^{MVS}(p)\ \tilde{p}\Vert/s,\  where\ D_i^{MVS}(p)\neq0  .
\end{equation}


\paragraph{Normal Prior and Uncertainty Acquisition.}
Texture-less regions cannot be effectively constrained by photometric consistency loss, which always suffers from shape-radiance ambiguity. Therefore, we use the surface normal acquired from the neural network to guide the optimization of NeuralRoom. We take UncertSurfaceNormal \cite{bae2021estimating} as our normal estimation module. \cite{bae2021estimating} is a learning-based normal estimation method. It takes a single image as input, and the output contains an estimated normal map and corresponding uncertainty map. We assume that the position with large uncertainty is where the surface geometry changes sharply. The area with a smooth surface generally has less uncertainty. 

The learning-based approaches always take a normal map calculated by a depth map as ground truth which is captured by a depth sensor. However, there is usually noise in the depth map obtained by the depth sensor. The depth varies greatly on edges, and the depth of special material surfaces is typically unable to be collected. In addition, the depth map acquired by the depth sensor needs to be converted to the image sensor coordinate system through extrinsic parameters. This transform also causes the depth value to be missing at the boundary. Therefore, the estimation of edges is usually inaccurate. We need an indication to determine which parts of the normal estimation results are reliable. Edge detection may also be an effective indication. In our approach, we take the uncertainty map given by~\cite{bae2021estimating}.

We feed the image $I_i$ into~\cite{bae2021estimating} and obtain the normal estimation $N^{raw}_i$ and corresponding uncertainty $U_i$. Then, we use uncertainty as an indication to filter $N^{raw}_i$ to obtain a reliable estimation result $N_i$:
\begin{equation}
    N_i = N_i^{raw} \cdot Bool(U_i <= mean(U_i)).
\end{equation}
We take the mean value of $U_i$ as the threshold to filter $N_i^{raw}$ and obtain a reliable estimation result $N_i$. We also set a zero value for pixels where the normal is filtered. We use $N_i$ as a normal prior to guide the optimization of NeuralRoom. In addition, the normal prior is also used in the perturbation-residual training method.

\subsection{NeuralRoom Module}
\label{sec4.2}
\paragraph{NeuralRoom Rendering Method.}
Our NeuralRoom renderer follows the basic volume rendering model NeuS \cite{wang2021neus} while integrating additional prior information and the perturbation-residual restriction.

NeuralRoom uses two multilayer perceptrons (MLPs) to represent geometry $f$ and color functions $c$. The geometry function $f: \Bbb{R}^3\to\Bbb{R}$ maps point $x\in\Bbb{R}^3$ to the signed distance to the object. The color function $c:\Bbb{R}^3\times\Bbb{S}^2\to\Bbb{R}^3$ maps a point position $x\in\Bbb{R}^3$ and a viewing direction $v\in\Bbb{S}^2$ to the RGB color space. The surface $\mathcal{S}$ of the object is represented by the zero set of its signed distance function (SDF):
\begin{equation}
    \mathcal{S} = \{x\in\Bbb{R}^3\vert f(x)=0\}.
\end{equation}

As a volume rendering method, the NeuralRoom renderer represents scenes as a colorized volume with weights and integrates radiance along with rays via alpha blending. Each pixel determines a ray. This scheme samples $n$ points along the ray $r$: $\{x_i = o + t_iv \vert i=1,2,...,n, t_i<t_{i+1}\}$. $o$ is the position from which the ray is emitted, which is usually the center of the camera. $v$ is the direction of the ray. $t$ is the length of the ray that has been emitted. 

Substituting a query spatial position $x_i$ into the geometry function $f$, an SDF estimation should be $f(x_i)$. Then, a unimodal density distribution function $\phi_s(f(x_i))$ is introduced. $\phi_s(x_i)=se^{-sx_i}/{(1+e^{-sx_i})}^2$ which is the derivative of the sigmoid function $\Phi_s(x_i)=(1+e^{-sx_i})^{-1}$, $s$ is a learnable parameter. The discrete opacity values $\alpha$ are shown to be:
\begin{equation}
    \alpha_i = max\bigl(\frac{\Phi_s(f(x_i)-\Phi_s(f(x_{i+1})))}{\Phi_s(f(x_i))},\ 0\bigr).
\end{equation}
The scheme approximates the color, normal, and length of this ray by calculating:
\begin{equation}
    \hat{C}(r) = \sum_{i=1}^n M_i\alpha_i c(x_i,v),
    \label{render1}
\end{equation}
\begin{equation}
    \hat{D}(r) = \sum_{i=1}^n M_i\alpha_i t_{i},
    \label{render2}
\end{equation}
\begin{equation}
    \hat{N}(r) = \sum_{i=1}^n M_i  \alpha_i  gradient(x_i),
    \label{render3}
\end{equation}
where $M_i = \prod_{j=1}^{i-1}(1-\alpha_j)$ indicates the accumulated transmittance, $\alpha_i$ is the per voxel opacity value acquired from implicit geometry function $f$, and $\hat{C},\hat{D},\hat{N}$ represent rendering color, rendering depth and rendering normal respectively. The derivative of implicit geometric function $f$ with respect to the three coordinate directions at point $p_i$: $(\frac{\partial f}{\partial x},\frac{\partial f}{\partial y},\frac{\partial f}{\partial z}) $ is the gradient at that position. The derivation process is completed by PyTorch's automatic derivation~\cite{2019PyTorch}.

\paragraph{Guided Optimization}
Let $p$ be the 2D pixel coordinate in $I_i$. To optimize NeuralRoom, we first sample some pixels from a specific image $I_i$ and generate their corresponding rays in world space ${P = \{C_i(p),D_i(p),\mathcal{N}_i(p),o_i,v_i(p)\}}$, where $C_i$ is the color of the pixel acquired from $I_i$, $D_i$ is the corresponding distance prior, which is the length between the camera center and the intersection of the surface, $N_i$ is the corresponding normal prior in world coordinates, $o_i$ is the camera center in world coordinates, and $v_i$ is the ray direction. $o_i$ and $v_i$ are calculated from $K_i$ and $T_i^{opt}$. We assume that the batch size of sampling rays is $k$, and we sample $n$ points along each ray.

As a rendering-based neural surface reconstruction method, its most important loss function is to minimize the difference between the rendered pixel colors and the ground truth input corresponding pixel colors:
\begin{equation}
    \mathcal{L}_{color} = \frac{1}{k}\sum\limits_k\left | \hat{C}_k - C_k \right |,
\end{equation}
where $C_k$ is the corresponding pixel color, $\hat{C}_k$ is the corresponding rendering color.

\paragraph{Prior Loss}
The prior loss $\mathcal{L}_{prior}$ consists of two parts, the distance prior loss $\mathcal{L}_{prior\_D}$ and the normal prior loss$\mathcal{L}_{prior\_N}$:
\begin{equation}
    \mathcal{L}_{prior} = \mathcal{L}_{prior\_D} + \gamma\mathcal{L}_{prior\_\mathcal{N}}.
\end{equation}
We force the distance $\hat{D}_k$ between the camera center and the intersection of the ray and implicit surface to consist of the distance prior $D_k$. If a sampled ray has a corresponding distance prior, we perform the distance prior loss to constrain the intersection to a specific position:
\begin{equation}
    \mathcal{L}_{prior\_D} = \frac{1}{n}\sum\limits_{j=1}^{n}SmoothL1(\hat{D}_k^j,D_k^j),
\end{equation}
\begin{equation}
    SmoothL1(A,B) = \begin{cases} 0.5 * {\Vert A-B \Vert} ^2 / beta, & \Vert A-B \Vert <= beta, \\ \Vert A-B\Vert - 0.5 * beta & \Vert A-B\Vert>beta, \end{cases}
    \label{con:smoothl1}
\end{equation}
where $\hat{D}_k^j$ is the rendered depth if the pixel has a distance prior, $A$ and $B$ are two vectors of the same dimension used to illustrate $SmoothL1$, $SmoothL1$ is the smooth L1 loss ($beta=0.1$), $D_k$ is the corresponding distance prior, and $n$ is the number of rays that have a distance prior. Therefore, the intersection positions between these rays and the implicit surface are consistent with the distance prior. The distance prior makes the reconstruction results consistent with the MVS results in texture-rich areas and where the surface geometry changes sharply. In addition, since the rendering loss is dominant in the optimization process, this will reduce the impact of outliers in MVS results on reconstruction.

Next, we need to solve the most important problem of indoor reconstruction: shape-radiance ambiguity in the texture-less region. The normal prior loss is:
\begin{equation}
    \mathcal{L}_{prior\_\mathcal{N}} = \frac{1}{n}\sum\limits_{j=1}^n SmoothL1(\hat{\mathcal{N}}_k^j,\mathcal{N}_k^j),
\end{equation}
where SmoothL1 ($beta=0.2$) is defined in Equation \ref{con:smoothl1}, $\mathcal{N}_{k}^j$ is the valid normal prior, $\hat{\mathcal{N}_{k}^j}$ is the rendered normal, and $n$ is the number of rays that have a valid normal prior. This optimization term makes the surface normal in a texture-less region consistent with the normal prior. However, due to the noise in the priors, the lack of distance prior and inaccurate camera poses, the reconstructed texture-less surface may be unsmooth and even appear to have severe discontinuity caused by shape-radiance ambiguity. Therefore, we propose perturbation-residual restrictions to improve the reconstruction quality.
\subsection{Perturbation-residual Restrictions}
\label{sec4.3}

\begin{figure}[h]
  \centering
  \includegraphics[width=\linewidth]{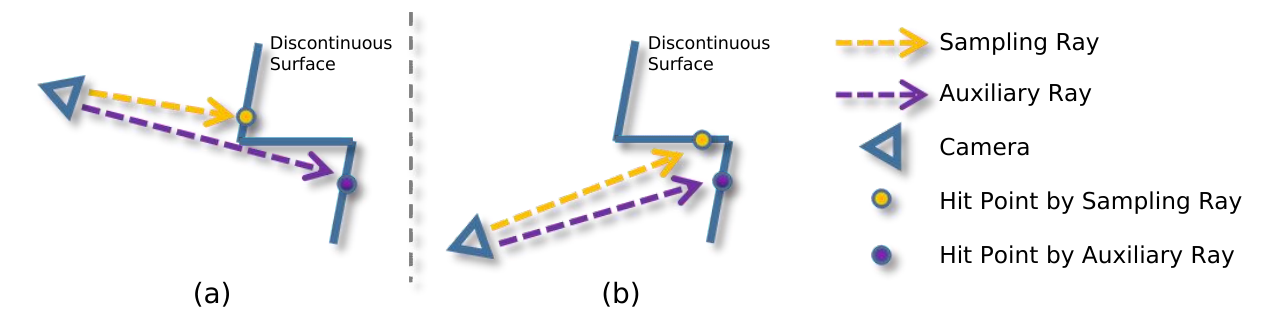}
  \caption{Two cases illustrating perturbation-residual restriction. The surface should be continuous but break in the cases. In case (a), the two hit points are far from each other with the same surface normal. Thus, $\mathcal{L}_{consist\_N}$ is small, and $\mathcal{L}_{smooth\_D}$ is large. In case (b), the two hit points are near each other, but the surface normal is quite different, which leads to a small $\mathcal{L}_{smooth\_D}$ and a large $\mathcal{L}_{consist\_N}$. Only when the local surface is smooth, the two losses will be small at the same time.}
 \label{pert}
\end{figure}

Although we use two priors to guide the optimization, still we face some problems. One is that the depth and normal estimation results always have noise. Another is that the normal estimation of a region may be inconsistent. The inaccurate camera pose, unstable light condition and poor photo imaging quality easily cause the above situation. 

The derivative of geometric network $f$ at the sample point is the normal of that position. The normal obtained by this computing method is a very local constraint. Because the sampling ray is discrete and random, the normal condition can only ensure that the normal corresponding to the sampling ray is consistent with the prior and cannot affect a larger neighborhood. The rendered color loss cannot play an effective role in the texture-less region, and the distance prior generally cannot be obtained by MVS. At the same time, the normal prior can not uniquely determine a spatial location, resulting in the optimized surface geometry fluctuating and even breaking into many parts. 

A natural solution, which may be the best, is to render the depth and normal of all rays in a patch from the net $f$ to give geometric constraints. However, each sampling ray needs to query the network hundreds of times to obtain the data required for rendering. Moreover, rendering everything in a patch will consume considerable memory and computing time. Therefore, we propose a compromise approach called perturbation-residual restriction to establish connections between sampling rays, making the reconstructed surface continuous.

The perturbation-residual restrictions assume that in a small region, the sampling points should have the same normal, and the distances to the camera center should be similar. We divide each optimization step into two stages. In the first stage, we randomly sample $k$ rays as before and render their color $\hat{C}$, depth $\hat{D}$ and normal $\hat{N}$. In the next stage, we perturb the ray direction. With the new sampling ray $P = \{o_k,v_k^{pert}\}$, we query the net $f$ again and render the corresponding depth $\hat{D}^{pert}$ and normal $\hat{N}^{pert}$. Then the residual is calculated with the first stage. The ray direction corresponding to a pixel $p_k:(u,v) \in I_i$ in the world coordinate is
\begin{equation}
    v_k = R_i^{-1}Normalize(K_i^{-1}\begin{pmatrix} u\\v\\1 \end{pmatrix}),
\end{equation}
\begin{equation}
    v_k^{pert} = R_i^{-1}Normalize(K_i^{-1}\begin{pmatrix} u + (a-0.5) * w\\v + (b-0.5)*w\\1 \end{pmatrix}),
\end{equation}
where $K_i$ is the intrinsic matrix of $I_i$, $R_i$ is the rotation matrix belonging to the extrinsic matrix $T_i$, $Normalize$ is the normalization function that normalizes the length of the vector to be 1, random variables $a,b \sim U[0,1]$ obey the uniform distribution on [0,1], and $w$ is a hyperparameter indicating the amplitude of perturbation. The perturbation-residual restrictions establish the connection between surrounding rays to share their geometric information. Then, we calculate the residual with the previous stage:
\begin{equation}
    \mathcal{L}_{res} = \delta\mathcal{L}_{smooth\_D} + \epsilon\mathcal{L}_{consist\_N},
\end{equation}
\begin{equation}
    \mathcal{L}_{smooth\_D} = \frac{1}{k}\sum\limits_k SmoothL1(\hat{D}_k,\hat{D}_k^{pert}),
\end{equation}
\begin{equation}
    \mathcal{L}_{consist\_N} = \frac{1}{N_{prior\_N}}\sum\limits_k SmoothL1({N}_k,\hat{N}_k^{pert}),
\end{equation}
where $SmoothL1$ is the smooth L1 loss ($beta=0.1$) defined in Equation \ref{con:smoothl1}, $N_{prior\_N}$ is the number of valid $N_k$, $\delta$ and $\epsilon$ are two hyperparameters. $\mathcal{L}_{smooth\_D}$ is a smoothing term. This term works on all sampling rays, aiming to make the depth of the sampling ray and the corresponding perturbation ray close. If $N_i$ has a valid value, then $\hat{N}_k^{pert}$ would be constrained by $\mathcal{L}_{consist\_N}$. These two optimization terms make the surface geometry in the flat region continuous and obey the normal prior. 

The final loss function is defined as
\begin{equation}
    \mathcal{L} = \mathcal{L}_{color} + \mathcal{L}_{prior} + \mathcal{L}_{res} + \mathcal{L}_{Eikonal},
\end{equation}
\begin{equation}
    \mathcal{L}_{Eikonal} = \frac{1}{nk}\sum_{n,k}(\left | \nabla f(x_{n,k})\right |-1)^2,
\end{equation}
where $\mathcal{L}_{Eikonal}$ is the Eikonal term on the sampling points to regularize the SDF.

\subsection{Mesh Extraction}
\label{sec4.4}
For each spatial position in the optimization area, we query the geometry function $f$ to obtain the corresponding TSDF value. Then, we use marching cube~\cite{lorensen1987marching} to extract the raw mesh.
There are many positions we do not need in the raw mesh, so we conduct ray-tracing to obtain the depth map for each pose and perform TSDF fusion~\cite{curless1996volumetric,newcombe2011kinectfusion} to generate our final mesh model. Ray-tracing and TSDF fusion are implemented based on Open3D~\cite{Zhou2018open3d}.

\section{Experiment}
\subsection{Experimental Setup}

\begin{figure*}[t] 
  \centering 
  \begin{minipage}[b]{\linewidth} 
  \newcommand{\myvspace}{6pt} 
  \newcommand{\widthOfFullPage}{0.135} 
  \newcommand{\widthOfMiniPage}{0.98}
  \subfloat[COLMAP]{
    \begin{minipage}[b]{\widthOfFullPage\linewidth} 
      \centering
      \includegraphics[width=\widthOfMiniPage\linewidth]{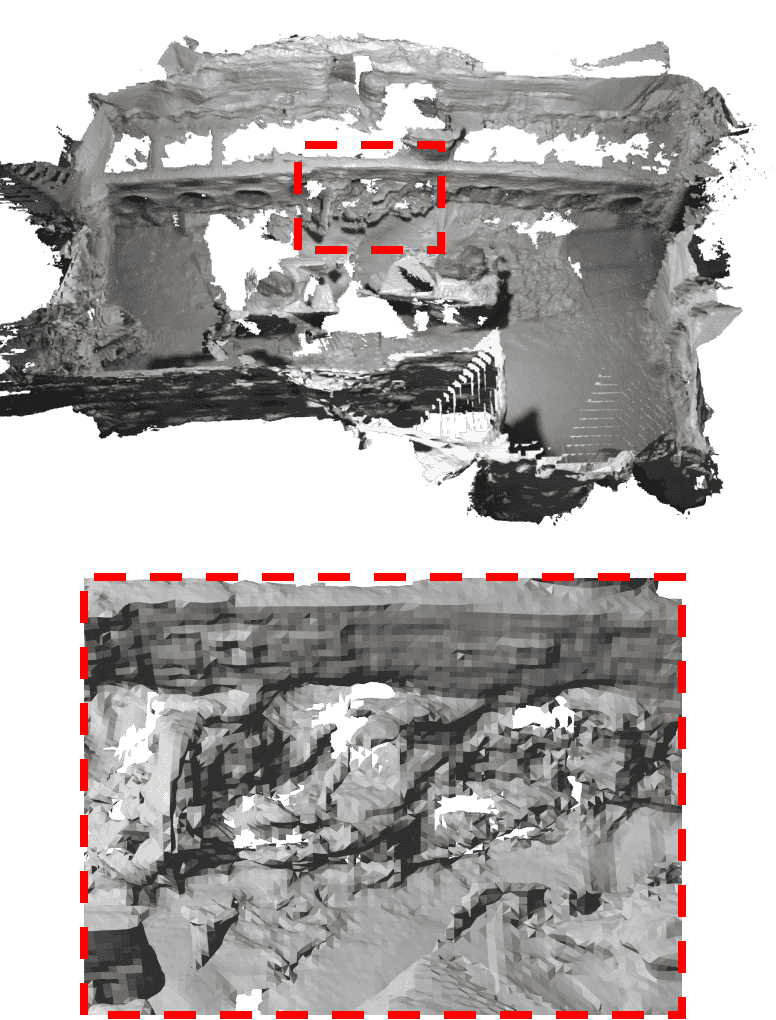}\vspace{\myvspace}
      \includegraphics[width=\widthOfMiniPage\linewidth]{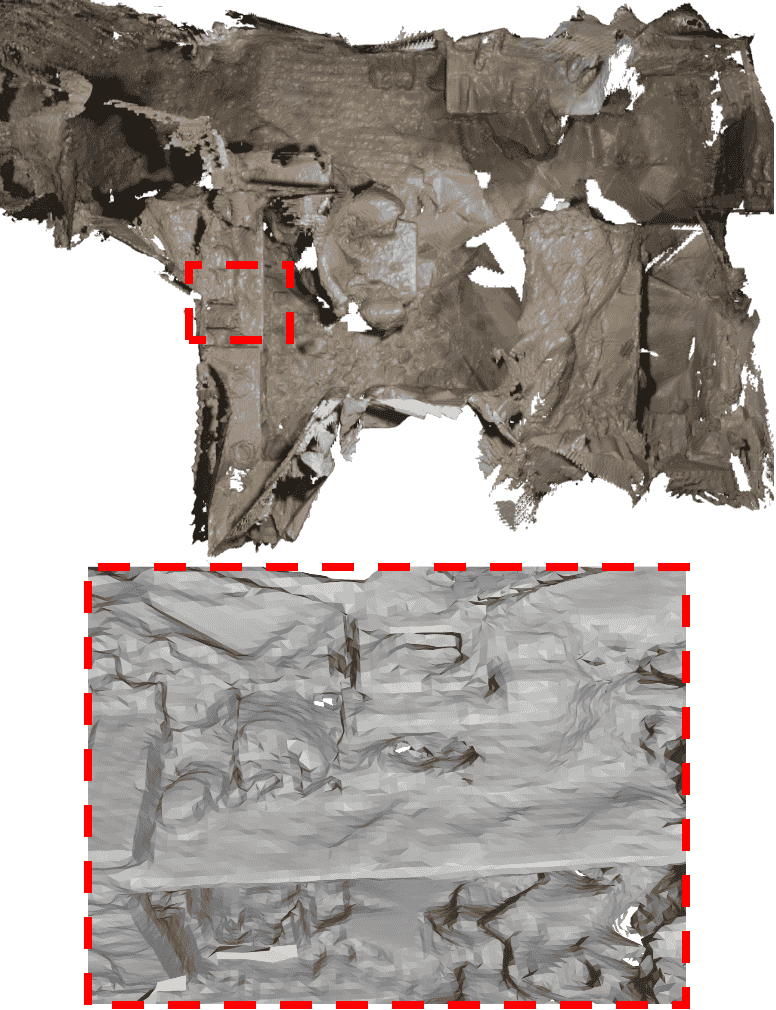}\vspace{\myvspace}
      \includegraphics[width=\widthOfMiniPage\linewidth]{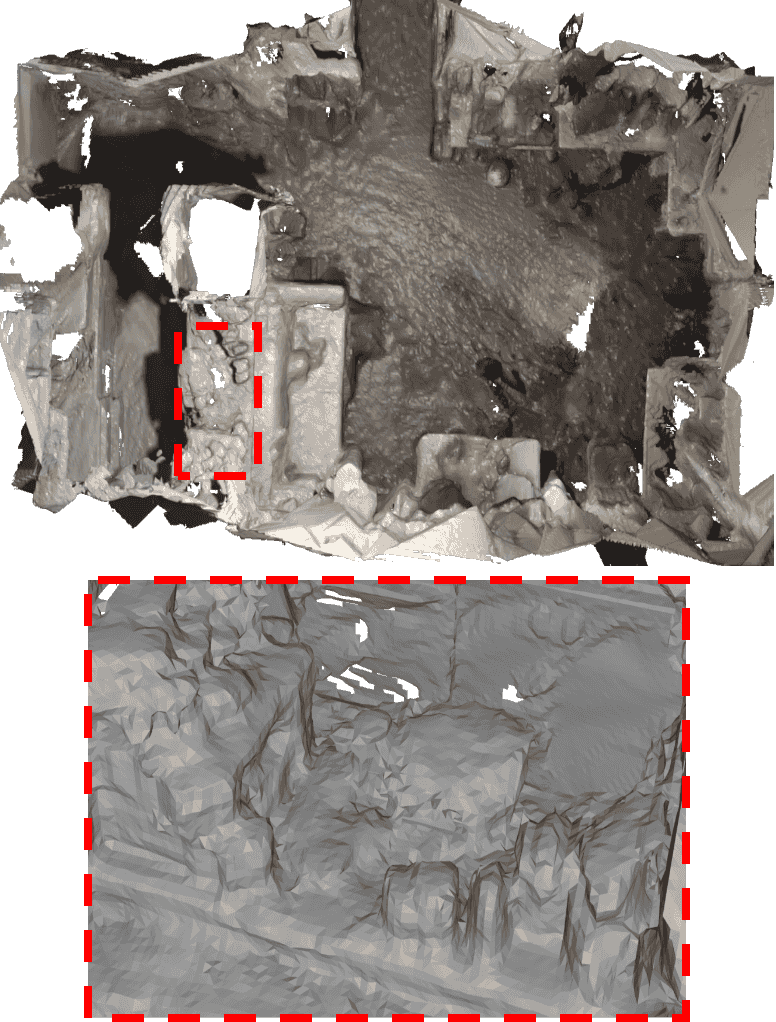}\vspace{\myvspace}
      \includegraphics[width=\widthOfMiniPage\linewidth]{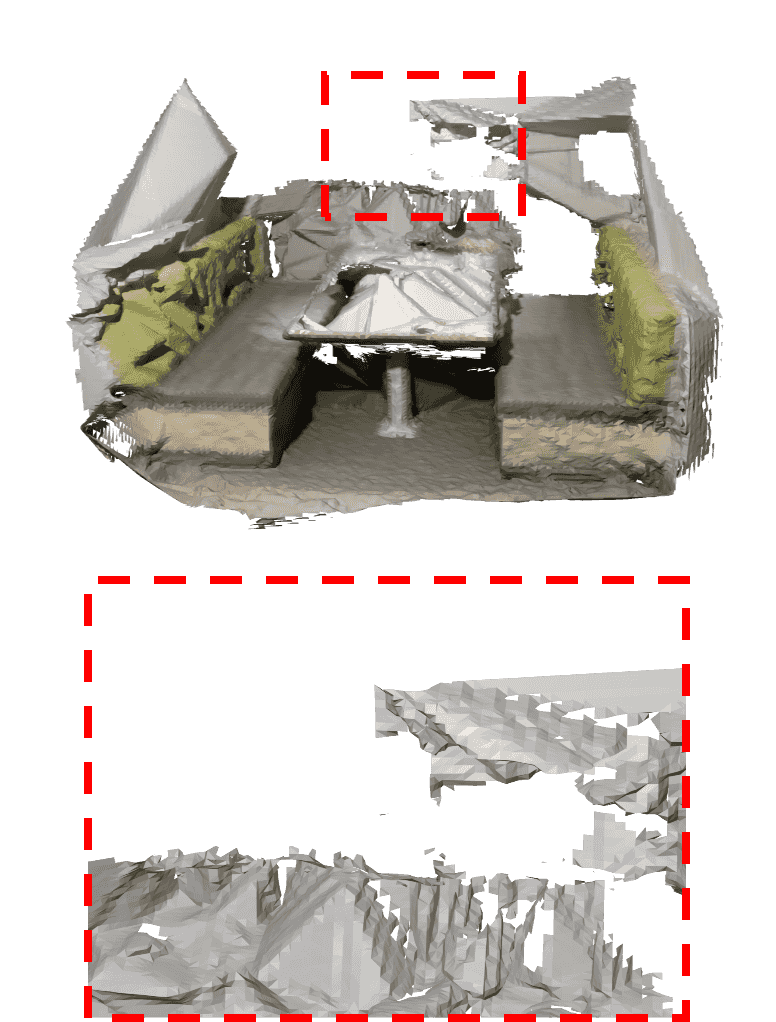}\vspace{\myvspace}
    \end{minipage}
  }
  \subfloat[Atlas]{
    \begin{minipage}[b]{\widthOfFullPage\linewidth} 
      \centering
      \includegraphics[width=\widthOfMiniPage\linewidth]{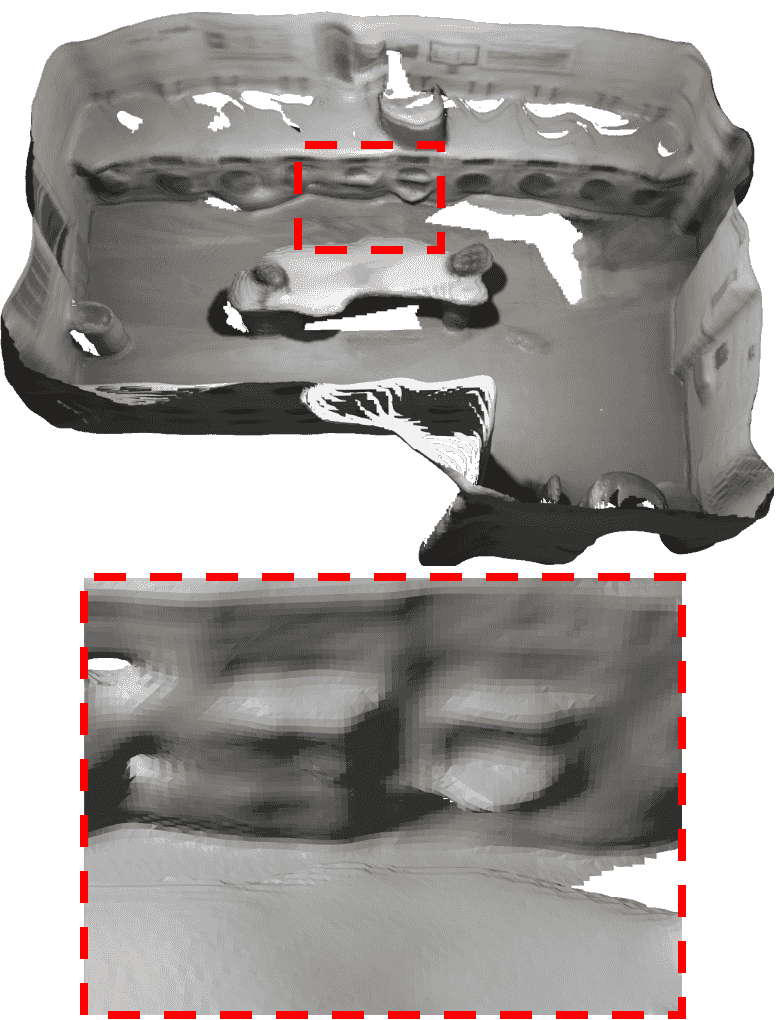}\vspace{\myvspace}
      \includegraphics[width=\widthOfMiniPage\linewidth]{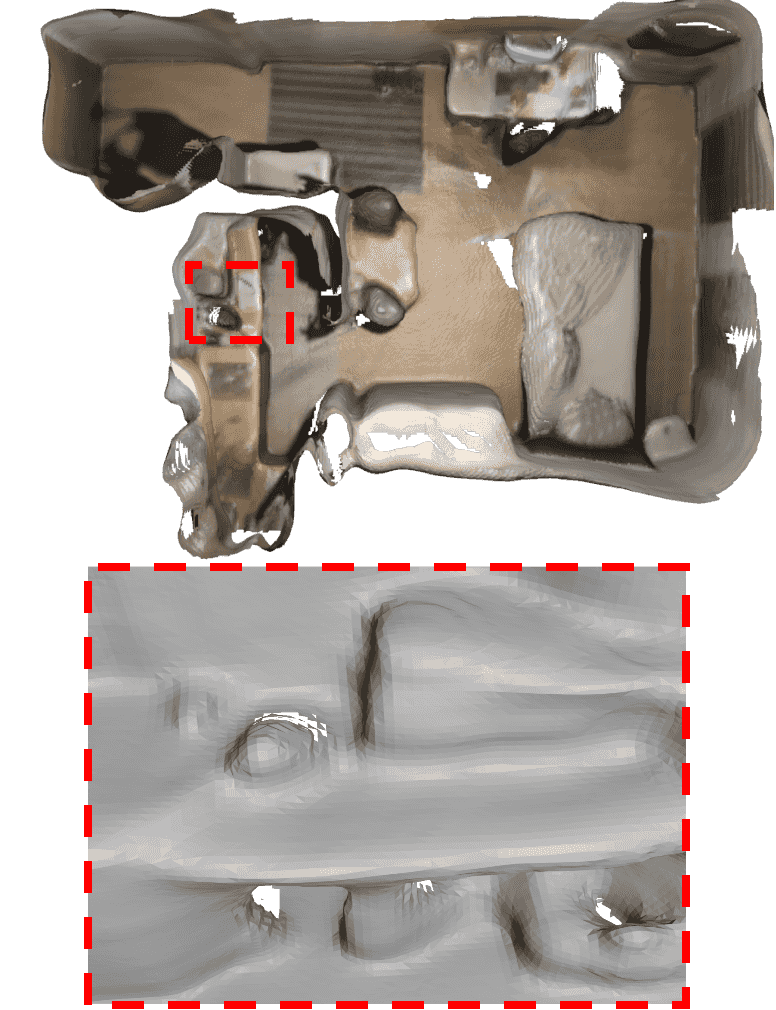}\vspace{\myvspace}
      \includegraphics[width=\widthOfMiniPage\linewidth]{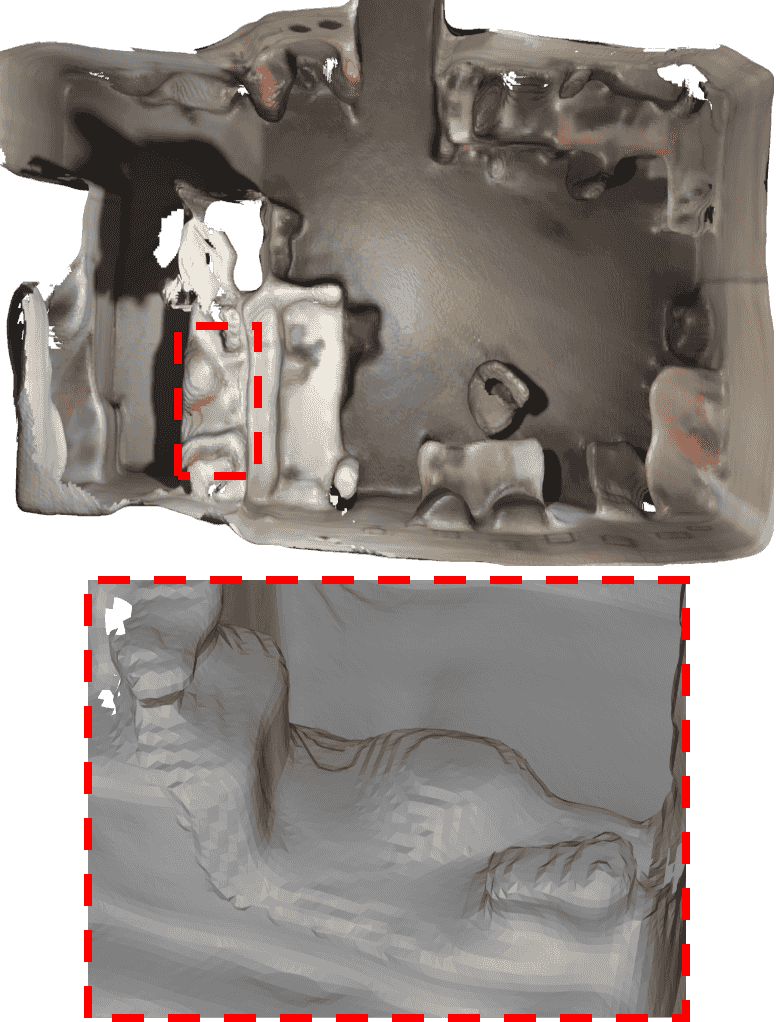}\vspace{\myvspace}
      \includegraphics[width=\widthOfMiniPage\linewidth]{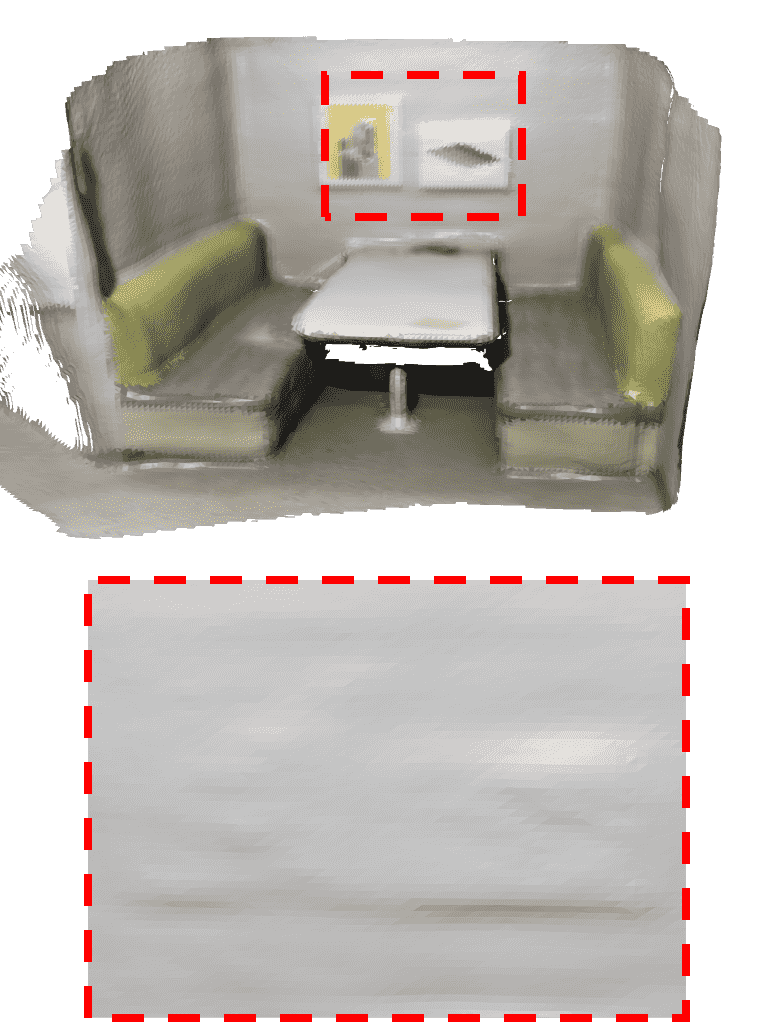}\vspace{\myvspace}
    \end{minipage}
  }
  \subfloat[NeuralRecon]{
    \begin{minipage}[b]{\widthOfFullPage\linewidth} 
      \centering
      \includegraphics[width=\widthOfMiniPage\linewidth]{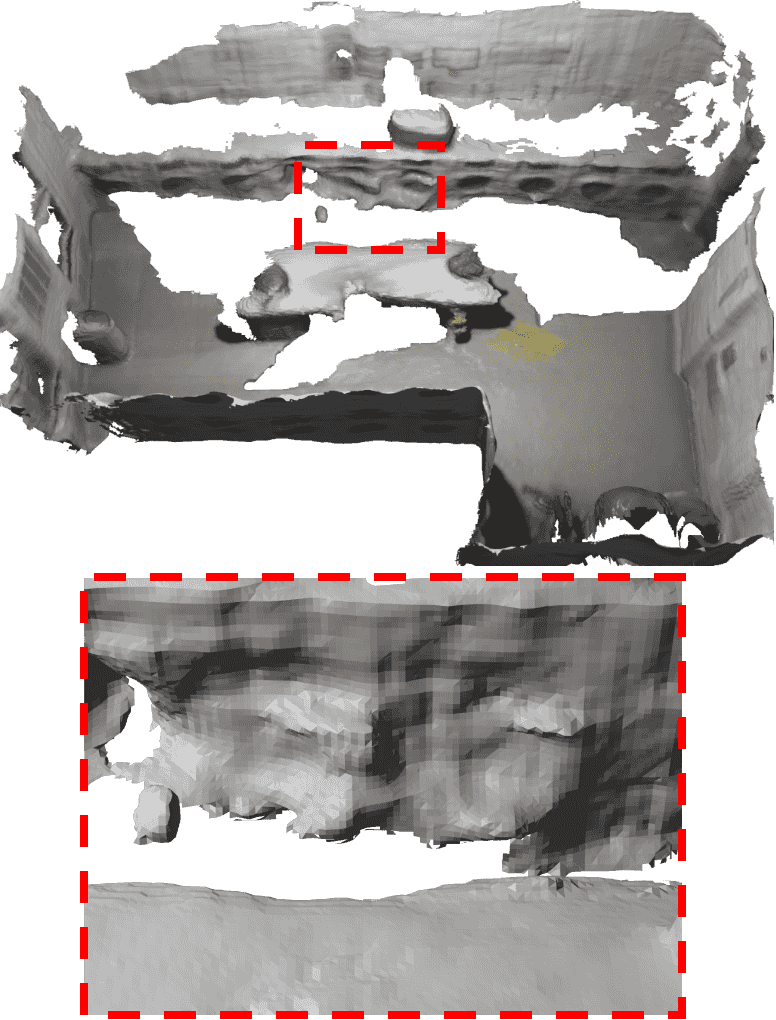}\vspace{\myvspace}
      \includegraphics[width=\widthOfMiniPage\linewidth]{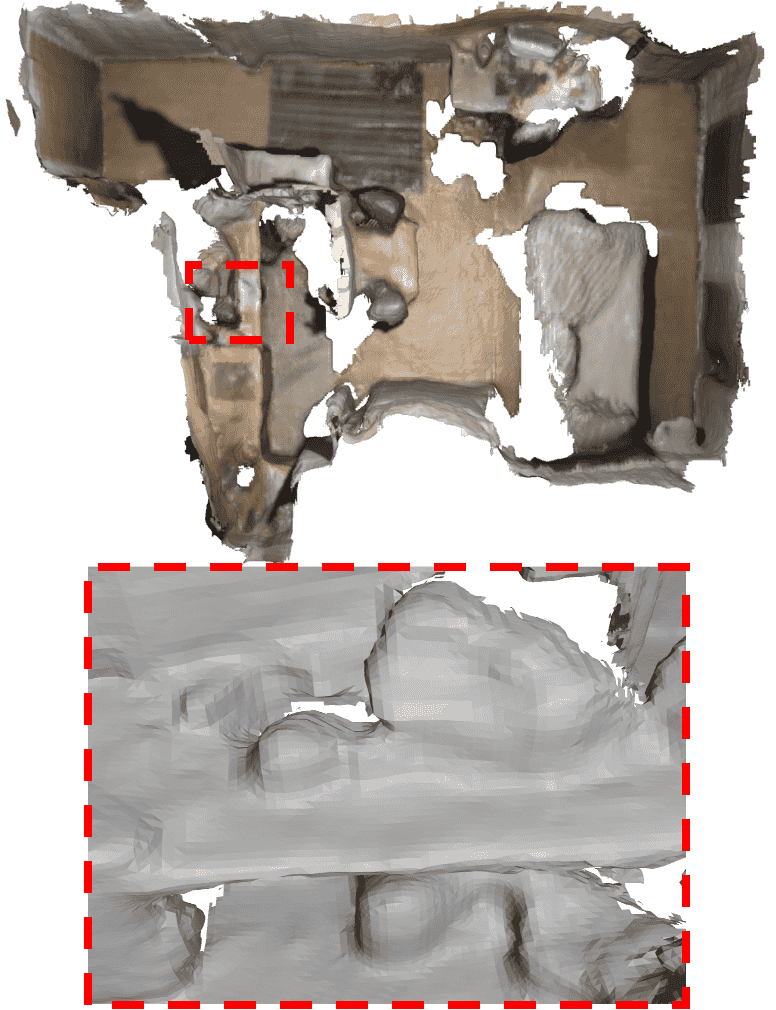}\vspace{\myvspace}
      \includegraphics[width=\widthOfMiniPage\linewidth]{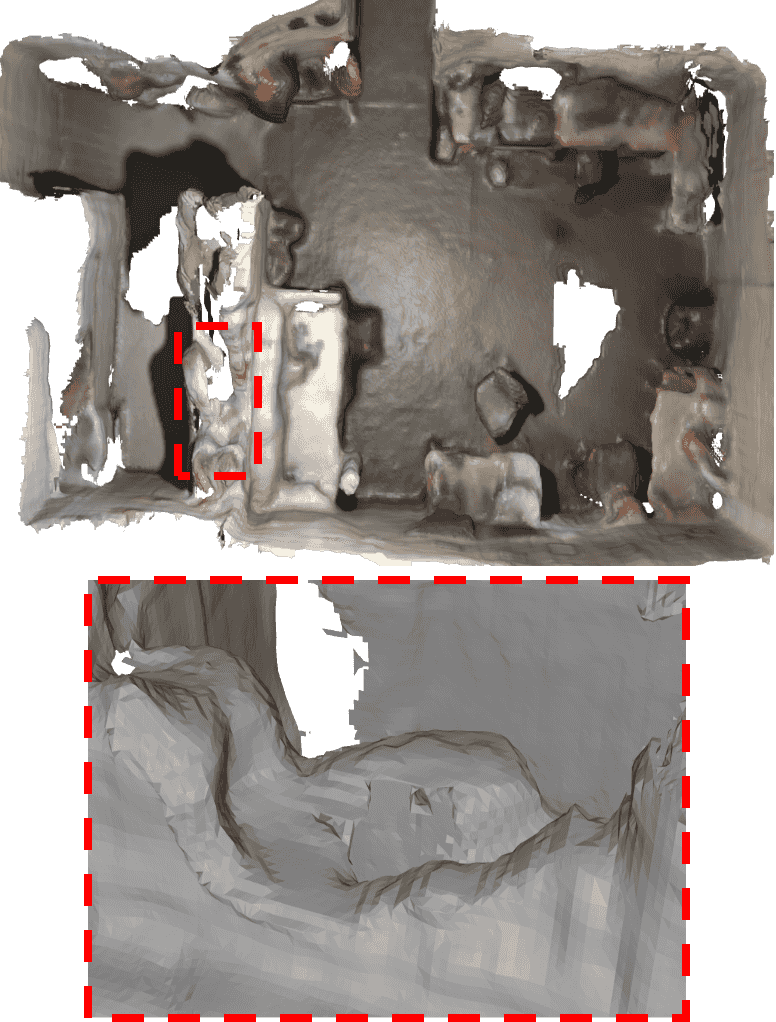}\vspace{\myvspace}
      \includegraphics[width=\widthOfMiniPage\linewidth]{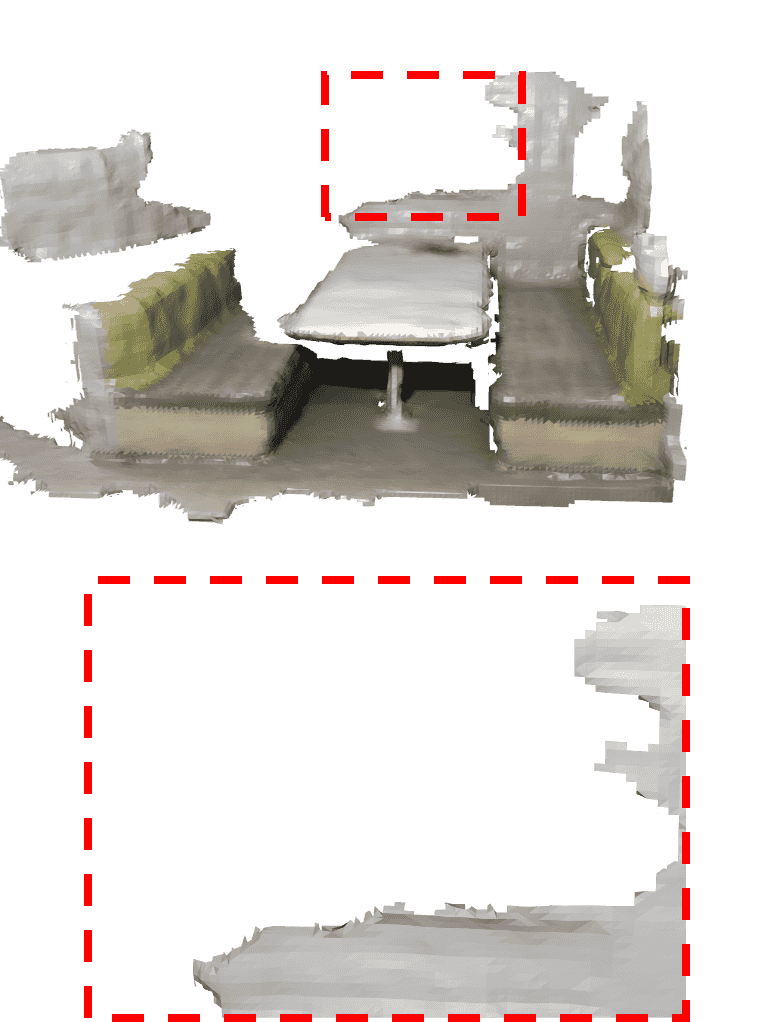}\vspace{\myvspace}
    \end{minipage}
  }
  \subfloat[3DVNet]{
    \begin{minipage}[b]{\widthOfFullPage\linewidth} 
      \centering
      \includegraphics[width=\widthOfMiniPage\linewidth]{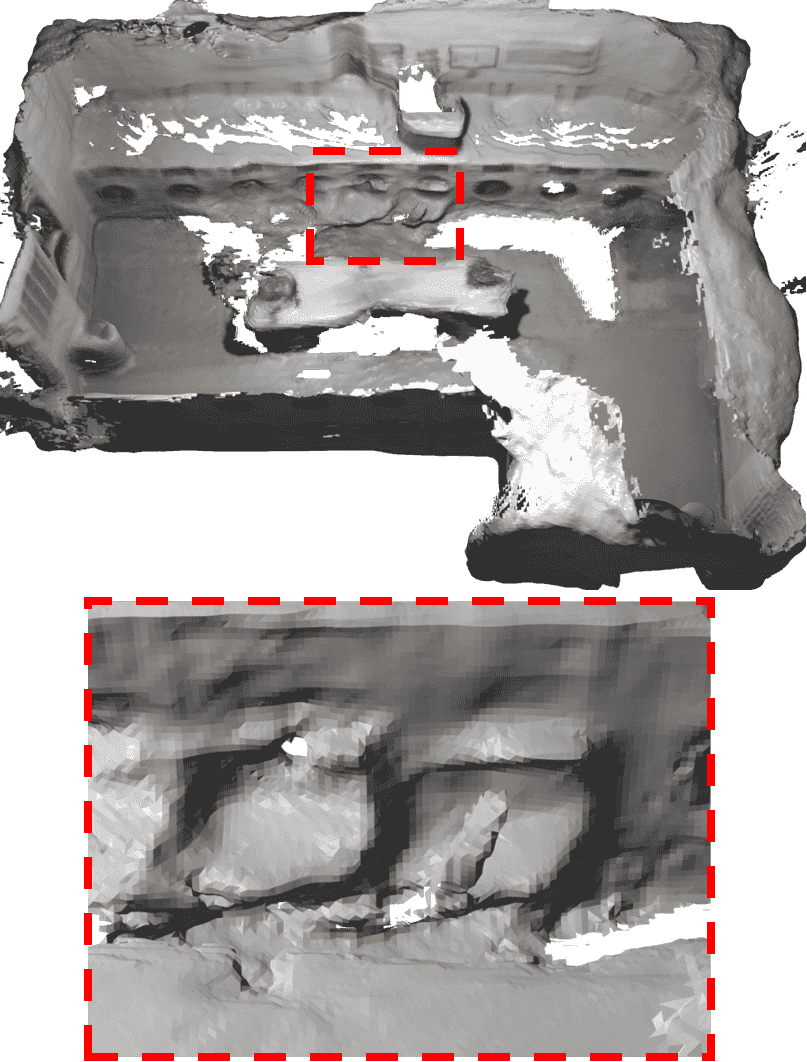}\vspace{\myvspace}
      \includegraphics[width=\widthOfMiniPage\linewidth]{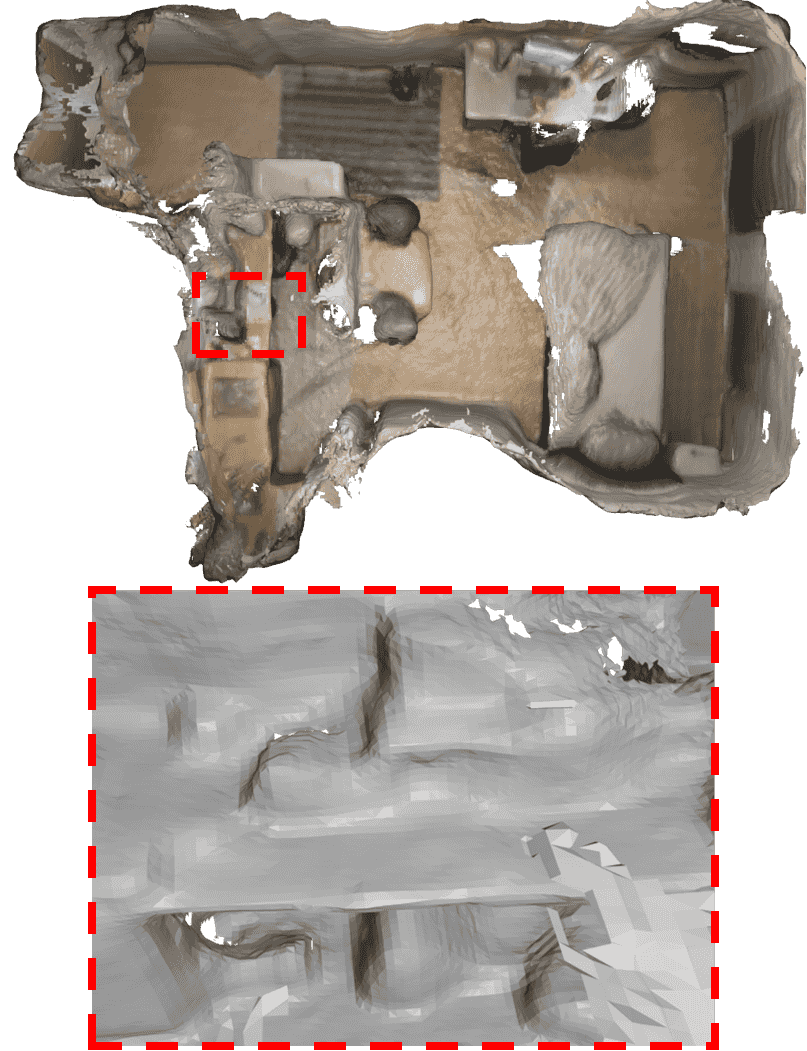}\vspace{\myvspace}
      \includegraphics[width=\widthOfMiniPage\linewidth]{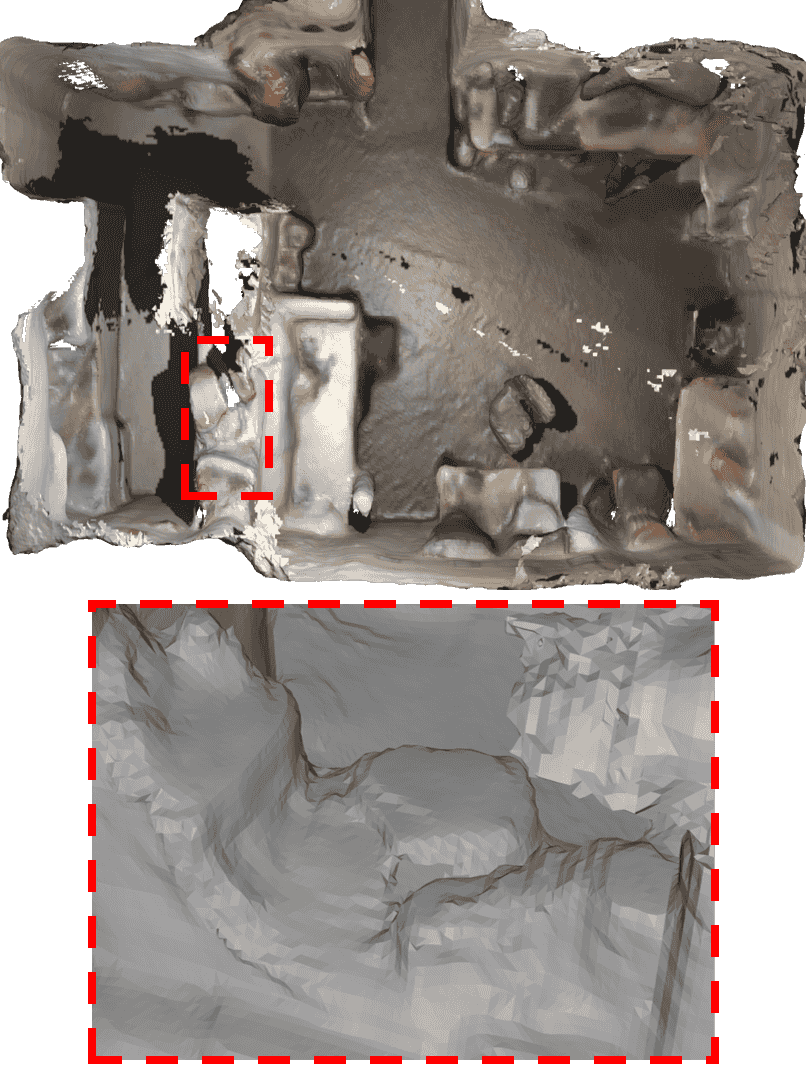}\vspace{\myvspace}
      \includegraphics[width=\widthOfMiniPage\linewidth]{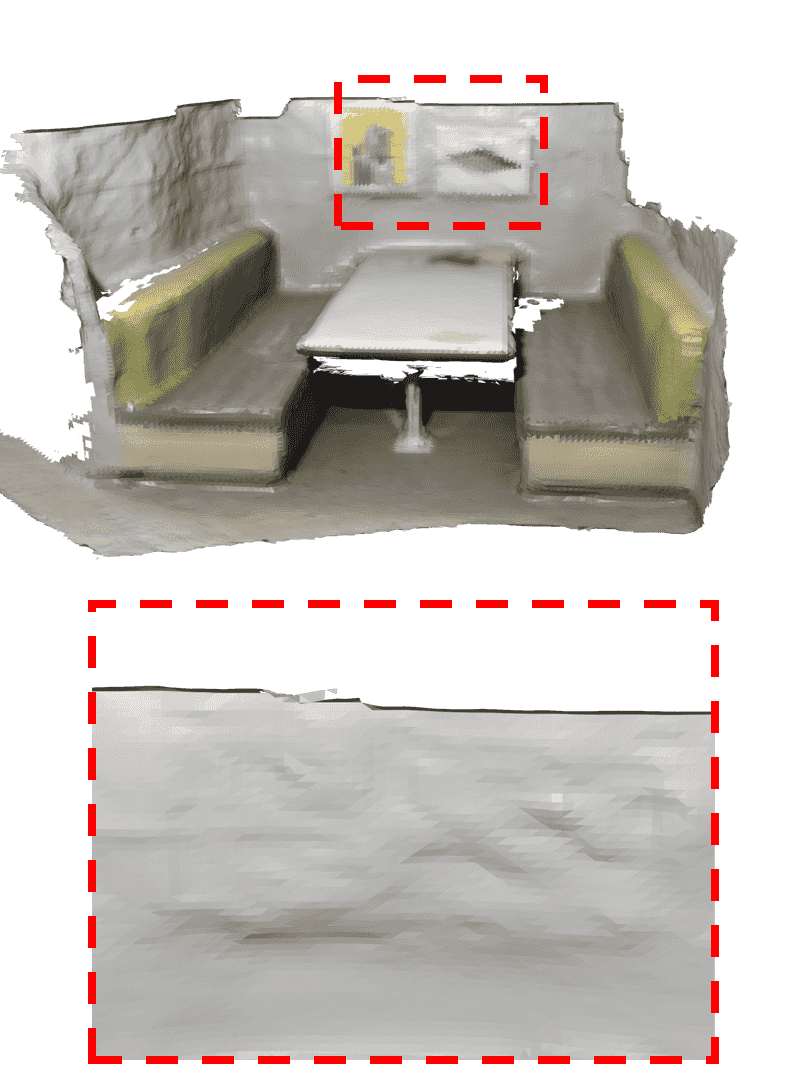}\vspace{\myvspace}
    \end{minipage}
  }
  \subfloat[ESTDepth]{
    \begin{minipage}[b]{\widthOfFullPage\linewidth} 
      \centering
      \includegraphics[width=\widthOfMiniPage\linewidth]{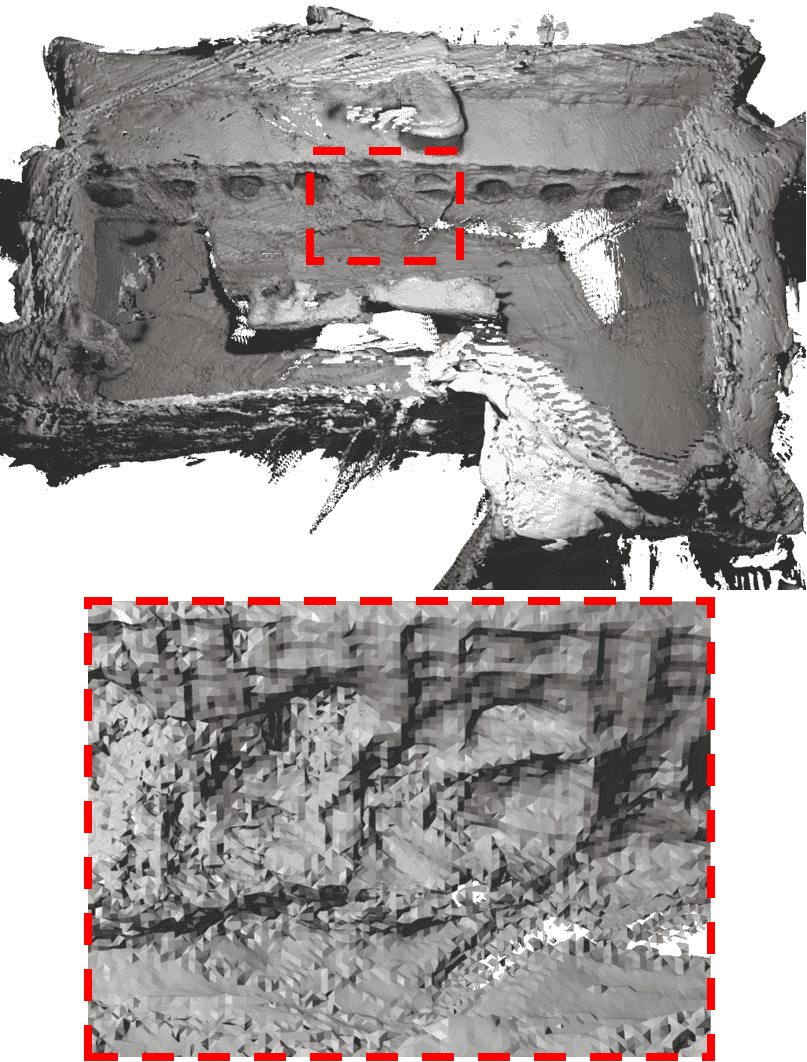}\vspace{\myvspace}
      \includegraphics[width=\widthOfMiniPage\linewidth]{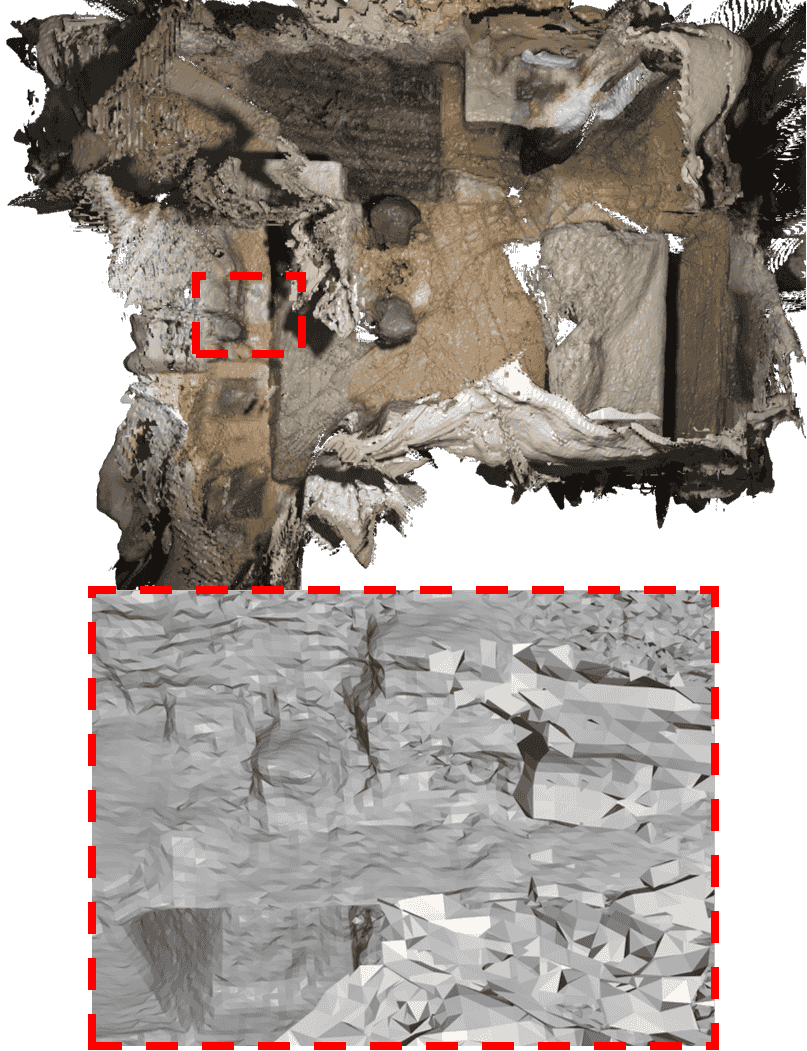}\vspace{\myvspace}
      \includegraphics[width=\widthOfMiniPage\linewidth]{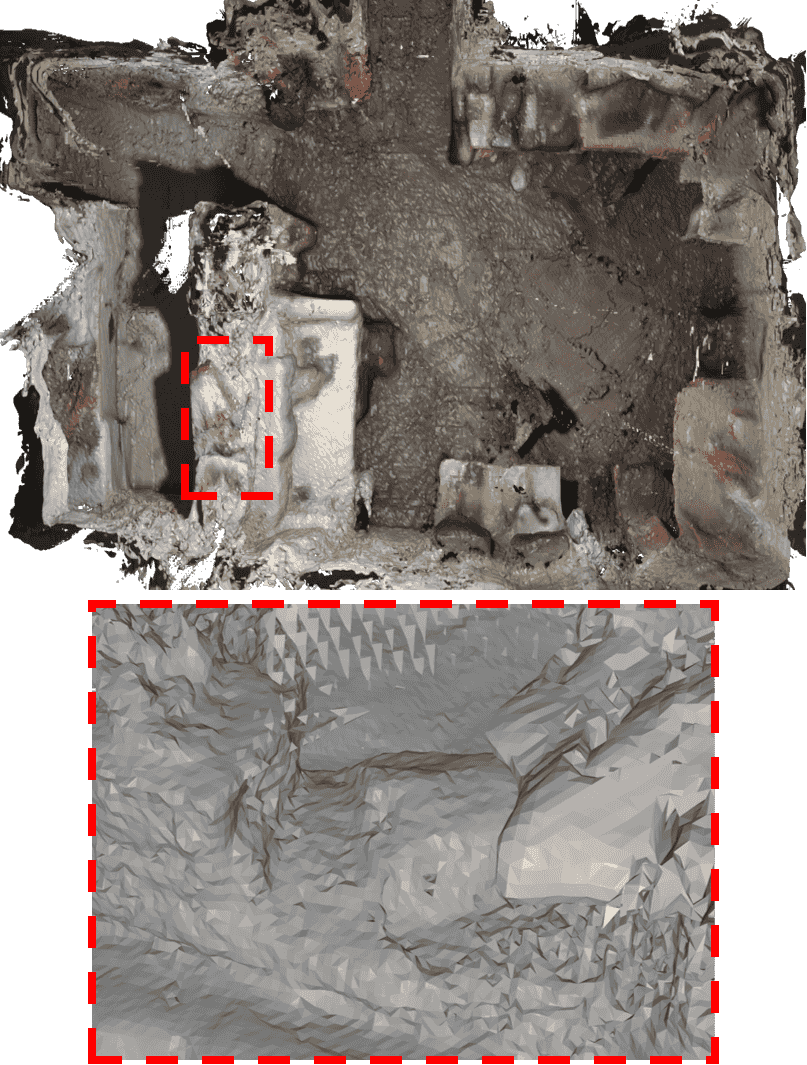}\vspace{\myvspace}
      \includegraphics[width=\widthOfMiniPage\linewidth]{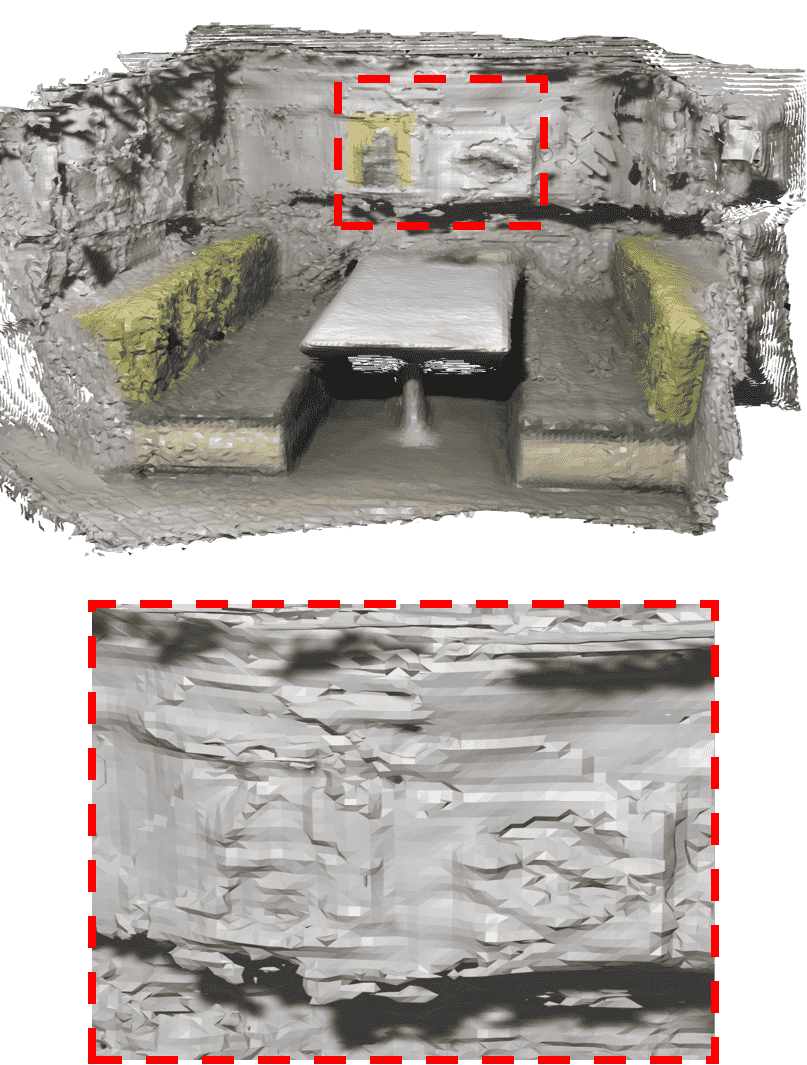}\vspace{\myvspace}
    \end{minipage}
  }
  \subfloat[Ours]{
    \begin{minipage}[b]{\widthOfFullPage\linewidth} 
      \centering
      \includegraphics[width=\widthOfMiniPage\linewidth]{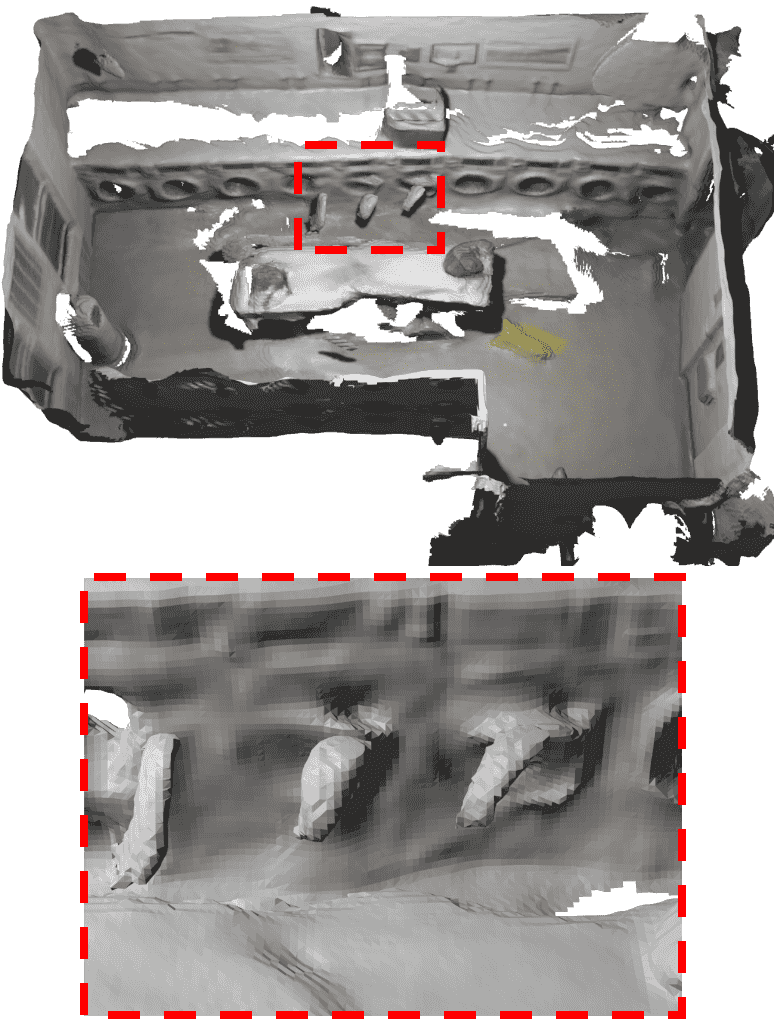}\vspace{\myvspace}
      \includegraphics[width=\widthOfMiniPage\linewidth]{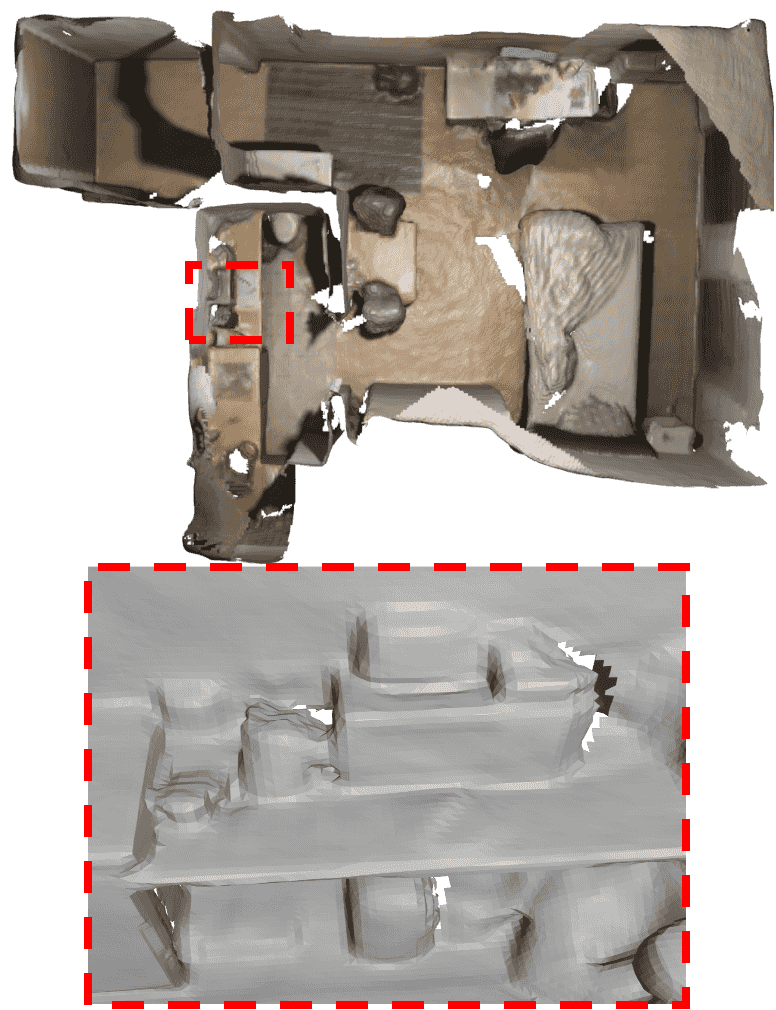}\vspace{\myvspace}
      \includegraphics[width=\widthOfMiniPage\linewidth]{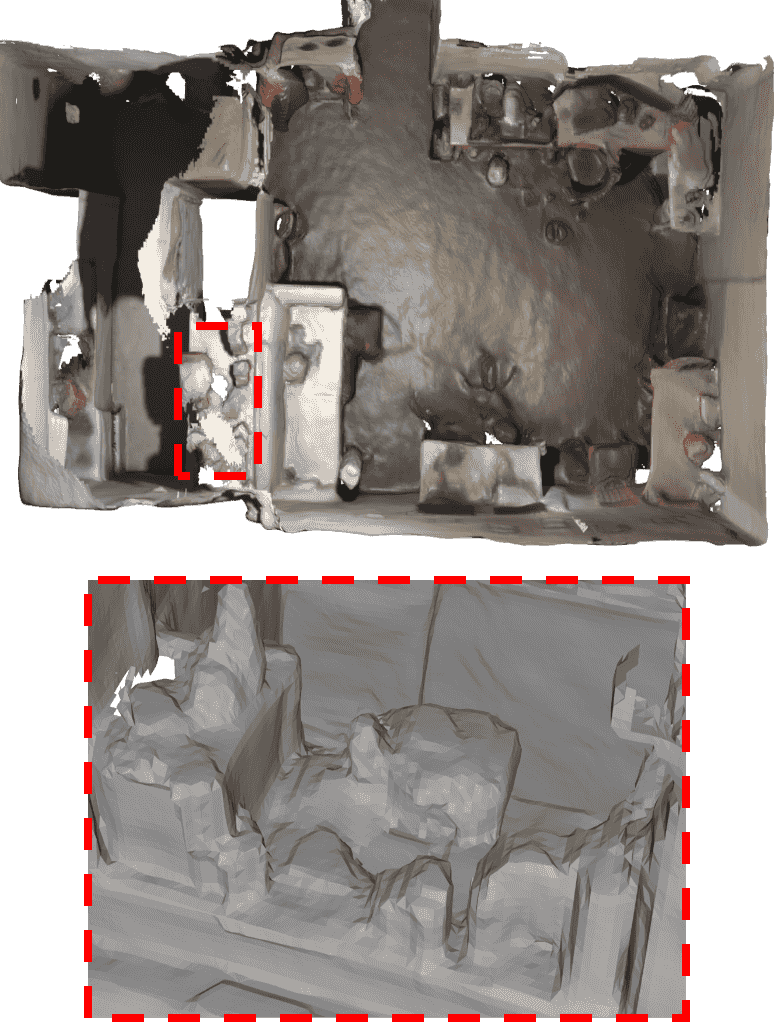}\vspace{\myvspace}
      \includegraphics[width=\widthOfMiniPage\linewidth]{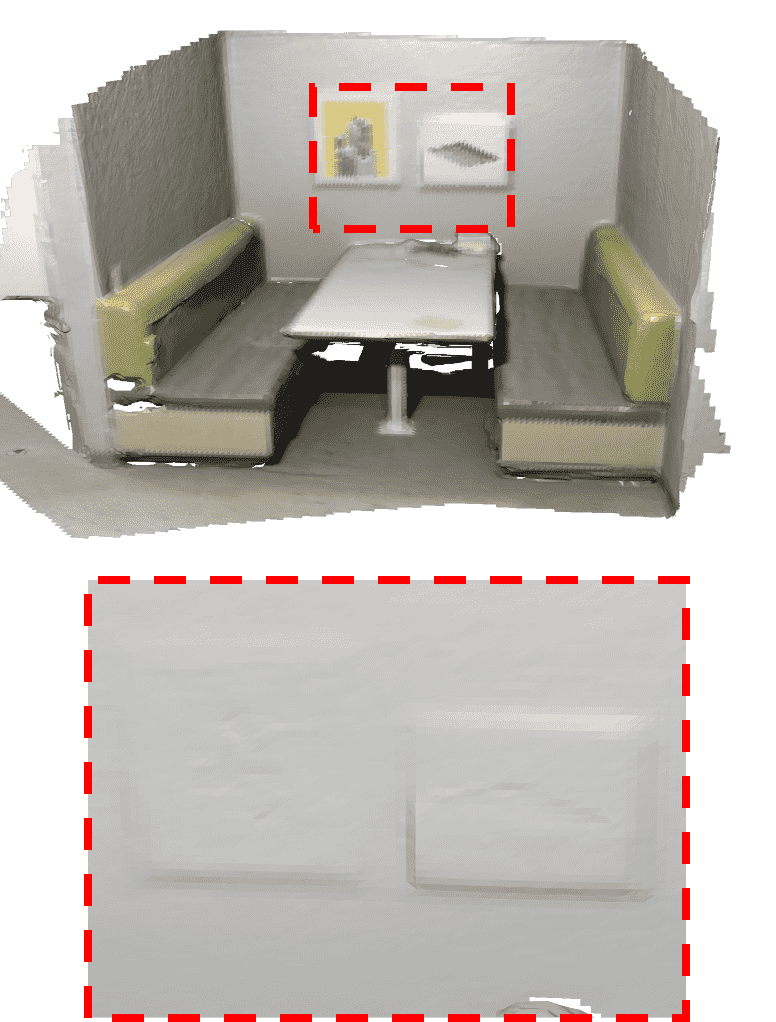}\vspace{\myvspace}
    \end{minipage}
  }
  \subfloat[GT]{
    \begin{minipage}[b]{\widthOfFullPage\linewidth} 
      \centering
      \includegraphics[width=\widthOfMiniPage\linewidth]{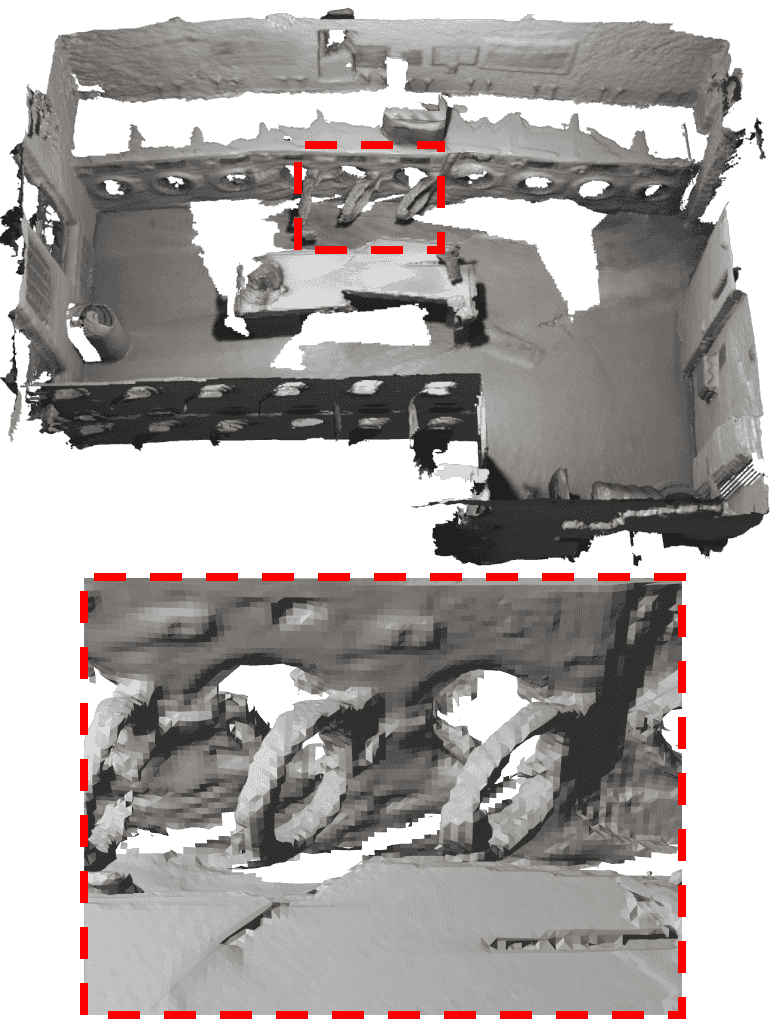}\vspace{\myvspace}
      \includegraphics[width=\widthOfMiniPage\linewidth]{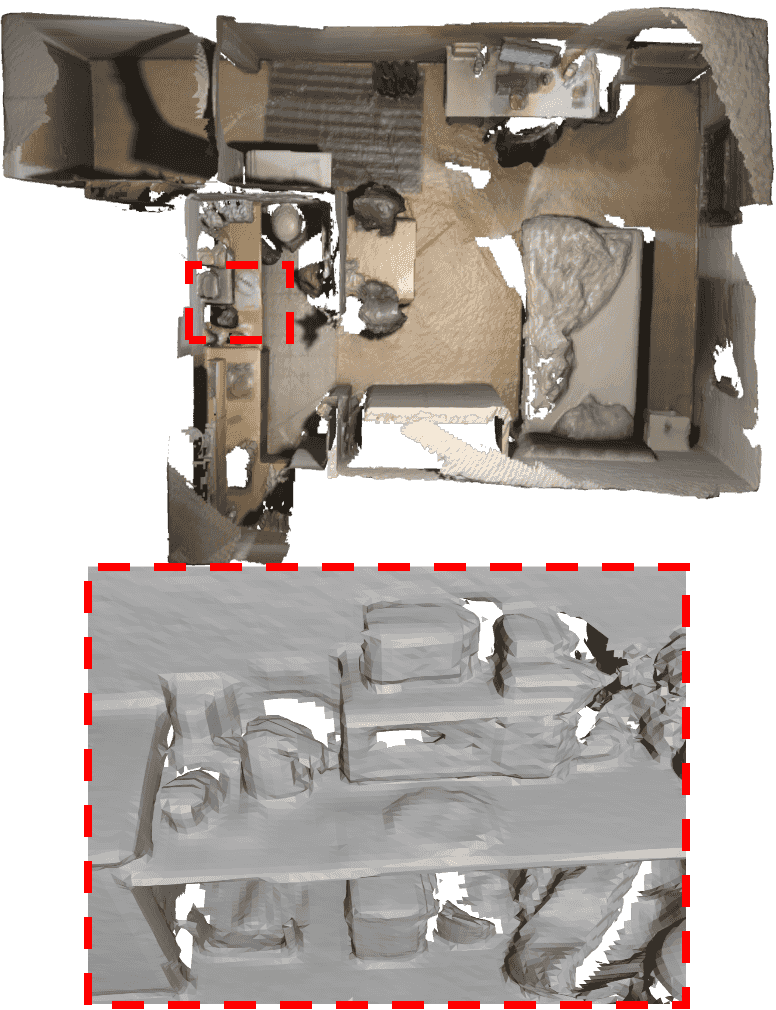}\vspace{\myvspace}
      \includegraphics[width=\widthOfMiniPage\linewidth]{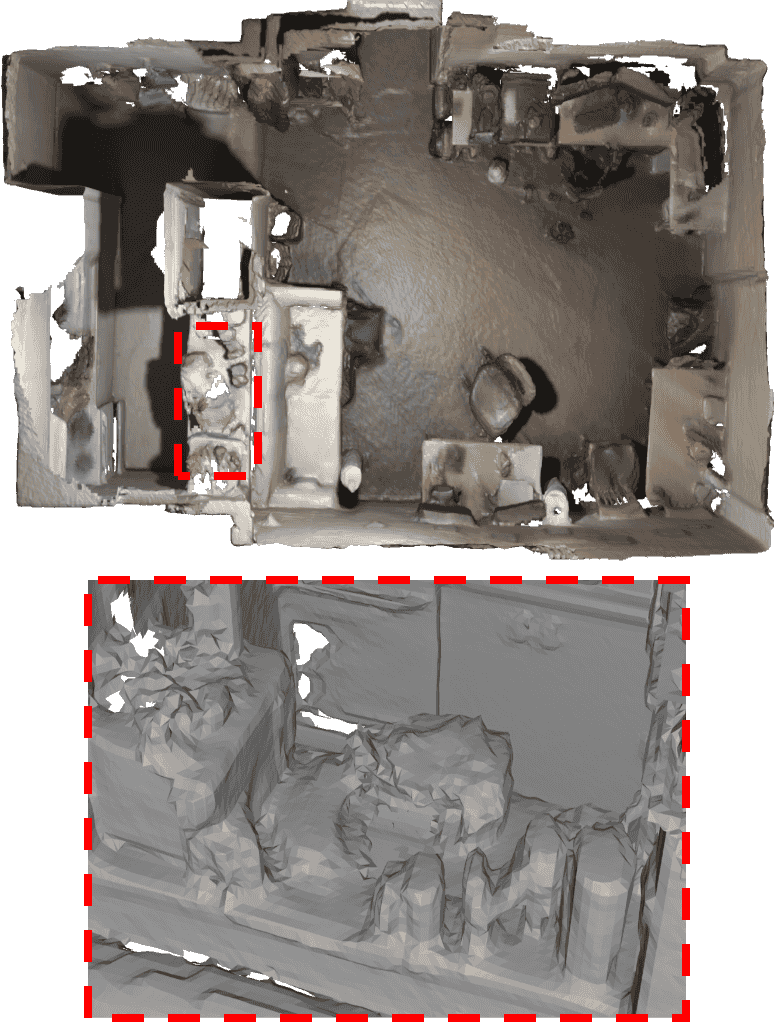}\vspace{\myvspace}
      \includegraphics[width=\widthOfMiniPage\linewidth]{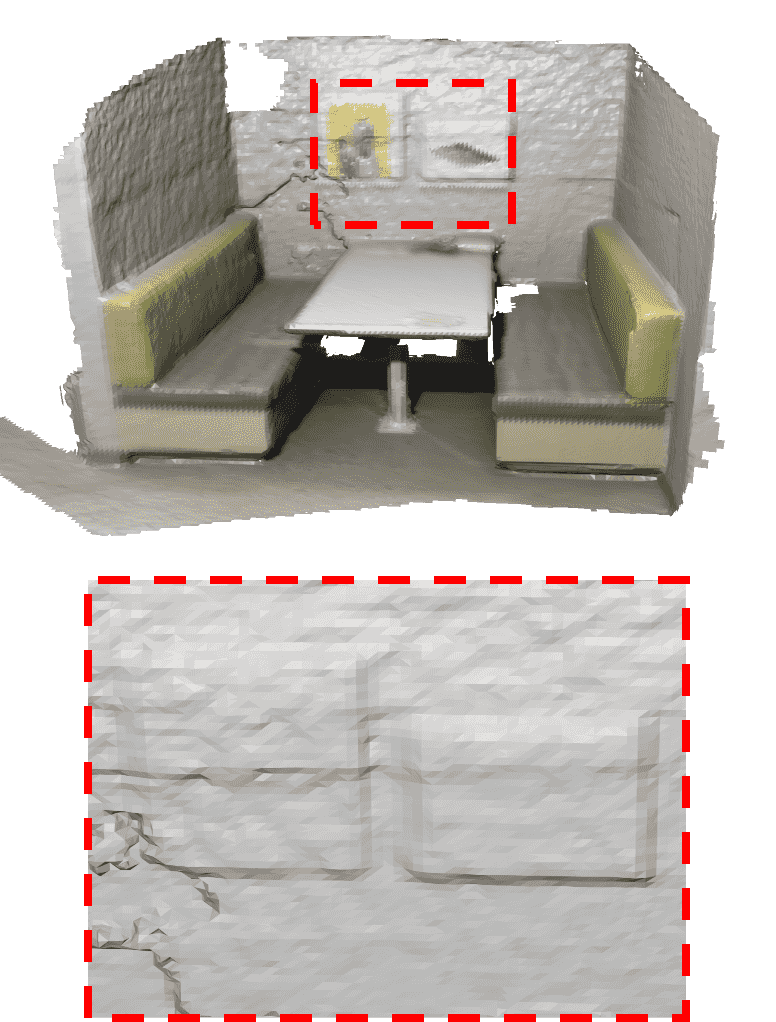}\vspace{\myvspace}
    \end{minipage}
  }
  \end{minipage}
  \caption{Visualization results on ScanNet. We compared our reconstruction results with those of COLMAP ~\cite{schonberger2016structure} and other advanced learning-based indoor scene reconstruction methods~\cite{murez2020atlas,sun2021neuralrecon,rich20213dvnet,long2021multi}. The reconstruction results of NeuralRoom exhibit similar scene integrity in visual perception to those of Atlas. The details are preserved better than other algorithms. The quantitative evaluation is shown in Table~\ref{quantitative_evaluation}.}
  \label{visual_comparisons}
\end{figure*}

\paragraph{Dataset.}
We evaluate our approach and baseline methods on the indoor dataset, ScanNet (V2)~\cite{dai2017scannet}. We randomly select 8 test scenes from the intersection of the test set of Framenet~\cite{huang2019framenet} and ScanNet~\cite{dai2017scannet}. The~\cite{bae2021estimating} network has been pretrained on the training set of~\cite{huang2019framenet} for evaluation. We take one photo out of approximately every ten adjacent photos with an image resolution of $1296\times968$. 



\paragraph{Implementation Details.}
The geometry function $f$ is modeled by an MLP, which consists of 8 hidden layers with a hidden size of 256. The color function $c$ is modeled by an MLP, which consists of 4 hidden layers with a size of 256. Positional encoding, initialization of the implicit neural representation, and coarse to fine sampling methods are similar to the method of~\cite{wang2021neus}. The numbers of coarse and fine sampling points for each ray are 64 and 64, respectively. We sample 512 rays per batch and train NeuralRoom for 200k iterations on a single NVIDIA RTX2080Ti GPU. The hyperparameters used in the experiment are as follows: $e=1.3, w=2.4,\gamma=0.001,\delta=0.001,$ and $\epsilon=0.001$.

\paragraph{Baselines.}
We do not compare our method with rendering-based reconstruction methods~\cite{wang2021neus,oechsle2021unisurf,volsdf2021} since these reconstruction methods often fails. We compare our method with the following baseline methods, NeuralRecon~\cite{sun2021neuralrecon} and Atlas~\cite{murez2020atlas}, two volumetric multiview indoor scene reconstruction methods that directly extract 3D surface from feature volume; COLMAP~\cite{schonberger2016structure} and ACMP~\cite{Xu2020ACMP}, two traditional PatchMatch-based MVS methods; 3DVNet~\cite{rich20213dvnet}, a learning-based multiview stereo method that combines the advantages of depth-based and volumetric multiview stereo approaches; and ESTDepth~\cite{long2021multi}, a learning-based multiview depth estimation method.

\paragraph{Evaluation Protocols.}
We choose a mesh as the 3D geometry representation for qualitative comparison and quantitative evaluation. For COLMAP, we perform Delaunay triangulation to form a mesh from a point cloud. For our method, Atlas, NeuralRecon and COLMAP with Delaunay triangulation, in which there are a large number of nonobservation areas. Therefore, we use ray tracing provided by Open3D~\cite{Zhou2018open3d} on the reconstructed model to obtain the depth map of each view. Then we use TSDF fusion~\cite{curless1996volumetric,newcombe2011kinectfusion} to obtain a trimmed 3D mesh. For ACMP, ESTDepth and 3DVNet, we use TSDF fusion to obtain a mesh. The voxel length is $0.02 m$, and the SDF truncation value is $0.12 m$. For quantitative evaluation, we use a regular voxel grid to create a uniformly downsampled point cloud from the input mesh. The side length of the voxel grid is $0.005 m$. The threshold of precision and recall is $0.05 m$.

\subsection{Geometry Prior Statistics}

We perform the quantitative evaluation for the prior on all test scenes. The distance prior is converted from the corresponding MVS depth map, so we evaluate the depth map instead. The depth map and normal map are evaluated in Table~\ref{tab_depthmap} and Table~\ref{tab_normal}. The evaluation of the normal map filtered by uncertainty has a \_50 extension. After filtering, we obtain highly accurate normal estimation results, for which the mean angle error is $8.183^{\circ}$.

\begin{table}[h]%
\caption{Quantitative evaluation of the depth map acquired by COLMAP.}
\label{tab_depthmap}
\begin{minipage}{\columnwidth}
\begin{center}
\begin{tabular}{ccccc}
  \toprule
  Comp & Abs Diff & Abs Rel & Sq Rel & RMSE\\ \midrule
   0.143 & 0.093 & 0.042 & 0.013 & 0.158\\
  \bottomrule
\end{tabular}
\end{center}
\bigskip\centering
\end{minipage}
\end{table}%

\begin{table}%
\caption{Quantitative evaluation of the acquired normal map. The filtered normal maps are with a $\_50$ extension.}
\label{tab_normal}
\begin{minipage}{\columnwidth}
\begin{center}
\begin{tabular}{cccccc}
  \toprule
  Mean & Median & RMSE & Mean\_50 & Median\_50 & RMSE\_50\\ \midrule
   14.691 & 7.329 & 23.224 & 8.183 & 4.931 & 13.030\\
  \bottomrule
\end{tabular}
\end{center}
\bigskip\centering
\end{minipage}
\end{table}%

\subsection{Evaluation Results}
\begin{figure}[!htb]
  \centering
  \includegraphics[width=\linewidth]{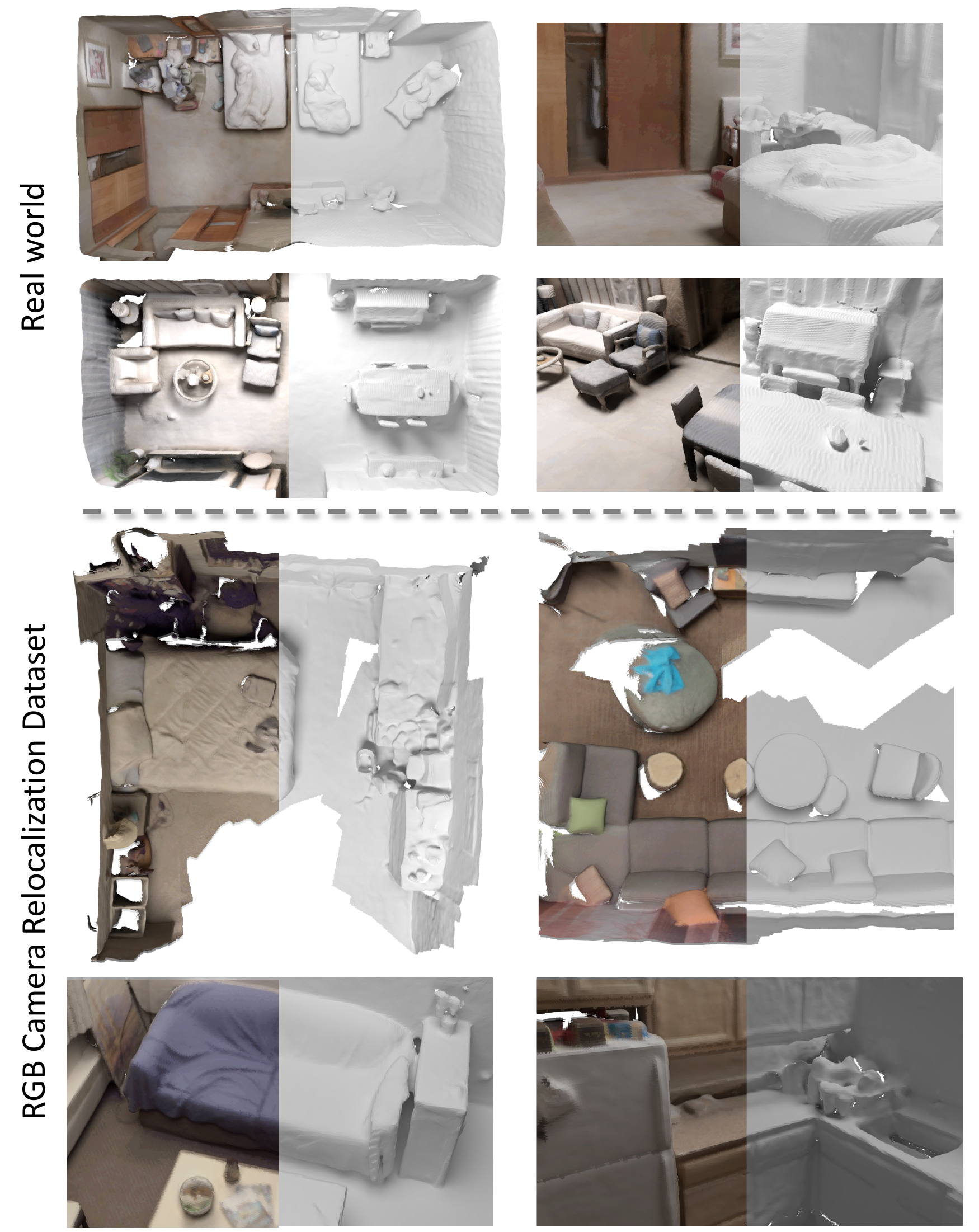}
  \caption{We applied NeuralRoom to real-world scenes and RGB Camera Relocalization Dataset~\cite{valentin2016learning}.}
 \label{real_world}
\end{figure}


We provide qualitative and quantitative comparisons on the ScanNet dataset to evaluate the performance of our system. Our method achieves state-of-the-art both quantitatively and qualitatively. Since all algorithms will produce some additional reconstruction areas, for a fair comparison, we perform ray tracing with ground truth camera parameters to get the corresponding depth, and then perform TSDF fusion to acquire cleaned 3D reconstruction.
\paragraph{Visual comparison.} Figure~\ref{visual_comparisons} shows our full scene and detailed visualization results compared with those of different reconstruction methods, including COLMAP~\cite{schonberger2016structure}, Atlas~\cite{murez2020atlas},  NeuralRecon~\cite{sun2021neuralrecon}, 3DVNet~\cite{rich20213dvnet}, ESTDepth~\cite{long2021multi} and ground truth~\cite{dai2017scannet}. 

COLMAP~\cite{schonberger2016structure} produces accurate geometry in rich textured areas, but it cannot handle texture-less areas. The details are preserved to some extent. The cross-view consistency of the ESTDepth~\cite{long2021multi} is not good, so the reconstruction results are not ideal. Atlas~\cite{murez2020atlas} and NeuralRecon~\cite{sun2021neuralrecon} are end-to-end 3D reconstruction methods that directly regress a TSDF from calibrated images. Atlas usually obtains a coarse but continuous reconstruction. The details of the scene are difficult to reconstruct by Atlas. The NeuralRecon can obtain better details but sacrifice completeness. 3DVNet~\cite{rich20213dvnet} is a learning-based multiview stereo method, it can produce better visual effects than other algorithms. However, the output depth map resolution of 3DVNet is low, which results in the reconstruction results with less precise details.

Compared with other algorithms, our reconstruction results have higher reconstruction integrity and finer details. The visual effect of some reconstructed areas is better than that of the ground truth. In addition, we tested our algorithm on real-world scenes and RGB Camera Relocalization Dataset~\cite{valentin2016learning}, Figure \ref{real_world} shows the reconstruction results. We use mobile phones to take high-resolution photos of two real-world scenes. The resolution of photos is $4608\times3456$. First, we use COLMAP~\cite{schonberger2016structure} to calculate camera parameters, obtain dense depth maps and define the bounding box. Next, we resize images and depth maps to $1296\times968$, then feed the images to the normal estimation network~\cite{bae2021estimating} to obtain the normal prior. The normal estimation network was only pretrained on the ScanNet~\cite{dai2017scannet} dataset without fine-tuning. The camera intrinsics are different from ScanNet, which may influence the accuracy of the normal estimation. Finally, we use NeuralRoom to reconstruct the scene. The results show our algorithm's robustness.


\paragraph{Quantitative comparison.} Table~\ref{quantitative_evaluation} reports the summary of 3D geometry metrics for different methods. COLMAP~\cite{schonberger2016structure} and ACMP~\cite{Xu2020ACMP} are traditional MVS methods, ESTDepth \cite{long2021multi} is a multiview depth estimation method, 3DVNet~\cite{rich20213dvnet} is a learning-based multiview stereo method, Atlas~\cite{murez2020atlas} and NeuralRecon~\cite{sun2021neuralrecon} are learning-based reconstruction methods. Since rendering-based reconstruction methods have a high probability that the satisfactory reconstruction of indoor scenes cannot be acquired through input images (Figure~\ref{shape-radiance-ambiguity}), we only evaluate the methods mentioned above.

The ACMP reconstructs the indoor scene with the plane hypothesis. The reconstruction always has many outliers, resulting in low accuracy and precision scores. Our method has a relatively balanced performance in accuracy and completeness. The overall performance is much better than that of other different types of methods. We believe that the improvements come from the following aspects:

\begin{table}[h]
\centering
\caption{Quantitative evaluation of reconstruction with existing methods on the ScanNet dataset. We report the average results for eight scenes from the test set.}
\label{quantitative_evaluation}
\scalebox{0.9}{
\begin{tabular}{lcccccc}
  \toprule
  Method & Prec$\uparrow$ & Recall$\uparrow$ & F-score$\uparrow$ & Acc$\downarrow$ & Comp$\downarrow$ & Overall$\downarrow$\\
  \midrule
   COLMAP & 45.136 & 44.510 & 44.678 & 0.108 & 0.136 & 0.122\\
   ACMP & 35.978 & \uline{70.691} & 47.622 & 0.152 & \uline{0.047} & 0.100\\
   ESTDepth & 38.217 & 50.992 & 43.589 & 0.144 & 0.075 & 0.110\\
   3DVNet & 64.961 & 64.562 & 64.665 & 0.071 & 0.061 & 0.066\\
   Atlas & 67.957 & 57.747 & 61.871 & \uline{0.050} & 0.090 & 0.070\\
   NeuralRecon & 63.851 & 47.401 & 54.208 & 0.054 & 0.128 & 0.091\\
   Ours & \uline{68.347} & 65.298 & \uline{66.756} & 0.051 & 0.058 & \uline{0.055}\\
  \bottomrule
\end{tabular}
}
\end{table}

\begin{figure*}[ht]
  \centering 
  \newcommand{\myvspace}{1pt} 
  \newcommand{\widthOfFullPage}{0.25} 
  \newcommand{\widthOfMiniPage}{0.90}
  \subfloat[Base]{
    \begin{minipage}[b]{\widthOfFullPage\linewidth}
      \centering
      \includegraphics[width=\widthOfMiniPage\linewidth]{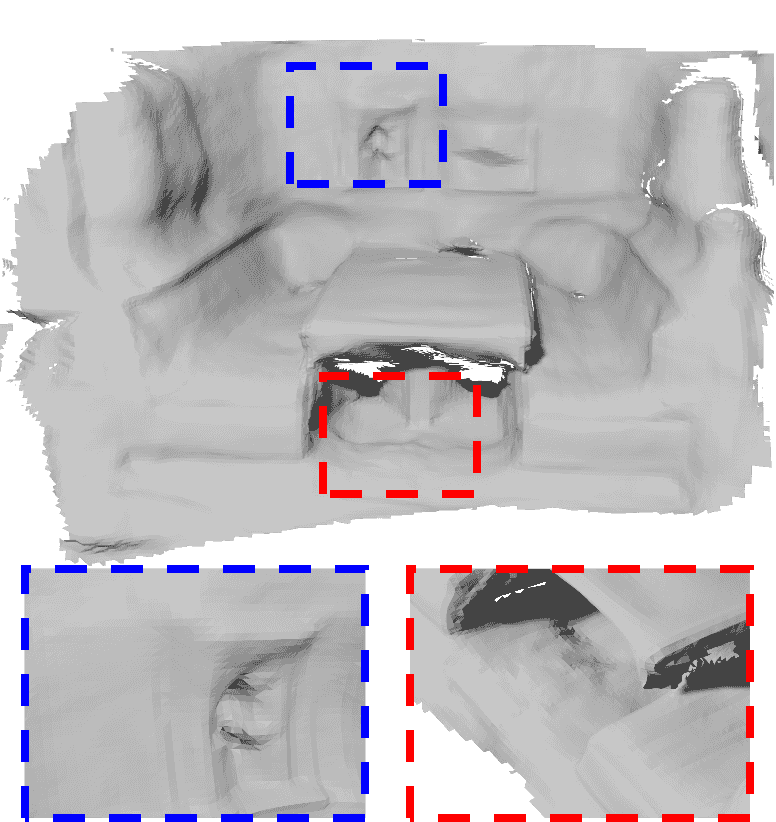}\vspace{\myvspace}
    \end{minipage}
  }
  \subfloat[Base + Distance prior]{
    \begin{minipage}[b]{\widthOfFullPage\linewidth}
      \centering
      \includegraphics[width=\widthOfMiniPage\linewidth]{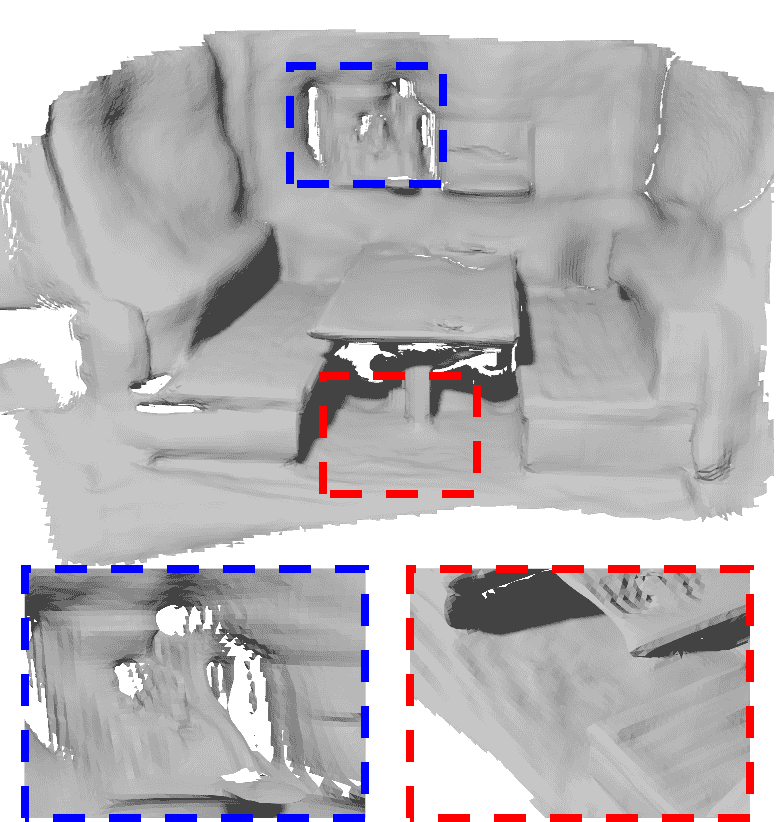}\vspace{\myvspace}
    \end{minipage}
  }
  \subfloat[Base + Normal prior]{
    \begin{minipage}[b]{\widthOfFullPage\linewidth}
      \centering
      \includegraphics[width=\widthOfMiniPage\linewidth]{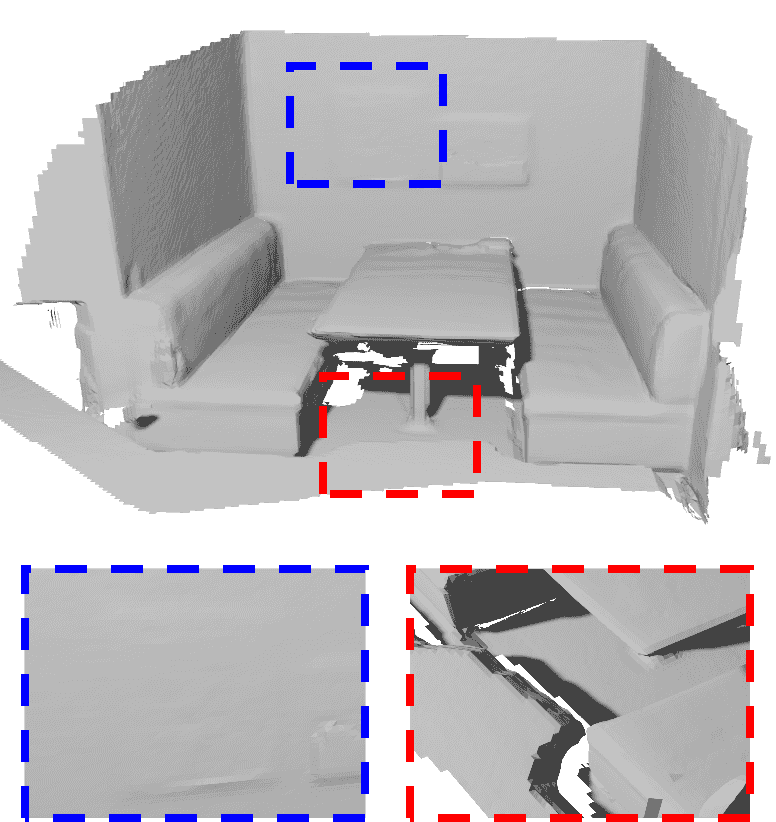}\vspace{\myvspace}
    \end{minipage}
  }
  \subfloat[Base + Prior]{
    \begin{minipage}[b]{\widthOfFullPage\linewidth}
      \centering
      \includegraphics[width=\widthOfMiniPage\linewidth]{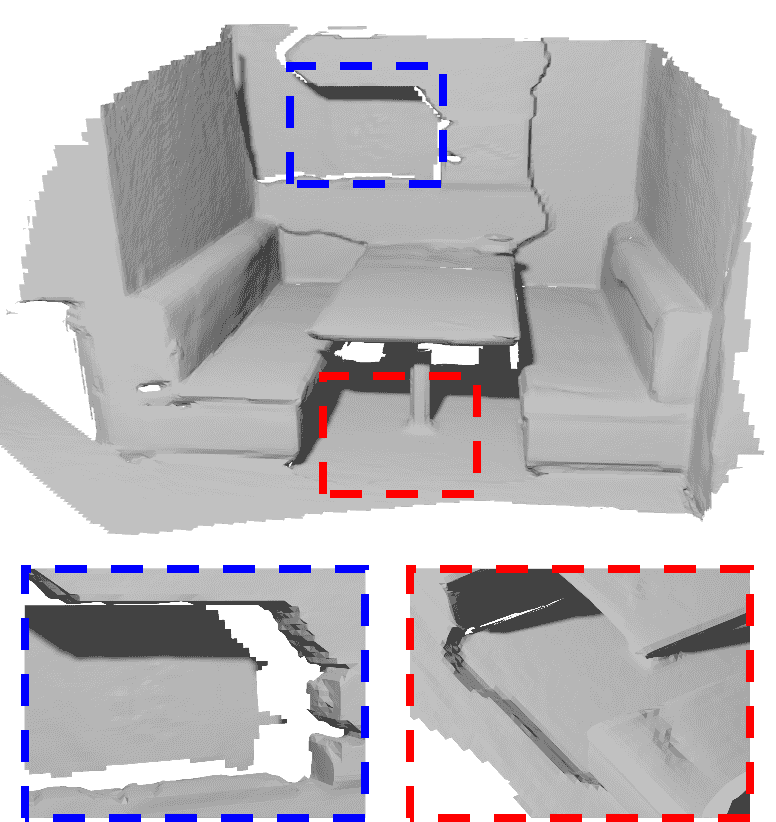}\vspace{\myvspace}
    \end{minipage}
  }
  \qquad 
  \subfloat[Base + Prior + Smooth]{
    \begin{minipage}[b]{\widthOfFullPage\linewidth}
      \centering
      \includegraphics[width=\widthOfMiniPage\linewidth]{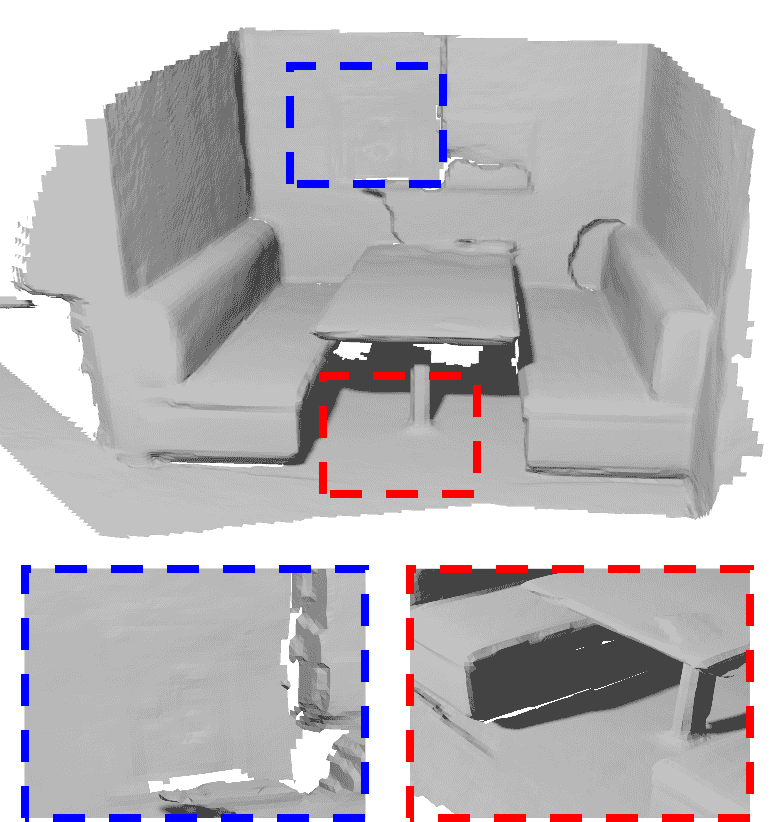}\vspace{\myvspace}
    \end{minipage}
  }
  \subfloat[Base + Prior + Consist]{
    \begin{minipage}[b]{\widthOfFullPage\linewidth}
      \centering
      \includegraphics[width=\widthOfMiniPage\linewidth]{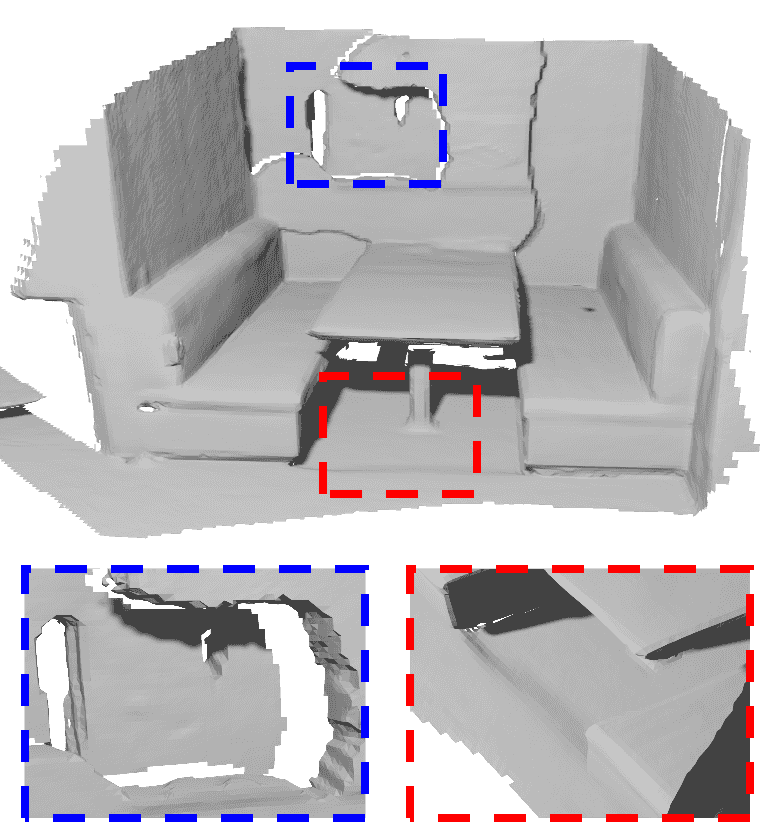}\vspace{\myvspace}
    \end{minipage}
  }
  \subfloat[Full]{
    \begin{minipage}[b]{\widthOfFullPage\linewidth}
      \centering
      \includegraphics[width=\widthOfMiniPage\linewidth]{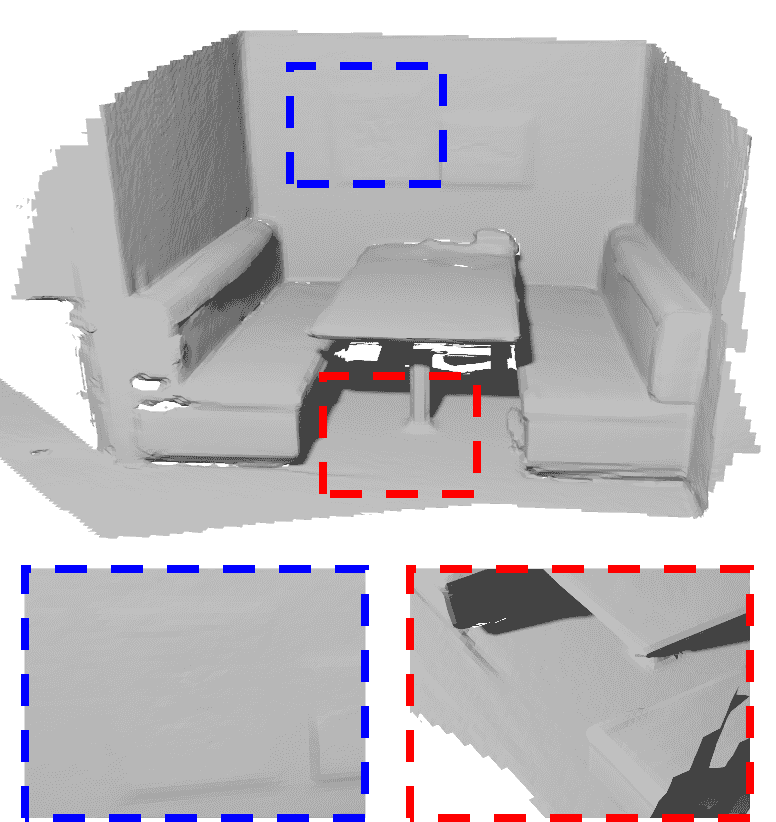}\vspace{\myvspace}
    \end{minipage}
  }
  \subfloat[GT]{
    \begin{minipage}[b]{\widthOfFullPage\linewidth}
      \centering
      \includegraphics[width=\widthOfMiniPage\linewidth]{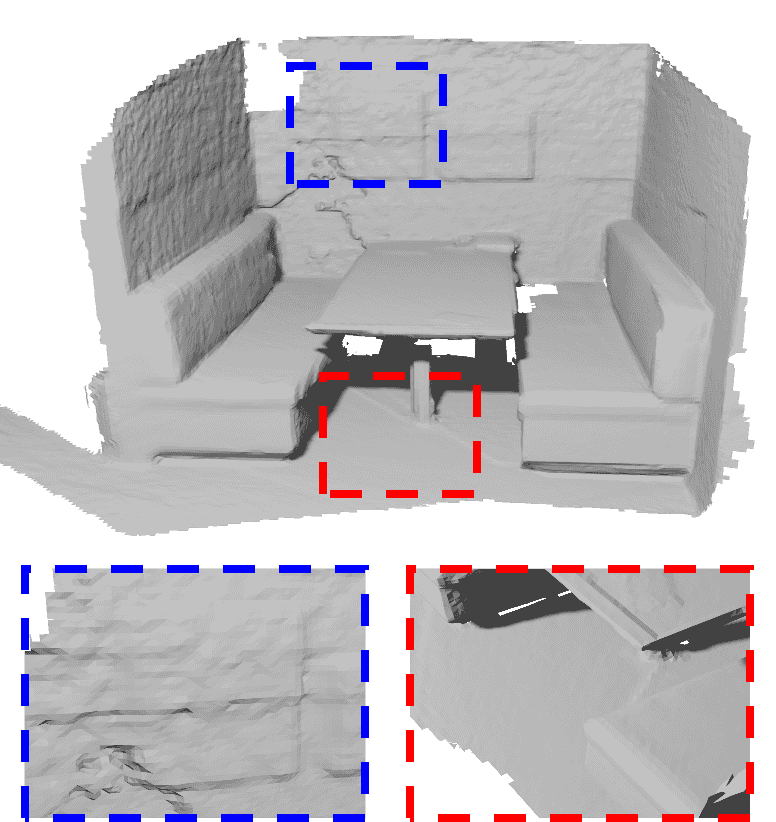}\vspace{\myvspace}
    \end{minipage}
  }
  \caption{Visualization results of the ablation study of a simple scene without a boundingbox. The analysis is presented in Section \ref{sec:ablation_st}. The quantitative comparisons are shown in Table \ref{ablation_study_table}.}
  \label{ablation_study_figure}
\end{figure*}

\begin{figure}[h]
  \centering
  \includegraphics[width=\linewidth]{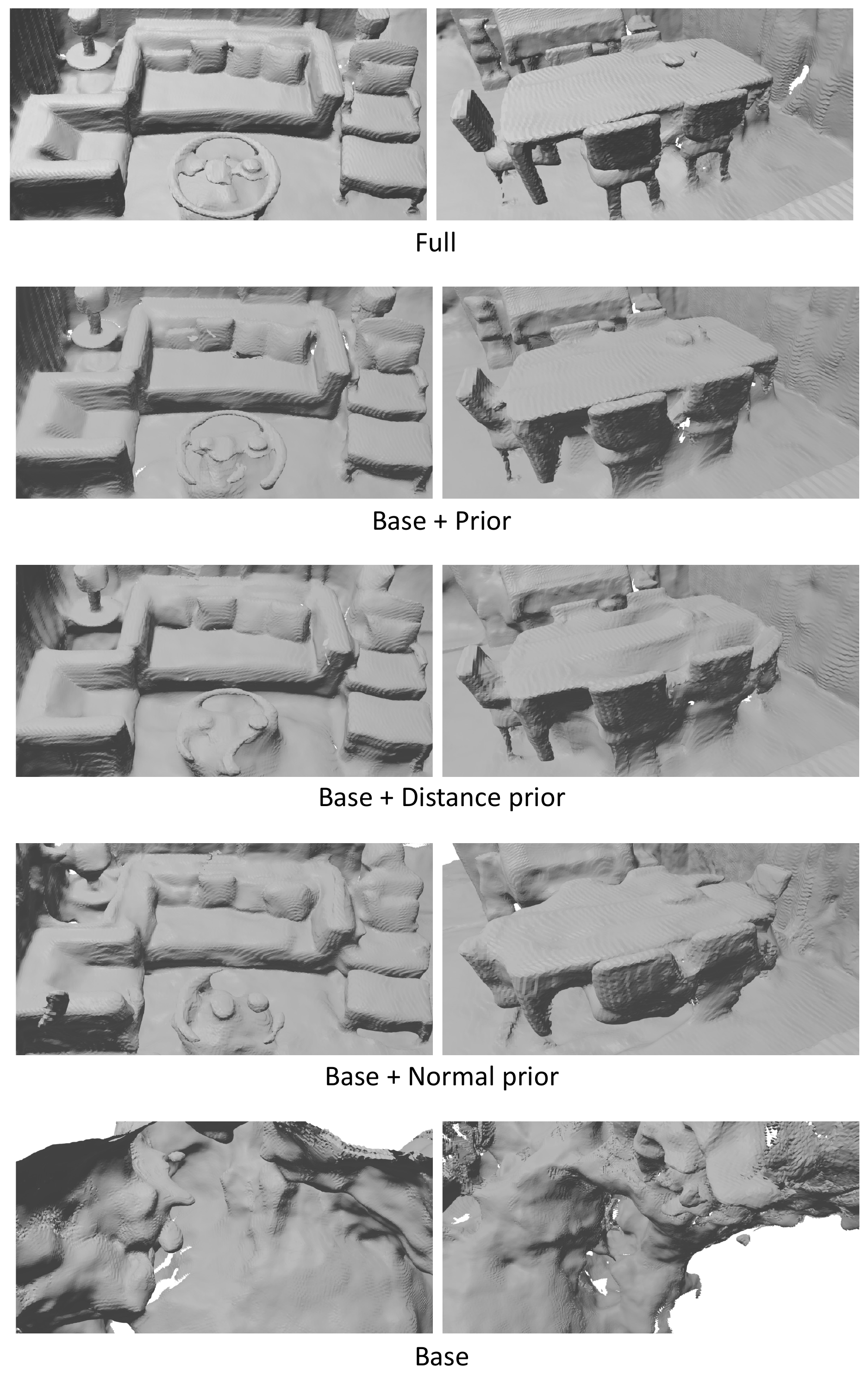}
  \caption{Visualization results of the ablation study on a real-world scene.}
  \label{Ablationreal}
\end{figure}

\paragraph{Distance prior.} The distance prior guides the implicit surface close to the corresponding point, which gives the NeuralRoom the ability to preserve the reliable spatial information. The weight of the distance prior loss in optimization can be adjusted according to its quality.

\paragraph{Normal prior.} The normal prior is the most important prior. With a normal prior, NeuralRoom can make the surface normal of the texture-less region consistent. This leads to better visual effects and quantitative evaluation. The weight of the normal prior loss in optimization can also be adjusted according to its quality.

\paragraph{Perturbation-residual restrictions.} The perturbation-residual restrictions ensure the continuity of the reconstructed scene and improve the accuracy and completeness. It establishes a connection between the sampling ray and the corresponding auxiliary ray.

\paragraph{Differentiable renderer.} The key to combining the above three aspects is our NeuralRoom differentiable renderer. The reliable priors reduce the possible spatial variation range of an implicit neural surface which helps the NeuralRoom alleviate shape-radiance ambiguity. The renderer takes the color loss as a primary loss to optimize an implicit neural surface that can render an image consistent with the input training images. In addition, the renderer has the ability to resist the influence caused by incorrect geometric priors. Similar to volumetric methods~\cite{murez2020atlas,sun2021neuralrecon}, the renderer considers the influence of all inputs in the reconstruction optimization, while the depth estimation method usually only considers the neighborhood.

\subsection{Ablation Study}
\label{sec:ablation_st}

To better understand the role of each optimization item, we performed ablation studies over each component of the proposed system. The experiment was conducted on Scene0801\_00 in ScanNet. Some simple scenes like that can be reconstructed without a bounding box and are sensitive to each loss term. The qualitative evaluation is shown in Figure~\ref{ablation_study_figure}, and the quantitative evaluation is shown in Table~\ref{ablation_study_table}. In addition, we also perform an ablation study on the complex real-world scene.

\begin{table}[h]
\caption{Ablation study of a simple scene without a boundingbox. Some scenes with simple geometry can be reconstructed without boundingbox like Scene0801\_00. We test the effect of each loss function in the method. This analysis shows that our full method performs best both visually and quantitatively.}
\centering
\label{ablation_study_table}
\scalebox{0.9}{
\begin{tabular}{lcccccc}
  \toprule
  \  & Method  & Comp$\downarrow$ & Acc$\downarrow$ & Overall$\downarrow$\\ 
  \midrule
  a & Base  & 0.055 & 0.157 & 0.106 \\
  b & Base + Distance prior  & 0.048 & 0.095 & 0.072 \\
  c & Base + Normal prior  & 0.050 & 0.059 & 0.054 \\
  d & Base + Prior  & 0.031 & 0.050 & 0.041 \\
  e & Base + Prior + Smooth  & 0.033 & 0.054 & 0.044 \\
  f & Base + Prior + Consist & 0.027 & 0.041 & 0.034 \\
  g & Full & \uline{0.022} & \uline{0.027} & \uline{0.024} \\
  \bottomrule
\end{tabular}
}
\end{table}

The distance prior $\mathcal{L}_{prior\_D}$ provides accurate 3D points, which helps improve the accuracy of the reconstruction. The rich textured and edge areas are well reconstructed. The normal prior $\mathcal{L}_{prior\_N}$ is the most important term of our system, which can significantly improve the completeness and accuracy of the scene. When both $\mathcal{L}_{prior\_D}$ and $\mathcal{L}_{prior\_N}$ participate in optimization at the same time, the accuracy and completeness of reconstruction are further improved. There are cracks in texture-less areas, although the quantitative results are better than before. The noise that exists in the priors and the lack of distance prior may cause this phenomenon. The residual-perturbation restrictions $\mathcal{L}_{smooth\_D}$ and $\mathcal{L}_{consist\_N}$ establish a connection between the sampling ray and its corresponding auxiliary ray, which improves the reconstruction quality of the surface. Using $\mathcal{L}_{consist\_N}$ alone can improve the completeness and accuracy, but more cracks appear in the scene. The normal of two points far apart in space can also be consistent, so only considering the normal constraint cannot determine a unique position in space. The $\mathcal{L}_{smooth\_D}$ term is used to limit the spatial distance between two points, which is designed to smooth the scene. When both local normal and smooth constraints are added to the optimization, we can obtain a better visual effect and quantitative evaluation result. In addition, we can adjust the corresponding weight according to the quality of the prior, so as to obtain a better reconstruction result.



\subsection{Computational cost}
When the layer number of Geo-MLP and Color-MLP are 8 and 4 respectively, and the number of sampling rays is 64+64, the GPU memory consumption is 9615M, and the time consumption is 16h on a single RTX 2080Ti. Without the perturbation-residual restriction, time and memory usage drop by about a quarter. When the number of sampling rays is reduced to 32+32, it takes 5353M and 10h. The reconstruction result will lose a little detail, but the overall reconstruction quality is OK. When the layers of Geo-MLP and Color-MLP are further reduced to 4 and 2 respectively, it takes 3405M and 7h. This setting is OK for reconstructing the scenes in the teaser and ablation study, but the capacity of MLP (4+2) is not enough for reconstructing some more complex scenes shown in Fig7. We believe that if the MLP is replaced with more advanced parameter representations like InstantNGP~\cite{muller2022instant} and TensoRF~\cite{tensorf}, the running speed can be increased significantly. 

\subsection{Advantages and Limitation}
\label{seclim}

\begin{figure}[h]
  \centering
  \includegraphics[width=\linewidth]{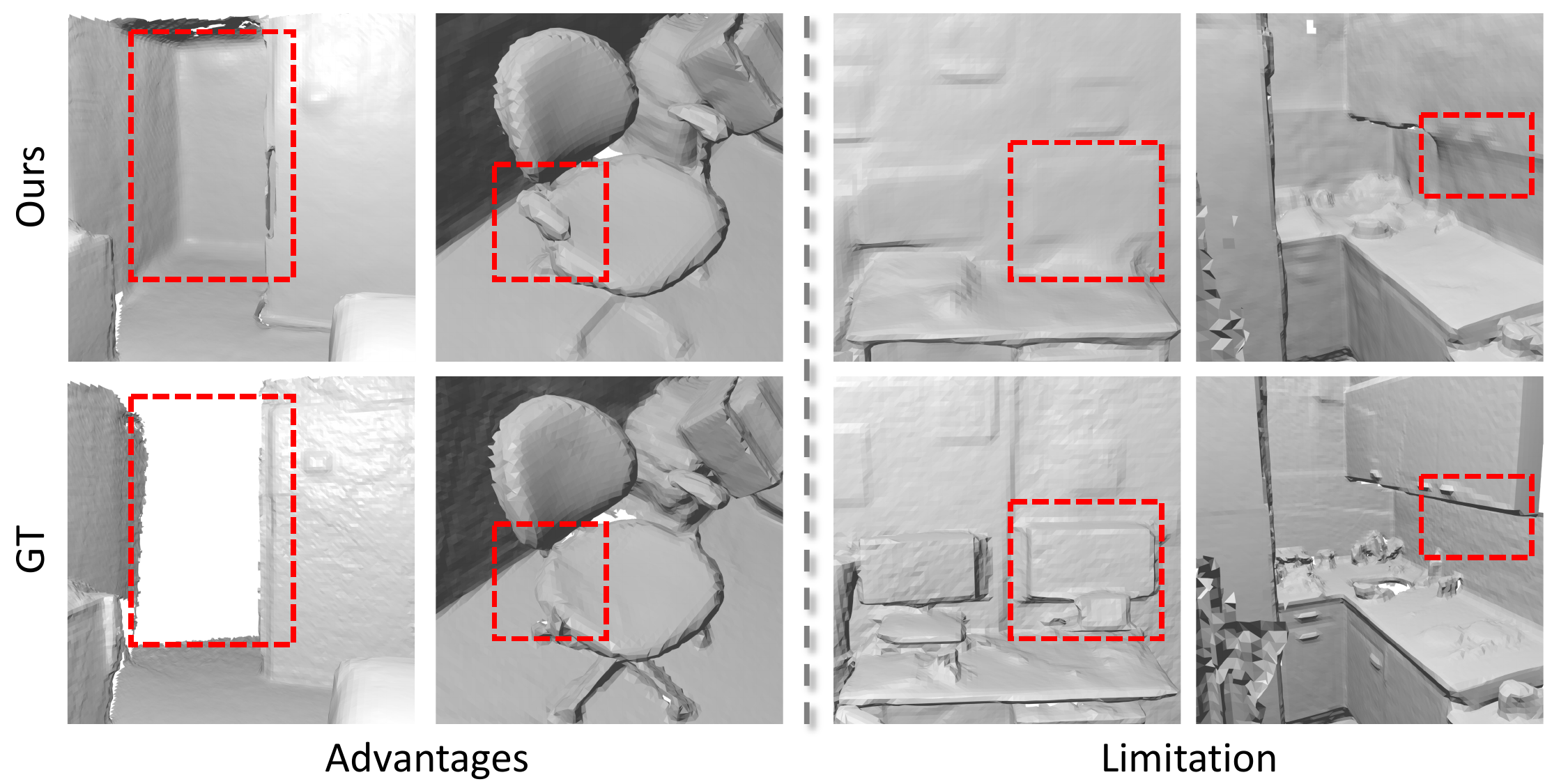}
  \caption{Advantages and limitations.}
  \label{adv_lmt}
\end{figure}

The depth sensor emitting infrared rays has difficulty to collect the depth information of the mirror, black area and distant objects, which results in incomplete reconstruction. NeuralRoom takes RGB images as input, which can restore this part of the scene (Figure~\ref{adv_lmt} left). However, there are several limitations of the proposed method.
\paragraph{Poses and priors.} Almost all rendering-based reconstruction methods rely on the accurate camera pose. Therefore, large poses and priors errors have a strong adverse impact on reconstruction results. In addition, when the two surfaces with the same normal are close to each other, if the correct distance prior is missing, the NeuralRoom will over smooth them (Figure~\ref{adv_lmt}, right). Using more advanced learning-based pose, depth and normal estimation methods to provide more reliable geometry priors and camera poses can ensure the reconstruction quality.
\paragraph{Computational cost.} Our method requires a large amount of computational resources for geometry prior computation and renderer training. Accelerating with an updated differentiable rendering~\cite{yu2021plenoxels,muller2022instant} architecture is a direction for improvement. Moreover, using a neural network such as~\cite{murez2020atlas} to quickly reconstruct a structure of the scene and then using a differentiable renderer or other learning-based methods to adjust the details of the indoor scene may be a good solution.

\subsection{NeuralRoom-Advanced}
\label{secnradv}
We use more advanced indoor scene reconstruction algorithms~\cite{murez2020atlas,sun2021neuralrecon,rich20213dvnet} instead of COLMAP~\cite{schonberger2016structure} to provide distance priors, and show the improved reconstruction quality of these algorithms achieved by our proposed NeuralRoom system.

\begin{table}[h]
\centering
\caption{Quantitative evaluation illustrating the improvement of existing reconstruction algorithms achieved by our proposed NeuralRoom system.}
\label{ipm_eva}
\scalebox{0.85}{
\begin{tabular}{lcccccc}
  \toprule
  Method & Prec$\uparrow$ & Recall$\uparrow$ & F-score$\uparrow$ & Acc$\downarrow$ & Comp$\downarrow$ & Overall$\downarrow$\\
  \midrule
   NeuralRoom & 68.347 & 65.298 & 66.756 & 0.051 & 0.058 & 0.055\\
   NR-Atlas & 73.003 & 69.062 & 70.948 & \uline{0.044} & 0.052 & \uline{0.048}\\
   NR-NeuralRecon & 69.977 & 66.965 & 68.393 & 0.054 & 0.059 & 0.056\\
   NR-3DVNet & \uline{73.339} & \uline{70.496} & \uline{71.865} & 0.046 & \uline{0.051} & \uline{0.048}\\
  \bottomrule
\end{tabular}
}
\end{table}

\begin{figure}[h]
  \centering 
  \newcommand{\myhspace}{-4pt} 
  \newcommand{\myabovecaptionskip}{0.1pt} 
  \newcommand{\widthOfFullPage}{0.31} 
  \newcommand{\widthOfMiniPage}{0.99}
  \captionsetup[subfloat]{labelsep=none,format=plain,labelformat=empty}
  \subfloat[Atlas]{
    \begin{minipage}[b]{\widthOfFullPage\linewidth}
      \centering
      \includegraphics[width=\widthOfMiniPage\linewidth]{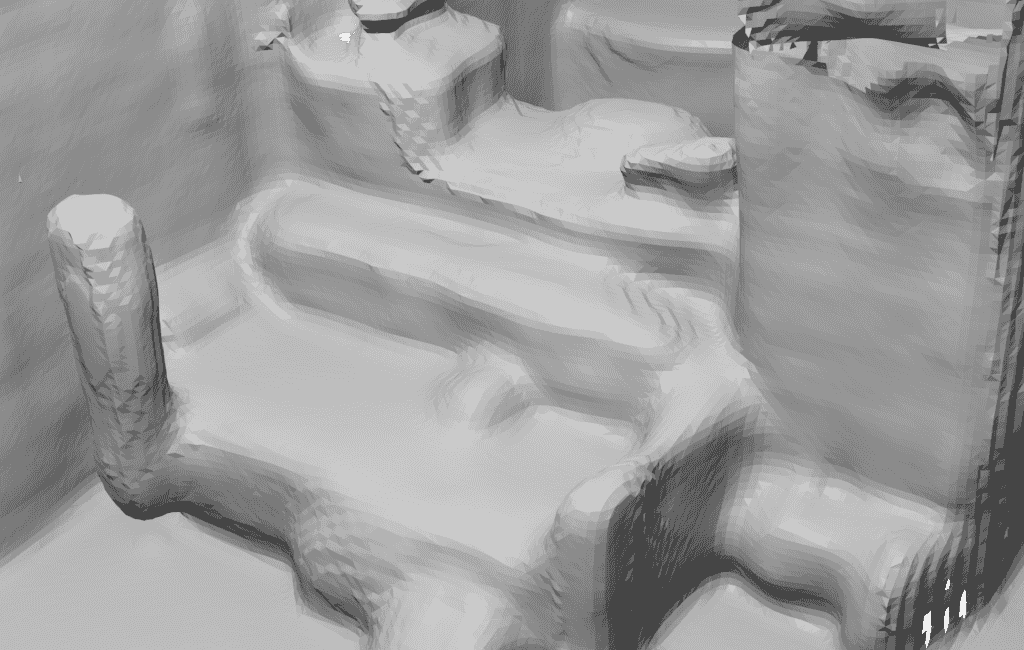}
    \end{minipage}\setlength{\abovecaptionskip}{\myabovecaptionskip} 
  }
  \subfloat[NR-Atlas]{
    \begin{minipage}[b]{\widthOfFullPage\linewidth}
      \centering
      \includegraphics[width=\widthOfMiniPage\linewidth]{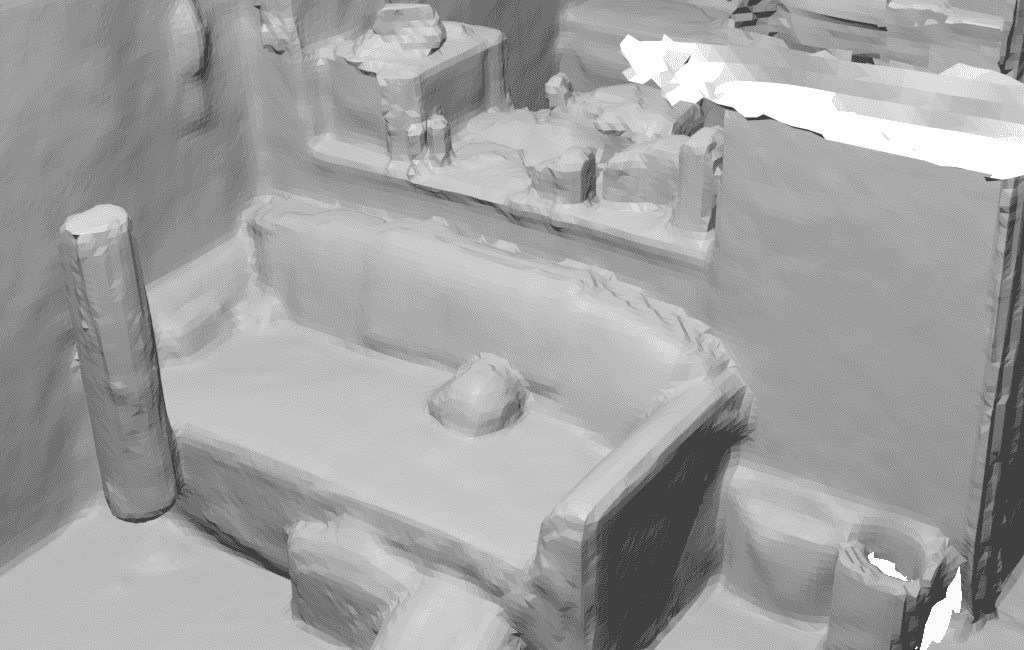}
    \end{minipage}
  }
  \subfloat[GT]{
    \begin{minipage}[b]{\widthOfFullPage\linewidth}
      \centering
      \includegraphics[width=\widthOfMiniPage\linewidth]{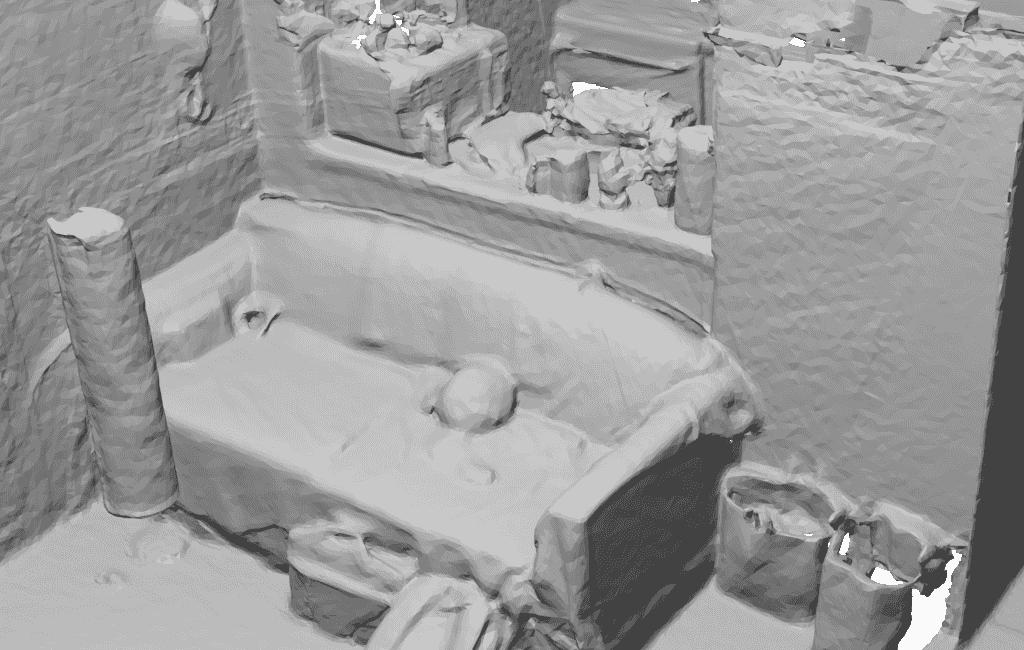}
    \end{minipage}
  }
  \\ 
  \subfloat[3DVNet]{
    \begin{minipage}[b]{\widthOfFullPage\linewidth}
      \centering
      \includegraphics[width=\widthOfMiniPage\linewidth]{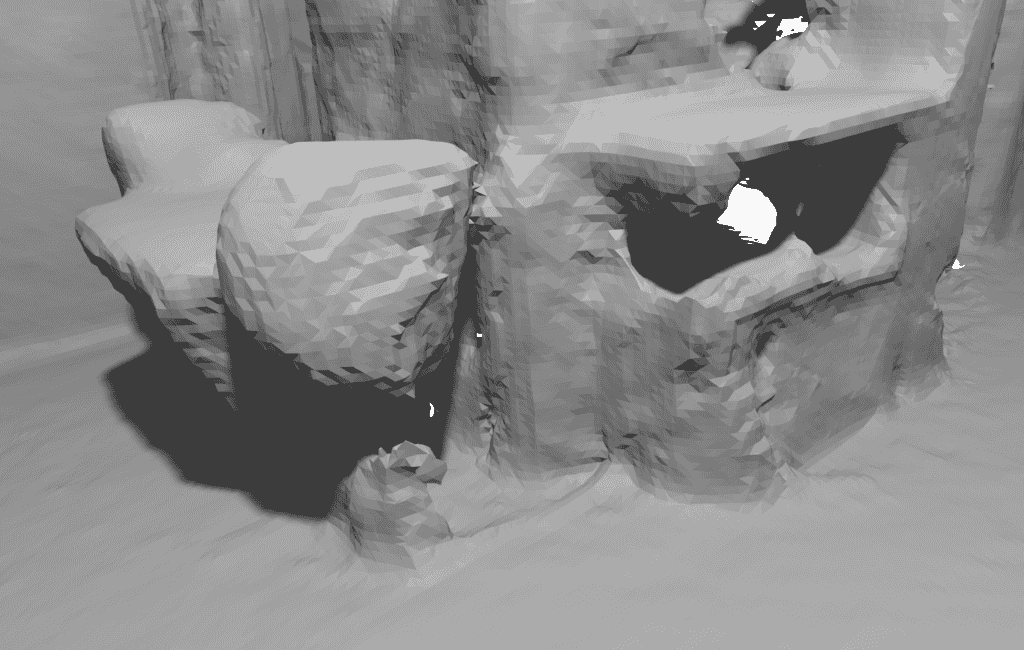}
    \end{minipage}
  }
  \subfloat[NR-3DVNet]{
    \begin{minipage}[b]{\widthOfFullPage\linewidth}
      \centering
      \includegraphics[width=\widthOfMiniPage\linewidth]{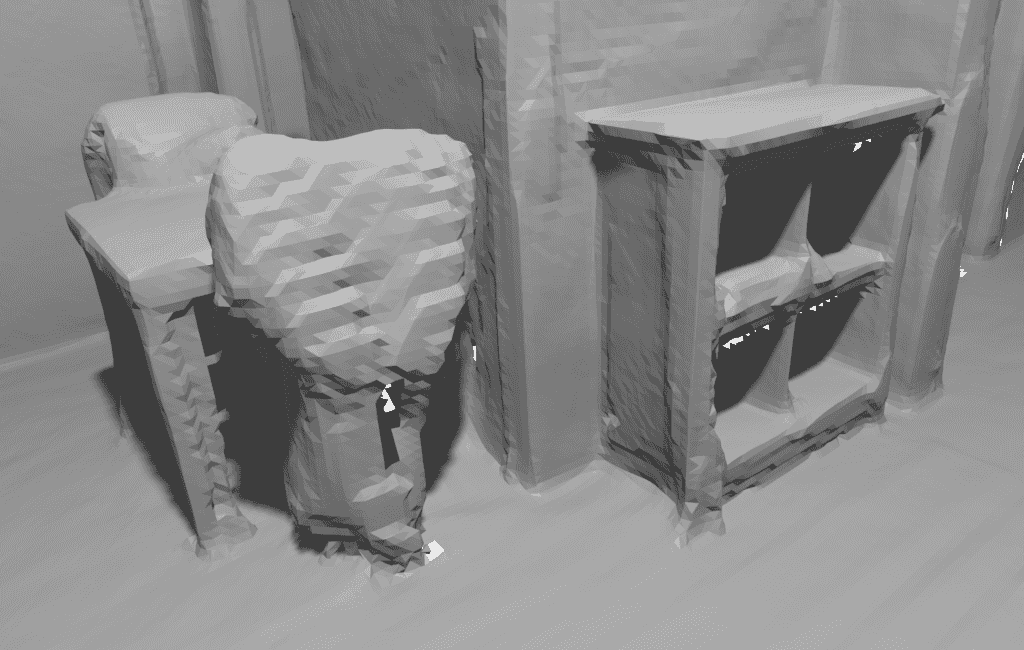}
    \end{minipage}
  }
  \subfloat[GT]{
    \begin{minipage}[b]{\widthOfFullPage\linewidth}
      \centering
      \includegraphics[width=\widthOfMiniPage\linewidth]{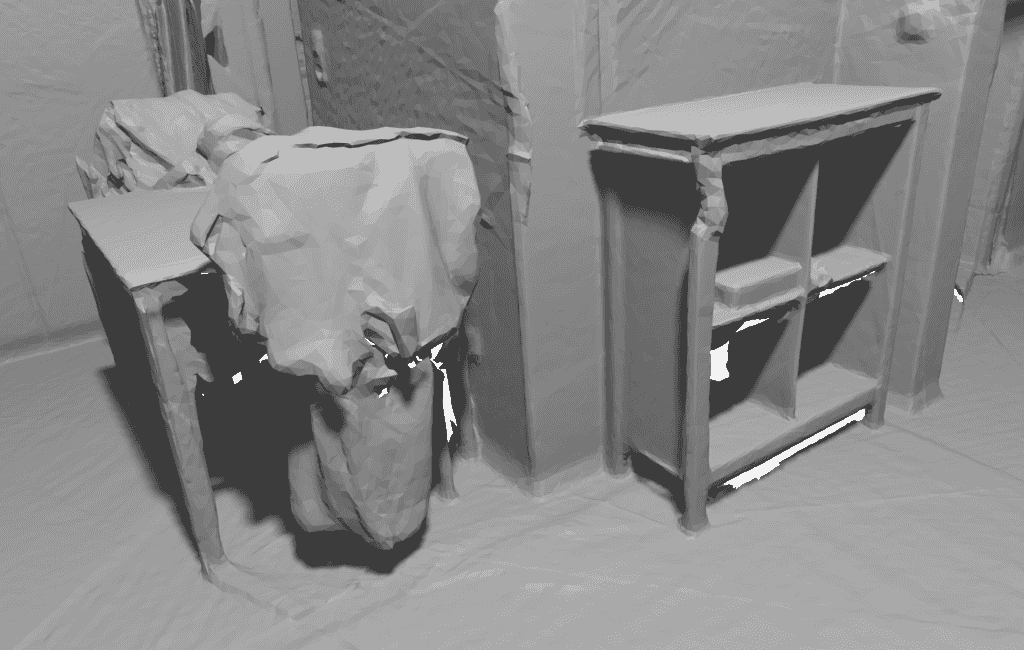}
    \end{minipage}
  }
  \\ 
  \subfloat[NeuralRecon]{
    \begin{minipage}[b]{\widthOfFullPage\linewidth}
      \centering
      \includegraphics[width=\widthOfMiniPage\linewidth]{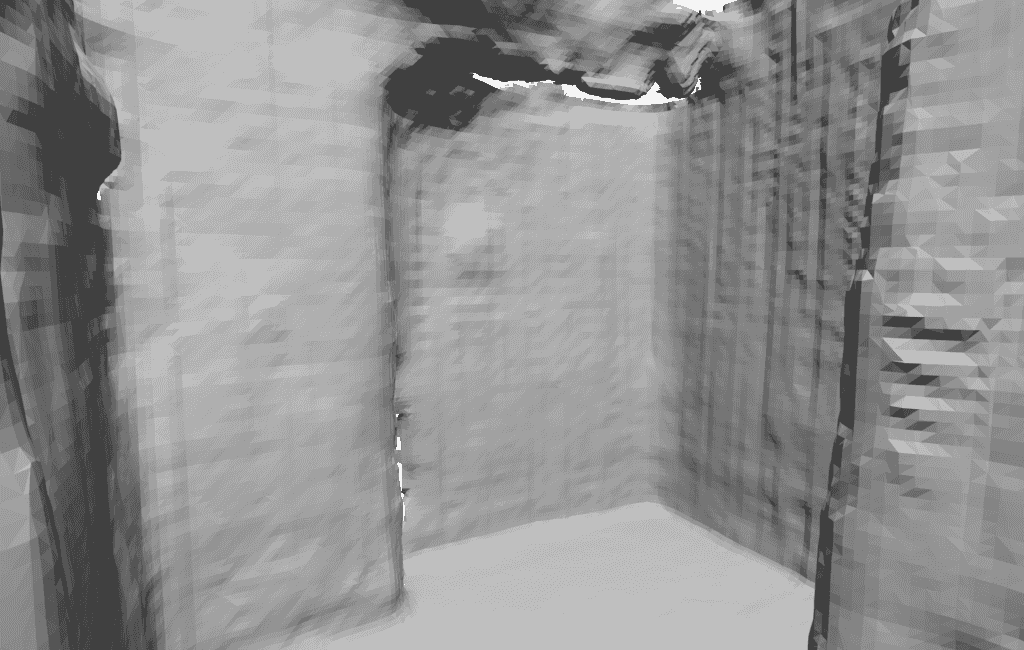}
    \end{minipage}
  }
  \subfloat[NR-NeuralRecon]{
    \begin{minipage}[b]{\widthOfFullPage\linewidth}
      \centering
      \includegraphics[width=\widthOfMiniPage\linewidth]{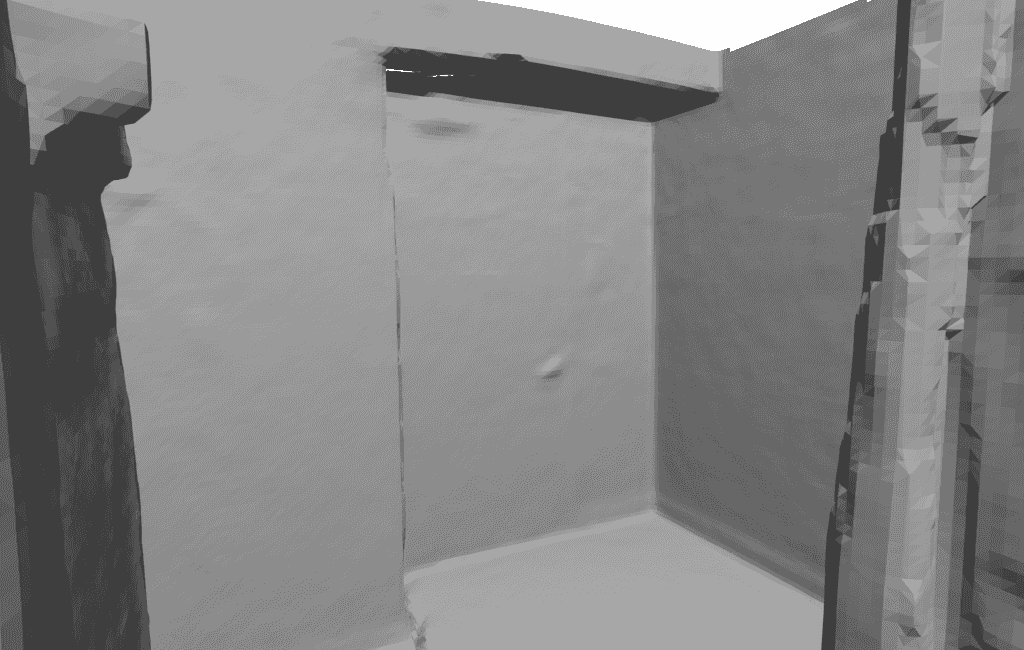}
    \end{minipage}
  }
  \subfloat[GT]{
    \begin{minipage}[b]{\widthOfFullPage\linewidth}
      \centering
      \includegraphics[width=\widthOfMiniPage\linewidth]{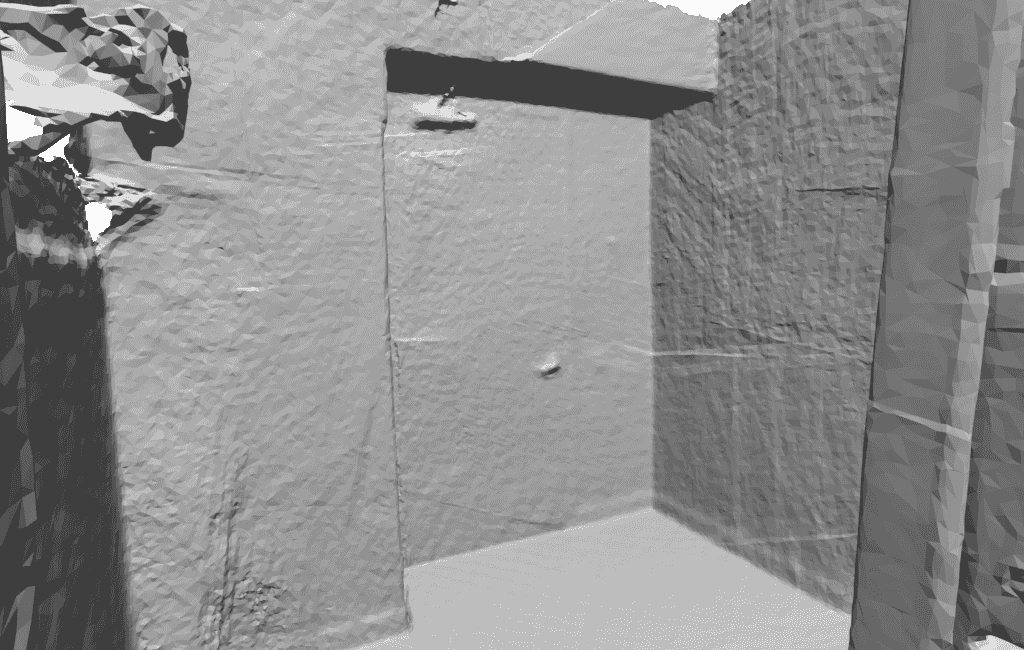}
    \end{minipage}
  }
  \caption{The improvement of reconstruction quality of different reconstruction algorithms by our proposed NeuralRoom system.}
  \label{impv3dv}
\end{figure}

These learning-based multiview reconstruction algorithms provide a distance prior with higher accuracy or better completeness than COLMAP~\cite{schonberger2016structure}, which will help further alleviate shape-radiance ambiguity resulting in better reconstructions. Table~\ref{ipm_eva} shows the quantitative evaluation and Figure~\ref{impv3dv} shows the qualitative comparisons, the related algorithms improved by NeuralRoom have an NR- prefix. In addition to replacing the distance prior module in NeuralRoom, researchers can further improve the reconstruction quality by replacing the normal estimation module and differentiable renderer with their own more advanced algorithms.

\section{Conclusions}
We present NeuralRoom, a rendering-based neural surface reconstruction method for reconstructing indoor scenes directly from a set of 2D images. The key to successful indoor scene reconstruction by using differentiable rendering is that we find complementarity between depth estimation and normal estimation methods, which helps alleviate inherent shape-radiance ambiguity. We also design geometric constraints for the renderer to obtain a more smoothing and complete surface reconstruction. NeuralRoom produces impressive reconstruction and successfully reconstructs the surface with no texture or rich texture. In addition, we show the reconstruction improvement of existing multiview reconstruction algorithms by incorporating NeuralRoom pipeline.


\begin{acks}
This work is partially supported by NSFC (No. 61972298) and Wuhan University-Huawei GeoInformatices Innovation Lab.
\end{acks}


\bibliographystyle{ACM-Reference-Format}
\normalem
\bibliography{sample-bibliography}


\begin{thebibliography}{91}


\ifx \showCODEN    \undefined \def \showCODEN     #1{\unskip}     \fi
\ifx \showDOI      \undefined \def \showDOI       #1{#1}\fi
\ifx \showISBNx    \undefined \def \showISBNx     #1{\unskip}     \fi
\ifx \showISBNxiii \undefined \def \showISBNxiii  #1{\unskip}     \fi
\ifx \showISSN     \undefined \def \showISSN      #1{\unskip}     \fi
\ifx \showLCCN     \undefined \def \showLCCN      #1{\unskip}     \fi
\ifx \shownote     \undefined \def \shownote      #1{#1}          \fi
\ifx \showarticletitle \undefined \def \showarticletitle #1{#1}   \fi
\ifx \showURL      \undefined \def \showURL       {\relax}        \fi
\providecommand\bibfield[2]{#2}
\providecommand\bibinfo[2]{#2}
\providecommand\natexlab[1]{#1}
\providecommand\showeprint[2][]{arXiv:#2}

\bibitem[\protect\citeauthoryear{Atzmon, Haim, Yariv, Israelov, Maron, and
  Lipman}{Atzmon et~al\mbox{.}}{2019}]%
        {atzmon2019controlling}
\bibfield{author}{\bibinfo{person}{Matan Atzmon}, \bibinfo{person}{Niv Haim},
  \bibinfo{person}{Lior Yariv}, \bibinfo{person}{Ofer Israelov},
  \bibinfo{person}{Haggai Maron}, {and} \bibinfo{person}{Yaron Lipman}.}
  \bibinfo{year}{2019}\natexlab{}.
\newblock \showarticletitle{Controlling neural level sets}.
\newblock \bibinfo{journal}{\emph{Advances in Neural Information Processing
  Systems}}  \bibinfo{volume}{32} (\bibinfo{year}{2019}).
\newblock


\bibitem[\protect\citeauthoryear{Atzmon and Lipman}{Atzmon and Lipman}{2020}]%
        {atzmon2020sal}
\bibfield{author}{\bibinfo{person}{Matan Atzmon} {and} \bibinfo{person}{Yaron
  Lipman}.} \bibinfo{year}{2020}\natexlab{}.
\newblock \showarticletitle{Sal: Sign agnostic learning of shapes from raw
  data}. In \bibinfo{booktitle}{\emph{Proceedings of the IEEE/CVF Conference on
  Computer Vision and Pattern Recognition}}. \bibinfo{pages}{2565--2574}.
\newblock


\bibitem[\protect\citeauthoryear{Bae, Budvytis, and Cipolla}{Bae
  et~al\mbox{.}}{2021}]%
        {bae2021estimating}
\bibfield{author}{\bibinfo{person}{Gwangbin Bae}, \bibinfo{person}{Ignas
  Budvytis}, {and} \bibinfo{person}{Roberto Cipolla}.}
  \bibinfo{year}{2021}\natexlab{}.
\newblock \showarticletitle{Estimating and Exploiting the Aleatoric Uncertainty
  in Surface Normal Estimation}. In \bibinfo{booktitle}{\emph{Proceedings of
  the IEEE/CVF International Conference on Computer Vision}}.
  \bibinfo{pages}{13137--13146}.
\newblock


\bibitem[\protect\citeauthoryear{Barron, Mildenhall, Tancik, Hedman,
  Martin-Brualla, and Srinivasan}{Barron et~al\mbox{.}}{2021}]%
        {barron2021mip}
\bibfield{author}{\bibinfo{person}{Jonathan~T Barron}, \bibinfo{person}{Ben
  Mildenhall}, \bibinfo{person}{Matthew Tancik}, \bibinfo{person}{Peter
  Hedman}, \bibinfo{person}{Ricardo Martin-Brualla}, {and}
  \bibinfo{person}{Pratul~P Srinivasan}.} \bibinfo{year}{2021}\natexlab{}.
\newblock \showarticletitle{Mip-nerf: A multiscale representation for
  anti-aliasing neural radiance fields}. In
  \bibinfo{booktitle}{\emph{Proceedings of the IEEE/CVF International
  Conference on Computer Vision}}. \bibinfo{pages}{5855--5864}.
\newblock


\bibitem[\protect\citeauthoryear{Bernardini, Mittleman, Rushmeier, Silva, and
  Taubin}{Bernardini et~al\mbox{.}}{1999}]%
        {bernardini1999ball}
\bibfield{author}{\bibinfo{person}{Fausto Bernardini}, \bibinfo{person}{Joshua
  Mittleman}, \bibinfo{person}{Holly Rushmeier}, \bibinfo{person}{Cl{\'a}udio
  Silva}, {and} \bibinfo{person}{Gabriel Taubin}.}
  \bibinfo{year}{1999}\natexlab{}.
\newblock \showarticletitle{The ball-pivoting algorithm for surface
  reconstruction}.
\newblock \bibinfo{journal}{\emph{IEEE transactions on visualization and
  computer graphics}} \bibinfo{volume}{5}, \bibinfo{number}{4}
  (\bibinfo{year}{1999}), \bibinfo{pages}{349--359}.
\newblock


\bibitem[\protect\citeauthoryear{Bokhovkin and Dai}{Bokhovkin and Dai}{2022}]%
        {bokhovkin2022neural}
\bibfield{author}{\bibinfo{person}{Alexey Bokhovkin} {and}
  \bibinfo{person}{Angela Dai}.} \bibinfo{year}{2022}\natexlab{}.
\newblock \showarticletitle{Neural Part Priors: Learning to Optimize Part-Based
  Object Completion in RGB-D Scans}.
\newblock \bibinfo{journal}{\emph{arXiv preprint arXiv:2203.09375}}
  (\bibinfo{year}{2022}).
\newblock


\bibitem[\protect\citeauthoryear{Buehler, Bosse, McMillan, Gortler, and
  Cohen}{Buehler et~al\mbox{.}}{2001}]%
        {buehler2001unstructured}
\bibfield{author}{\bibinfo{person}{Chris Buehler}, \bibinfo{person}{Michael
  Bosse}, \bibinfo{person}{Leonard McMillan}, \bibinfo{person}{Steven Gortler},
  {and} \bibinfo{person}{Michael Cohen}.} \bibinfo{year}{2001}\natexlab{}.
\newblock \showarticletitle{Unstructured lumigraph rendering}. In
  \bibinfo{booktitle}{\emph{Proceedings of the 28th annual conference on
  Computer graphics and interactive techniques}}. \bibinfo{pages}{425--432}.
\newblock


\bibitem[\protect\citeauthoryear{Chai, Tong, Chan, and Shum}{Chai
  et~al\mbox{.}}{2000}]%
        {chai2000plenoptic}
\bibfield{author}{\bibinfo{person}{Jin-Xiang Chai}, \bibinfo{person}{Xin Tong},
  \bibinfo{person}{Shing-Chow Chan}, {and} \bibinfo{person}{Heung-Yeung Shum}.}
  \bibinfo{year}{2000}\natexlab{}.
\newblock \showarticletitle{Plenoptic sampling}. In
  \bibinfo{booktitle}{\emph{Proceedings of the 27th annual conference on
  Computer graphics and interactive techniques}}. \bibinfo{pages}{307--318}.
\newblock


\bibitem[\protect\citeauthoryear{Chen, Xu, Geiger, Yu, and Su}{Chen
  et~al\mbox{.}}{2022}]%
        {tensorf}
\bibfield{author}{\bibinfo{person}{Anpei Chen}, \bibinfo{person}{Zexiang Xu},
  \bibinfo{person}{Andreas Geiger}, \bibinfo{person}{Jingyi Yu}, {and}
  \bibinfo{person}{Hao Su}.} \bibinfo{year}{2022}\natexlab{}.
\newblock \showarticletitle{TensoRF: Tensorial Radiance Fields}.
\newblock \bibinfo{journal}{\emph{arXiv preprint arXiv:2203.09517}}
  (\bibinfo{year}{2022}).
\newblock


\bibitem[\protect\citeauthoryear{Chen, Xu, Zhao, Zhang, Xiang, Yu, and Su}{Chen
  et~al\mbox{.}}{2021}]%
        {chen2021mvsnerf}
\bibfield{author}{\bibinfo{person}{Anpei Chen}, \bibinfo{person}{Zexiang Xu},
  \bibinfo{person}{Fuqiang Zhao}, \bibinfo{person}{Xiaoshuai Zhang},
  \bibinfo{person}{Fanbo Xiang}, \bibinfo{person}{Jingyi Yu}, {and}
  \bibinfo{person}{Hao Su}.} \bibinfo{year}{2021}\natexlab{}.
\newblock \showarticletitle{Mvsnerf: Fast generalizable radiance field
  reconstruction from multi-view stereo}. In
  \bibinfo{booktitle}{\emph{Proceedings of the IEEE/CVF International
  Conference on Computer Vision}}. \bibinfo{pages}{14124--14133}.
\newblock


\bibitem[\protect\citeauthoryear{Chen, Han, Xu, and Su}{Chen
  et~al\mbox{.}}{2019}]%
        {chen2019point}
\bibfield{author}{\bibinfo{person}{Rui Chen}, \bibinfo{person}{Songfang Han},
  \bibinfo{person}{Jing Xu}, {and} \bibinfo{person}{Hao Su}.}
  \bibinfo{year}{2019}\natexlab{}.
\newblock \showarticletitle{Point-based multi-view stereo network}. In
  \bibinfo{booktitle}{\emph{Proceedings of the IEEE/CVF International
  Conference on Computer Vision}}. \bibinfo{pages}{1538--1547}.
\newblock


\bibitem[\protect\citeauthoryear{Cheng, Xu, Zhu, Li, Li, Ramamoorthi, and
  Su}{Cheng et~al\mbox{.}}{2020}]%
        {cheng2020deep}
\bibfield{author}{\bibinfo{person}{Shuo Cheng}, \bibinfo{person}{Zexiang Xu},
  \bibinfo{person}{Shilin Zhu}, \bibinfo{person}{Zhuwen Li},
  \bibinfo{person}{Li~Erran Li}, \bibinfo{person}{Ravi Ramamoorthi}, {and}
  \bibinfo{person}{Hao Su}.} \bibinfo{year}{2020}\natexlab{}.
\newblock \showarticletitle{Deep stereo using adaptive thin volume
  representation with uncertainty awareness}. In
  \bibinfo{booktitle}{\emph{Proceedings of the IEEE/CVF Conference on Computer
  Vision and Pattern Recognition}}. \bibinfo{pages}{2524--2534}.
\newblock


\bibitem[\protect\citeauthoryear{Choy, Xu, Gwak, Chen, and Savarese}{Choy
  et~al\mbox{.}}{2016}]%
        {choy20163d}
\bibfield{author}{\bibinfo{person}{Christopher~B Choy}, \bibinfo{person}{Danfei
  Xu}, \bibinfo{person}{JunYoung Gwak}, \bibinfo{person}{Kevin Chen}, {and}
  \bibinfo{person}{Silvio Savarese}.} \bibinfo{year}{2016}\natexlab{}.
\newblock \showarticletitle{3d-r2n2: A unified approach for single and
  multi-view 3d object reconstruction}. In \bibinfo{booktitle}{\emph{European
  conference on computer vision}}. Springer, \bibinfo{pages}{628--644}.
\newblock


\bibitem[\protect\citeauthoryear{Collins}{Collins}{1996}]%
        {collins1996space}
\bibfield{author}{\bibinfo{person}{Robert~T Collins}.}
  \bibinfo{year}{1996}\natexlab{}.
\newblock \showarticletitle{A space-sweep approach to true multi-image
  matching}. In \bibinfo{booktitle}{\emph{Proceedings CVPR IEEE Computer
  Society Conference on Computer Vision and Pattern Recognition}}. IEEE,
  \bibinfo{pages}{358--363}.
\newblock


\bibitem[\protect\citeauthoryear{Curless and Levoy}{Curless and Levoy}{1996}]%
        {curless1996volumetric}
\bibfield{author}{\bibinfo{person}{Brian Curless} {and} \bibinfo{person}{Marc
  Levoy}.} \bibinfo{year}{1996}\natexlab{}.
\newblock \showarticletitle{A volumetric method for building complex models
  from range images}. In \bibinfo{booktitle}{\emph{Proceedings of the 23rd
  annual conference on Computer graphics and interactive techniques}}.
  \bibinfo{pages}{303--312}.
\newblock


\bibitem[\protect\citeauthoryear{Dai, Chang, Savva, Halber, Funkhouser, and
  Nie{\ss}ner}{Dai et~al\mbox{.}}{2017}]%
        {dai2017scannet}
\bibfield{author}{\bibinfo{person}{Angela Dai}, \bibinfo{person}{Angel~X.
  Chang}, \bibinfo{person}{Manolis Savva}, \bibinfo{person}{Maciej Halber},
  \bibinfo{person}{Thomas Funkhouser}, {and} \bibinfo{person}{Matthias
  Nie{\ss}ner}.} \bibinfo{year}{2017}\natexlab{}.
\newblock \showarticletitle{ScanNet: Richly-annotated 3D Reconstructions of
  Indoor Scenes}. In \bibinfo{booktitle}{\emph{Proc. Computer Vision and
  Pattern Recognition (CVPR), IEEE}}.
\newblock


\bibitem[\protect\citeauthoryear{Dai, Diller, and Nie{\ss}ner}{Dai
  et~al\mbox{.}}{2020}]%
        {sg-nn}
\bibfield{author}{\bibinfo{person}{Angela Dai}, \bibinfo{person}{Christian
  Diller}, {and} \bibinfo{person}{Matthias Nie{\ss}ner}.}
  \bibinfo{year}{2020}\natexlab{}.
\newblock \showarticletitle{Sg-nn: Sparse generative neural networks for
  self-supervised scene completion of rgb-d scans}. In
  \bibinfo{booktitle}{\emph{Proceedings of the IEEE/CVF Conference on Computer
  Vision and Pattern Recognition}}. \bibinfo{pages}{849--858}.
\newblock


\bibitem[\protect\citeauthoryear{Dai, Siddiqui, Thies, Valentin, and
  Nie{\ss}ner}{Dai et~al\mbox{.}}{2021}]%
        {dai2021spsg}
\bibfield{author}{\bibinfo{person}{Angela Dai}, \bibinfo{person}{Yawar
  Siddiqui}, \bibinfo{person}{Justus Thies}, \bibinfo{person}{Julien Valentin},
  {and} \bibinfo{person}{Matthias Nie{\ss}ner}.}
  \bibinfo{year}{2021}\natexlab{}.
\newblock \showarticletitle{Spsg: Self-supervised photometric scene generation
  from rgb-d scans}. In \bibinfo{booktitle}{\emph{Proceedings of the IEEE/CVF
  Conference on Computer Vision and Pattern Recognition}}.
  \bibinfo{pages}{1747--1756}.
\newblock


\bibitem[\protect\citeauthoryear{Debevec, Taylor, and Malik}{Debevec
  et~al\mbox{.}}{1996}]%
        {debevec1996modeling}
\bibfield{author}{\bibinfo{person}{Paul~E Debevec}, \bibinfo{person}{Camillo~J
  Taylor}, {and} \bibinfo{person}{Jitendra Malik}.}
  \bibinfo{year}{1996}\natexlab{}.
\newblock \showarticletitle{Modeling and rendering architecture from
  photographs: A hybrid geometry-and image-based approach}. In
  \bibinfo{booktitle}{\emph{Proceedings of the 23rd annual conference on
  Computer graphics and interactive techniques}}. \bibinfo{pages}{11--20}.
\newblock


\bibitem[\protect\citeauthoryear{Ding, Yuan, Zhu, Zhang, Liu, Wang, and
  Liu}{Ding et~al\mbox{.}}{2021}]%
        {ding2021transmvsnet}
\bibfield{author}{\bibinfo{person}{Yikang Ding}, \bibinfo{person}{Wentao Yuan},
  \bibinfo{person}{Qingtian Zhu}, \bibinfo{person}{Haotian Zhang},
  \bibinfo{person}{Xiangyue Liu}, \bibinfo{person}{Yuanjiang Wang}, {and}
  \bibinfo{person}{Xiao Liu}.} \bibinfo{year}{2021}\natexlab{}.
\newblock \showarticletitle{TransMVSNet: Global Context-aware Multi-view Stereo
  Network with Transformers}.
\newblock \bibinfo{journal}{\emph{arXiv preprint arXiv:2111.14600}}
  (\bibinfo{year}{2021}).
\newblock


\bibitem[\protect\citeauthoryear{Do, Vuong, Roumeliotis, and Park}{Do
  et~al\mbox{.}}{2020}]%
        {do2020surface}
\bibfield{author}{\bibinfo{person}{Tien Do}, \bibinfo{person}{Khiem Vuong},
  \bibinfo{person}{Stergios~I Roumeliotis}, {and} \bibinfo{person}{Hyun~Soo
  Park}.} \bibinfo{year}{2020}\natexlab{}.
\newblock \showarticletitle{Surface normal estimation of tilted images via
  spatial rectifier}. In \bibinfo{booktitle}{\emph{European Conference on
  Computer Vision}}. Springer, \bibinfo{pages}{265--280}.
\newblock


\bibitem[\protect\citeauthoryear{Galliani, Lasinger, and Schindler}{Galliani
  et~al\mbox{.}}{2015}]%
        {galliani2015massively}
\bibfield{author}{\bibinfo{person}{Silvano Galliani}, \bibinfo{person}{Katrin
  Lasinger}, {and} \bibinfo{person}{Konrad Schindler}.}
  \bibinfo{year}{2015}\natexlab{}.
\newblock \showarticletitle{Massively parallel multiview stereopsis by surface
  normal diffusion}. In \bibinfo{booktitle}{\emph{Proceedings of the IEEE
  International Conference on Computer Vision}}. \bibinfo{pages}{873--881}.
\newblock


\bibitem[\protect\citeauthoryear{Genova, Cole, Vlasic, Sarna, Freeman, and
  Funkhouser}{Genova et~al\mbox{.}}{2019}]%
        {genova2019learning}
\bibfield{author}{\bibinfo{person}{Kyle Genova}, \bibinfo{person}{Forrester
  Cole}, \bibinfo{person}{Daniel Vlasic}, \bibinfo{person}{Aaron Sarna},
  \bibinfo{person}{William~T Freeman}, {and} \bibinfo{person}{Thomas
  Funkhouser}.} \bibinfo{year}{2019}\natexlab{}.
\newblock \showarticletitle{Learning shape templates with structured implicit
  functions}. In \bibinfo{booktitle}{\emph{Proceedings of the IEEE/CVF
  International Conference on Computer Vision}}. \bibinfo{pages}{7154--7164}.
\newblock


\bibitem[\protect\citeauthoryear{Gu, Fan, Zhu, Dai, Tan, and Tan}{Gu
  et~al\mbox{.}}{2020}]%
        {gu2020cascade}
\bibfield{author}{\bibinfo{person}{Xiaodong Gu}, \bibinfo{person}{Zhiwen Fan},
  \bibinfo{person}{Siyu Zhu}, \bibinfo{person}{Zuozhuo Dai},
  \bibinfo{person}{Feitong Tan}, {and} \bibinfo{person}{Ping Tan}.}
  \bibinfo{year}{2020}\natexlab{}.
\newblock \showarticletitle{Cascade cost volume for high-resolution multi-view
  stereo and stereo matching}. In \bibinfo{booktitle}{\emph{Proceedings of the
  IEEE/CVF Conference on Computer Vision and Pattern Recognition}}.
  \bibinfo{pages}{2495--2504}.
\newblock


\bibitem[\protect\citeauthoryear{Guo, Peng, Lin, Wang, Zhang, Bao, and
  Zhou}{Guo et~al\mbox{.}}{2022}]%
        {manhattanindoor2022}
\bibfield{author}{\bibinfo{person}{Haoyu Guo}, \bibinfo{person}{Sida Peng},
  \bibinfo{person}{Haotong Lin}, \bibinfo{person}{Qianqian Wang},
  \bibinfo{person}{Guofeng Zhang}, \bibinfo{person}{Hujun Bao}, {and}
  \bibinfo{person}{Xiaowei Zhou}.} \bibinfo{year}{2022}\natexlab{}.
\newblock \showarticletitle{Neural 3D Scene Reconstruction with the
  Manhattan-world Assumption}. In \bibinfo{booktitle}{\emph{Proceedings of the
  IEEE/CVF Conference on Computer Vision and Pattern Recognition}}.
  \bibinfo{pages}{5511--5520}.
\newblock


\bibitem[\protect\citeauthoryear{Hou, Kannala, and Solin}{Hou
  et~al\mbox{.}}{2019}]%
        {hou2019multi}
\bibfield{author}{\bibinfo{person}{Yuxin Hou}, \bibinfo{person}{Juho Kannala},
  {and} \bibinfo{person}{Arno Solin}.} \bibinfo{year}{2019}\natexlab{}.
\newblock \showarticletitle{Multi-view stereo by temporal nonparametric
  fusion}. In \bibinfo{booktitle}{\emph{Proceedings of the IEEE/CVF
  International Conference on Computer Vision}}. \bibinfo{pages}{2651--2660}.
\newblock


\bibitem[\protect\citeauthoryear{Huang, Zhou, Funkhouser, and Guibas}{Huang
  et~al\mbox{.}}{2019}]%
        {huang2019framenet}
\bibfield{author}{\bibinfo{person}{Jingwei Huang}, \bibinfo{person}{Yichao
  Zhou}, \bibinfo{person}{Thomas Funkhouser}, {and} \bibinfo{person}{Leonidas~J
  Guibas}.} \bibinfo{year}{2019}\natexlab{}.
\newblock \showarticletitle{Framenet: Learning local canonical frames of 3d
  surfaces from a single rgb image}. In \bibinfo{booktitle}{\emph{Proceedings
  of the IEEE/CVF International Conference on Computer Vision}}.
  \bibinfo{pages}{8638--8647}.
\newblock


\bibitem[\protect\citeauthoryear{Ji, Gall, Zheng, Liu, and Fang}{Ji
  et~al\mbox{.}}{2017}]%
        {ji2017surfacenet}
\bibfield{author}{\bibinfo{person}{Mengqi Ji}, \bibinfo{person}{Juergen Gall},
  \bibinfo{person}{Haitian Zheng}, \bibinfo{person}{Yebin Liu}, {and}
  \bibinfo{person}{Lu Fang}.} \bibinfo{year}{2017}\natexlab{}.
\newblock \showarticletitle{Surfacenet: An end-to-end 3d neural network for
  multiview stereopsis}. In \bibinfo{booktitle}{\emph{Proceedings of the IEEE
  International Conference on Computer Vision}}. \bibinfo{pages}{2307--2315}.
\newblock


\bibitem[\protect\citeauthoryear{Ji, Zhang, Dai, and Fang}{Ji
  et~al\mbox{.}}{2020}]%
        {ji2020surfacenet+}
\bibfield{author}{\bibinfo{person}{Mengqi Ji}, \bibinfo{person}{Jinzhi Zhang},
  \bibinfo{person}{Qionghai Dai}, {and} \bibinfo{person}{Lu Fang}.}
  \bibinfo{year}{2020}\natexlab{}.
\newblock \showarticletitle{SurfaceNet+: An end-to-end 3D neural network for
  very sparse multi-view stereopsis}.
\newblock \bibinfo{journal}{\emph{IEEE Transactions on Pattern Analysis and
  Machine Intelligence}} \bibinfo{volume}{43}, \bibinfo{number}{11}
  (\bibinfo{year}{2020}), \bibinfo{pages}{4078--4093}.
\newblock


\bibitem[\protect\citeauthoryear{Jiang, Ding, Hu, and Huang}{Jiang
  et~al\mbox{.}}{2021}]%
        {jiang2021plnet}
\bibfield{author}{\bibinfo{person}{Hualie Jiang}, \bibinfo{person}{Laiyan
  Ding}, \bibinfo{person}{Junjie Hu}, {and} \bibinfo{person}{Rui Huang}.}
  \bibinfo{year}{2021}\natexlab{}.
\newblock \showarticletitle{PLNet: Plane and Line Priors for Unsupervised
  Indoor Depth Estimation}. In \bibinfo{booktitle}{\emph{2021 International
  Conference on 3D Vision (3DV)}}. IEEE, \bibinfo{pages}{741--750}.
\newblock


\bibitem[\protect\citeauthoryear{Kar, H{\"a}ne, and Malik}{Kar
  et~al\mbox{.}}{2017}]%
        {kar2017learning}
\bibfield{author}{\bibinfo{person}{Abhishek Kar}, \bibinfo{person}{Christian
  H{\"a}ne}, {and} \bibinfo{person}{Jitendra Malik}.}
  \bibinfo{year}{2017}\natexlab{}.
\newblock \showarticletitle{Learning a multi-view stereo machine}.
\newblock \bibinfo{journal}{\emph{Advances in neural information processing
  systems}}  \bibinfo{volume}{30} (\bibinfo{year}{2017}).
\newblock


\bibitem[\protect\citeauthoryear{Kazhdan, Bolitho, and Hoppe}{Kazhdan
  et~al\mbox{.}}{2006}]%
        {kazhdan2006poisson}
\bibfield{author}{\bibinfo{person}{Michael Kazhdan}, \bibinfo{person}{Matthew
  Bolitho}, {and} \bibinfo{person}{Hugues Hoppe}.}
  \bibinfo{year}{2006}\natexlab{}.
\newblock \showarticletitle{Poisson surface reconstruction}. In
  \bibinfo{booktitle}{\emph{Proceedings of the fourth Eurographics symposium on
  Geometry processing}}, Vol.~\bibinfo{volume}{7}.
\newblock


\bibitem[\protect\citeauthoryear{Kellnhofer, Jebe, Jones, Spicer, Pulli, and
  Wetzstein}{Kellnhofer et~al\mbox{.}}{2021}]%
        {kellnhofer2021neural}
\bibfield{author}{\bibinfo{person}{Petr Kellnhofer}, \bibinfo{person}{Lars~C
  Jebe}, \bibinfo{person}{Andrew Jones}, \bibinfo{person}{Ryan Spicer},
  \bibinfo{person}{Kari Pulli}, {and} \bibinfo{person}{Gordon Wetzstein}.}
  \bibinfo{year}{2021}\natexlab{}.
\newblock \showarticletitle{Neural lumigraph rendering}. In
  \bibinfo{booktitle}{\emph{Proceedings of the IEEE/CVF Conference on Computer
  Vision and Pattern Recognition}}. \bibinfo{pages}{4287--4297}.
\newblock


\bibitem[\protect\citeauthoryear{Kuhn, Sormann, Rossi, Erdler, and
  Fraundorfer}{Kuhn et~al\mbox{.}}{2020}]%
        {kuhn2020deepc}
\bibfield{author}{\bibinfo{person}{Andreas Kuhn}, \bibinfo{person}{Christian
  Sormann}, \bibinfo{person}{Mattia Rossi}, \bibinfo{person}{Oliver Erdler},
  {and} \bibinfo{person}{Friedrich Fraundorfer}.}
  \bibinfo{year}{2020}\natexlab{}.
\newblock \showarticletitle{Deepc-mvs: Deep confidence prediction for
  multi-view stereo reconstruction}. In \bibinfo{booktitle}{\emph{2020
  International Conference on 3D Vision (3DV)}}. IEEE,
  \bibinfo{pages}{404--413}.
\newblock


\bibitem[\protect\citeauthoryear{Liao, Fu, Yan, Luo, and Xiao}{Liao
  et~al\mbox{.}}{2021}]%
        {liao2021adaptive}
\bibfield{author}{\bibinfo{person}{Jie Liao}, \bibinfo{person}{Yanping Fu},
  \bibinfo{person}{Qingan Yan}, \bibinfo{person}{Fei Luo}, {and}
  \bibinfo{person}{Chunxia Xiao}.} \bibinfo{year}{2021}\natexlab{}.
\newblock \showarticletitle{Adaptive depth estimation for pyramid multi-view
  stereo}.
\newblock \bibinfo{journal}{\emph{Computers \& Graphics}}  \bibinfo{volume}{97}
  (\bibinfo{year}{2021}), \bibinfo{pages}{268--278}.
\newblock


\bibitem[\protect\citeauthoryear{Liu, Tang, and Shen}{Liu
  et~al\mbox{.}}{2020b}]%
        {liu2020depth}
\bibfield{author}{\bibinfo{person}{Hongmin Liu}, \bibinfo{person}{Xincheng
  Tang}, {and} \bibinfo{person}{Shuhan Shen}.}
  \bibinfo{year}{2020}\natexlab{b}.
\newblock \showarticletitle{Depth-map completion for large indoor scene
  reconstruction}.
\newblock \bibinfo{journal}{\emph{Pattern Recognition}}  \bibinfo{volume}{99}
  (\bibinfo{year}{2020}), \bibinfo{pages}{107112}.
\newblock


\bibitem[\protect\citeauthoryear{Liu, Gu, Zaw~Lin, Chua, and Theobalt}{Liu
  et~al\mbox{.}}{2020a}]%
        {liu2020neural}
\bibfield{author}{\bibinfo{person}{Lingjie Liu}, \bibinfo{person}{Jiatao Gu},
  \bibinfo{person}{Kyaw Zaw~Lin}, \bibinfo{person}{Tat-Seng Chua}, {and}
  \bibinfo{person}{Christian Theobalt}.} \bibinfo{year}{2020}\natexlab{a}.
\newblock \showarticletitle{Neural sparse voxel fields}.
\newblock \bibinfo{journal}{\emph{Advances in Neural Information Processing
  Systems}}  \bibinfo{volume}{33} (\bibinfo{year}{2020}),
  \bibinfo{pages}{15651--15663}.
\newblock


\bibitem[\protect\citeauthoryear{Long, Lin, Liu, Li, Theobalt, Yang, and
  Wang}{Long et~al\mbox{.}}{2021a}]%
        {long2021adaptive}
\bibfield{author}{\bibinfo{person}{Xiaoxiao Long}, \bibinfo{person}{Cheng Lin},
  \bibinfo{person}{Lingjie Liu}, \bibinfo{person}{Wei Li},
  \bibinfo{person}{Christian Theobalt}, \bibinfo{person}{Ruigang Yang}, {and}
  \bibinfo{person}{Wenping Wang}.} \bibinfo{year}{2021}\natexlab{a}.
\newblock \showarticletitle{Adaptive surface normal constraint for depth
  estimation}. In \bibinfo{booktitle}{\emph{Proceedings of the IEEE/CVF
  International Conference on Computer Vision}}. \bibinfo{pages}{12849--12858}.
\newblock


\bibitem[\protect\citeauthoryear{Long, Liu, Li, Theobalt, and Wang}{Long
  et~al\mbox{.}}{2021b}]%
        {long2021multi}
\bibfield{author}{\bibinfo{person}{Xiaoxiao Long}, \bibinfo{person}{Lingjie
  Liu}, \bibinfo{person}{Wei Li}, \bibinfo{person}{Christian Theobalt}, {and}
  \bibinfo{person}{Wenping Wang}.} \bibinfo{year}{2021}\natexlab{b}.
\newblock \showarticletitle{Multi-view depth estimation using epipolar
  spatio-temporal networks}. In \bibinfo{booktitle}{\emph{Proceedings of the
  IEEE/CVF Conference on Computer Vision and Pattern Recognition}}.
  \bibinfo{pages}{8258--8267}.
\newblock


\bibitem[\protect\citeauthoryear{Lorensen and Cline}{Lorensen and
  Cline}{1987}]%
        {lorensen1987marching}
\bibfield{author}{\bibinfo{person}{William~E Lorensen} {and}
  \bibinfo{person}{Harvey~E Cline}.} \bibinfo{year}{1987}\natexlab{}.
\newblock \showarticletitle{Marching cubes: A high resolution 3D surface
  construction algorithm}.
\newblock \bibinfo{journal}{\emph{ACM siggraph computer graphics}}
  \bibinfo{volume}{21}, \bibinfo{number}{4} (\bibinfo{year}{1987}),
  \bibinfo{pages}{163--169}.
\newblock


\bibitem[\protect\citeauthoryear{Merrell, Akbarzadeh, Wang, Mordohai, Frahm,
  Yang, Nist{\'e}r, and Pollefeys}{Merrell et~al\mbox{.}}{2007}]%
        {merrell2007real}
\bibfield{author}{\bibinfo{person}{Paul Merrell}, \bibinfo{person}{Amir
  Akbarzadeh}, \bibinfo{person}{Liang Wang}, \bibinfo{person}{Philippos
  Mordohai}, \bibinfo{person}{Jan-Michael Frahm}, \bibinfo{person}{Ruigang
  Yang}, \bibinfo{person}{David Nist{\'e}r}, {and} \bibinfo{person}{Marc
  Pollefeys}.} \bibinfo{year}{2007}\natexlab{}.
\newblock \showarticletitle{Real-time visibility-based fusion of depth maps}.
  In \bibinfo{booktitle}{\emph{2007 IEEE 11th International Conference on
  Computer Vision}}. IEEE, \bibinfo{pages}{1--8}.
\newblock


\bibitem[\protect\citeauthoryear{Mescheder, Oechsle, Niemeyer, Nowozin, and
  Geiger}{Mescheder et~al\mbox{.}}{2019}]%
        {mescheder2019occupancy}
\bibfield{author}{\bibinfo{person}{Lars Mescheder}, \bibinfo{person}{Michael
  Oechsle}, \bibinfo{person}{Michael Niemeyer}, \bibinfo{person}{Sebastian
  Nowozin}, {and} \bibinfo{person}{Andreas Geiger}.}
  \bibinfo{year}{2019}\natexlab{}.
\newblock \showarticletitle{Occupancy networks: Learning 3d reconstruction in
  function space}. In \bibinfo{booktitle}{\emph{Proceedings of the IEEE/CVF
  Conference on Computer Vision and Pattern Recognition}}.
  \bibinfo{pages}{4460--4470}.
\newblock


\bibitem[\protect\citeauthoryear{Michalkiewicz, Pontes, Jack, Baktashmotlagh,
  and Eriksson}{Michalkiewicz et~al\mbox{.}}{2019}]%
        {michalkiewicz2019implicit}
\bibfield{author}{\bibinfo{person}{Mateusz Michalkiewicz},
  \bibinfo{person}{Jhony~K Pontes}, \bibinfo{person}{Dominic Jack},
  \bibinfo{person}{Mahsa Baktashmotlagh}, {and} \bibinfo{person}{Anders
  Eriksson}.} \bibinfo{year}{2019}\natexlab{}.
\newblock \showarticletitle{Implicit surface representations as layers in
  neural networks}. In \bibinfo{booktitle}{\emph{Proceedings of the IEEE/CVF
  International Conference on Computer Vision}}. \bibinfo{pages}{4743--4752}.
\newblock


\bibitem[\protect\citeauthoryear{Mildenhall, Srinivasan, Tancik, Barron,
  Ramamoorthi, and Ng}{Mildenhall et~al\mbox{.}}{2020}]%
        {mildenhall2020nerf}
\bibfield{author}{\bibinfo{person}{Ben Mildenhall}, \bibinfo{person}{Pratul~P
  Srinivasan}, \bibinfo{person}{Matthew Tancik}, \bibinfo{person}{Jonathan~T
  Barron}, \bibinfo{person}{Ravi Ramamoorthi}, {and} \bibinfo{person}{Ren Ng}.}
  \bibinfo{year}{2020}\natexlab{}.
\newblock \showarticletitle{Nerf: Representing scenes as neural radiance fields
  for view synthesis}. In \bibinfo{booktitle}{\emph{European conference on
  computer vision}}. Springer, \bibinfo{pages}{405--421}.
\newblock


\bibitem[\protect\citeauthoryear{M{\"u}ller, Evans, Schied, and
  Keller}{M{\"u}ller et~al\mbox{.}}{2022}]%
        {muller2022instant}
\bibfield{author}{\bibinfo{person}{Thomas M{\"u}ller}, \bibinfo{person}{Alex
  Evans}, \bibinfo{person}{Christoph Schied}, {and} \bibinfo{person}{Alexander
  Keller}.} \bibinfo{year}{2022}\natexlab{}.
\newblock \showarticletitle{Instant Neural Graphics Primitives with a
  Multiresolution Hash Encoding}.
\newblock \bibinfo{journal}{\emph{arXiv preprint arXiv:2201.05989}}
  (\bibinfo{year}{2022}).
\newblock


\bibitem[\protect\citeauthoryear{Murez, As, Bartolozzi, Sinha, Badrinarayanan,
  and Rabinovich}{Murez et~al\mbox{.}}{2020}]%
        {murez2020atlas}
\bibfield{author}{\bibinfo{person}{Zak Murez}, \bibinfo{person}{Tarrence~van
  As}, \bibinfo{person}{James Bartolozzi}, \bibinfo{person}{Ayan Sinha},
  \bibinfo{person}{Vijay Badrinarayanan}, {and} \bibinfo{person}{Andrew
  Rabinovich}.} \bibinfo{year}{2020}\natexlab{}.
\newblock \showarticletitle{Atlas: End-to-end 3d scene reconstruction from
  posed images}. In \bibinfo{booktitle}{\emph{European Conference on Computer
  Vision}}. Springer, \bibinfo{pages}{414--431}.
\newblock


\bibitem[\protect\citeauthoryear{Newcombe, Izadi, Hilliges, Molyneaux, Kim,
  Davison, Kohi, Shotton, Hodges, and Fitzgibbon}{Newcombe
  et~al\mbox{.}}{2011}]%
        {newcombe2011kinectfusion}
\bibfield{author}{\bibinfo{person}{Richard~A Newcombe},
  \bibinfo{person}{Shahram Izadi}, \bibinfo{person}{Otmar Hilliges},
  \bibinfo{person}{David Molyneaux}, \bibinfo{person}{David Kim},
  \bibinfo{person}{Andrew~J Davison}, \bibinfo{person}{Pushmeet Kohi},
  \bibinfo{person}{Jamie Shotton}, \bibinfo{person}{Steve Hodges}, {and}
  \bibinfo{person}{Andrew Fitzgibbon}.} \bibinfo{year}{2011}\natexlab{}.
\newblock \showarticletitle{Kinectfusion: Real-time dense surface mapping and
  tracking}. In \bibinfo{booktitle}{\emph{2011 10th IEEE international
  symposium on mixed and augmented reality}}. IEEE, \bibinfo{pages}{127--136}.
\newblock


\bibitem[\protect\citeauthoryear{Niemeyer, Barron, Mildenhall, Sajjadi, Geiger,
  and Radwan}{Niemeyer et~al\mbox{.}}{2021}]%
        {niemeyer2021regnerf}
\bibfield{author}{\bibinfo{person}{Michael Niemeyer},
  \bibinfo{person}{Jonathan~T Barron}, \bibinfo{person}{Ben Mildenhall},
  \bibinfo{person}{Mehdi~SM Sajjadi}, \bibinfo{person}{Andreas Geiger}, {and}
  \bibinfo{person}{Noha Radwan}.} \bibinfo{year}{2021}\natexlab{}.
\newblock \showarticletitle{RegNeRF: Regularizing Neural Radiance Fields for
  View Synthesis from Sparse Inputs}.
\newblock \bibinfo{journal}{\emph{arXiv preprint arXiv:2112.00724}}
  (\bibinfo{year}{2021}).
\newblock


\bibitem[\protect\citeauthoryear{Niemeyer, Mescheder, Oechsle, and
  Geiger}{Niemeyer et~al\mbox{.}}{2020}]%
        {niemeyer2020differentiable}
\bibfield{author}{\bibinfo{person}{Michael Niemeyer}, \bibinfo{person}{Lars
  Mescheder}, \bibinfo{person}{Michael Oechsle}, {and} \bibinfo{person}{Andreas
  Geiger}.} \bibinfo{year}{2020}\natexlab{}.
\newblock \showarticletitle{Differentiable volumetric rendering: Learning
  implicit 3d representations without 3d supervision}. In
  \bibinfo{booktitle}{\emph{Proceedings of the IEEE/CVF Conference on Computer
  Vision and Pattern Recognition}}. \bibinfo{pages}{3504--3515}.
\newblock


\bibitem[\protect\citeauthoryear{Oechsle, Peng, and Geiger}{Oechsle
  et~al\mbox{.}}{2021}]%
        {oechsle2021unisurf}
\bibfield{author}{\bibinfo{person}{Michael Oechsle}, \bibinfo{person}{Songyou
  Peng}, {and} \bibinfo{person}{Andreas Geiger}.}
  \bibinfo{year}{2021}\natexlab{}.
\newblock \showarticletitle{Unisurf: Unifying neural implicit surfaces and
  radiance fields for multi-view reconstruction}. In
  \bibinfo{booktitle}{\emph{Proceedings of the IEEE/CVF International
  Conference on Computer Vision}}. \bibinfo{pages}{5589--5599}.
\newblock


\bibitem[\protect\citeauthoryear{Park, Florence, Straub, Newcombe, and
  Lovegrove}{Park et~al\mbox{.}}{2019}]%
        {park2019deepsdf}
\bibfield{author}{\bibinfo{person}{Jeong~Joon Park}, \bibinfo{person}{Peter
  Florence}, \bibinfo{person}{Julian Straub}, \bibinfo{person}{Richard
  Newcombe}, {and} \bibinfo{person}{Steven Lovegrove}.}
  \bibinfo{year}{2019}\natexlab{}.
\newblock \showarticletitle{Deepsdf: Learning continuous signed distance
  functions for shape representation}. In \bibinfo{booktitle}{\emph{Proceedings
  of the IEEE/CVF Conference on Computer Vision and Pattern Recognition}}.
  \bibinfo{pages}{165--174}.
\newblock


\bibitem[\protect\citeauthoryear{Paszke, Gross, Massa, Lerer, and
  Chintala}{Paszke et~al\mbox{.}}{2019}]%
        {2019PyTorch}
\bibfield{author}{\bibinfo{person}{A. Paszke}, \bibinfo{person}{S. Gross},
  \bibinfo{person}{F. Massa}, \bibinfo{person}{A. Lerer}, {and}
  \bibinfo{person}{S. Chintala}.} \bibinfo{year}{2019}\natexlab{}.
\newblock \showarticletitle{PyTorch: An Imperative Style, High-Performance Deep
  Learning Library}.
\newblock


\bibitem[\protect\citeauthoryear{Peng, Niemeyer, Mescheder, Pollefeys, and
  Geiger}{Peng et~al\mbox{.}}{2020}]%
        {peng2020convolutional}
\bibfield{author}{\bibinfo{person}{Songyou Peng}, \bibinfo{person}{Michael
  Niemeyer}, \bibinfo{person}{Lars Mescheder}, \bibinfo{person}{Marc
  Pollefeys}, {and} \bibinfo{person}{Andreas Geiger}.}
  \bibinfo{year}{2020}\natexlab{}.
\newblock \showarticletitle{Convolutional occupancy networks}. In
  \bibinfo{booktitle}{\emph{European Conference on Computer Vision}}. Springer,
  \bibinfo{pages}{523--540}.
\newblock


\bibitem[\protect\citeauthoryear{Philip, Morgenthaler, Gharbi, and
  Drettakis}{Philip et~al\mbox{.}}{2021}]%
        {philip2021free}
\bibfield{author}{\bibinfo{person}{Julien Philip},
  \bibinfo{person}{S{\'e}bastien Morgenthaler}, \bibinfo{person}{Micha{\"e}l
  Gharbi}, {and} \bibinfo{person}{George Drettakis}.}
  \bibinfo{year}{2021}\natexlab{}.
\newblock \showarticletitle{Free-viewpoint indoor neural relighting from
  multi-view stereo}.
\newblock \bibinfo{journal}{\emph{ACM Transactions on Graphics (TOG)}}
  \bibinfo{volume}{40}, \bibinfo{number}{5} (\bibinfo{year}{2021}),
  \bibinfo{pages}{1--18}.
\newblock


\bibitem[\protect\citeauthoryear{Rich, Stier, Sen, and H{\"o}llerer}{Rich
  et~al\mbox{.}}{2021}]%
        {rich20213dvnet}
\bibfield{author}{\bibinfo{person}{Alexander Rich}, \bibinfo{person}{Noah
  Stier}, \bibinfo{person}{Pradeep Sen}, {and} \bibinfo{person}{Tobias
  H{\"o}llerer}.} \bibinfo{year}{2021}\natexlab{}.
\newblock \showarticletitle{3DVNet: Multi-View Depth Prediction and Volumetric
  Refinement}. In \bibinfo{booktitle}{\emph{2021 International Conference on 3D
  Vision (3DV)}}. IEEE, \bibinfo{pages}{700--709}.
\newblock


\bibitem[\protect\citeauthoryear{Roessle, Barron, Mildenhall, Srinivasan, and
  Nie{\ss}ner}{Roessle et~al\mbox{.}}{2021}]%
        {roessle2021dense}
\bibfield{author}{\bibinfo{person}{Barbara Roessle},
  \bibinfo{person}{Jonathan~T Barron}, \bibinfo{person}{Ben Mildenhall},
  \bibinfo{person}{Pratul~P Srinivasan}, {and} \bibinfo{person}{Matthias
  Nie{\ss}ner}.} \bibinfo{year}{2021}\natexlab{}.
\newblock \showarticletitle{Dense Depth Priors for Neural Radiance Fields from
  Sparse Input Views}.
\newblock \bibinfo{journal}{\emph{arXiv preprint arXiv:2112.03288}}
  (\bibinfo{year}{2021}).
\newblock


\bibitem[\protect\citeauthoryear{Saito, Huang, Natsume, Morishima, Kanazawa,
  and Li}{Saito et~al\mbox{.}}{2019}]%
        {saito2019pifu}
\bibfield{author}{\bibinfo{person}{Shunsuke Saito}, \bibinfo{person}{Zeng
  Huang}, \bibinfo{person}{Ryota Natsume}, \bibinfo{person}{Shigeo Morishima},
  \bibinfo{person}{Angjoo Kanazawa}, {and} \bibinfo{person}{Hao Li}.}
  \bibinfo{year}{2019}\natexlab{}.
\newblock \showarticletitle{Pifu: Pixel-aligned implicit function for
  high-resolution clothed human digitization}. In
  \bibinfo{booktitle}{\emph{Proceedings of the IEEE/CVF International
  Conference on Computer Vision}}. \bibinfo{pages}{2304--2314}.
\newblock


\bibitem[\protect\citeauthoryear{Saito, Simon, Saragih, and Joo}{Saito
  et~al\mbox{.}}{2020}]%
        {saito2020pifuhd}
\bibfield{author}{\bibinfo{person}{Shunsuke Saito}, \bibinfo{person}{Tomas
  Simon}, \bibinfo{person}{Jason Saragih}, {and} \bibinfo{person}{Hanbyul
  Joo}.} \bibinfo{year}{2020}\natexlab{}.
\newblock \showarticletitle{Pifuhd: Multi-level pixel-aligned implicit function
  for high-resolution 3d human digitization}. In
  \bibinfo{booktitle}{\emph{Proceedings of the IEEE/CVF Conference on Computer
  Vision and Pattern Recognition}}. \bibinfo{pages}{84--93}.
\newblock


\bibitem[\protect\citeauthoryear{Schonberger and Frahm}{Schonberger and
  Frahm}{2016}]%
        {schonberger2016structure}
\bibfield{author}{\bibinfo{person}{Johannes~L Schonberger} {and}
  \bibinfo{person}{Jan-Michael Frahm}.} \bibinfo{year}{2016}\natexlab{}.
\newblock \showarticletitle{Structure-from-motion revisited}. In
  \bibinfo{booktitle}{\emph{Proceedings of the IEEE conference on computer
  vision and pattern recognition}}. \bibinfo{pages}{4104--4113}.
\newblock


\bibitem[\protect\citeauthoryear{Seitz, Curless, Diebel, Scharstein, and
  Szeliski}{Seitz et~al\mbox{.}}{2006}]%
        {seitz2006comparison}
\bibfield{author}{\bibinfo{person}{Steven~M Seitz}, \bibinfo{person}{Brian
  Curless}, \bibinfo{person}{James Diebel}, \bibinfo{person}{Daniel
  Scharstein}, {and} \bibinfo{person}{Richard Szeliski}.}
  \bibinfo{year}{2006}\natexlab{}.
\newblock \showarticletitle{A comparison and evaluation of multi-view stereo
  reconstruction algorithms}. In \bibinfo{booktitle}{\emph{2006 IEEE computer
  society conference on computer vision and pattern recognition (CVPR'06)}},
  Vol.~\bibinfo{volume}{1}. IEEE, \bibinfo{pages}{519--528}.
\newblock


\bibitem[\protect\citeauthoryear{Sun, Xie, Chen, Zhou, and Bao}{Sun
  et~al\mbox{.}}{2021a}]%
        {sun2021neuralrecon}
\bibfield{author}{\bibinfo{person}{Jiaming Sun}, \bibinfo{person}{Yiming Xie},
  \bibinfo{person}{Linghao Chen}, \bibinfo{person}{Xiaowei Zhou}, {and}
  \bibinfo{person}{Hujun Bao}.} \bibinfo{year}{2021}\natexlab{a}.
\newblock \showarticletitle{NeuralRecon: Real-time coherent 3D reconstruction
  from monocular video}. In \bibinfo{booktitle}{\emph{Proceedings of the
  IEEE/CVF Conference on Computer Vision and Pattern Recognition}}.
  \bibinfo{pages}{15598--15607}.
\newblock


\bibitem[\protect\citeauthoryear{Sun, Zheng, Shi, Xu, and Liu}{Sun
  et~al\mbox{.}}{2021b}]%
        {sun2021phi}
\bibfield{author}{\bibinfo{person}{Shang Sun}, \bibinfo{person}{Yunan Zheng},
  \bibinfo{person}{Xuelei Shi}, \bibinfo{person}{Zhenyu Xu}, {and}
  \bibinfo{person}{Yiguang Liu}.} \bibinfo{year}{2021}\natexlab{b}.
\newblock \showarticletitle{PHI-MVS: Plane Hypothesis Inference Multi-view
  Stereo for Large-Scale Scene Reconstruction}.
\newblock \bibinfo{journal}{\emph{arXiv preprint arXiv:2104.06165}}
  (\bibinfo{year}{2021}).
\newblock


\bibitem[\protect\citeauthoryear{Tancik, Casser, Yan, Pradhan, Mildenhall,
  Srinivasan, Barron, and Kretzschmar}{Tancik et~al\mbox{.}}{2022}]%
        {tancik2022block}
\bibfield{author}{\bibinfo{person}{Matthew Tancik}, \bibinfo{person}{Vincent
  Casser}, \bibinfo{person}{Xinchen Yan}, \bibinfo{person}{Sabeek Pradhan},
  \bibinfo{person}{Ben Mildenhall}, \bibinfo{person}{Pratul~P Srinivasan},
  \bibinfo{person}{Jonathan~T Barron}, {and} \bibinfo{person}{Henrik
  Kretzschmar}.} \bibinfo{year}{2022}\natexlab{}.
\newblock \showarticletitle{Block-NeRF: Scalable Large Scene Neural View
  Synthesis}.
\newblock \bibinfo{journal}{\emph{arXiv preprint arXiv:2202.05263}}
  (\bibinfo{year}{2022}).
\newblock


\bibitem[\protect\citeauthoryear{Teed and Deng}{Teed and Deng}{2018}]%
        {teed2018deepv2d}
\bibfield{author}{\bibinfo{person}{Zachary Teed} {and} \bibinfo{person}{Jia
  Deng}.} \bibinfo{year}{2018}\natexlab{}.
\newblock \showarticletitle{Deepv2d: Video to depth with differentiable
  structure from motion}.
\newblock \bibinfo{journal}{\emph{arXiv preprint arXiv:1812.04605}}
  (\bibinfo{year}{2018}).
\newblock


\bibitem[\protect\citeauthoryear{Valentin, Dai, Nie{\ss}ner, Kohli, Torr,
  Izadi, and Keskin}{Valentin et~al\mbox{.}}{2016}]%
        {valentin2016learning}
\bibfield{author}{\bibinfo{person}{Julien Valentin}, \bibinfo{person}{Angela
  Dai}, \bibinfo{person}{Matthias Nie{\ss}ner}, \bibinfo{person}{Pushmeet
  Kohli}, \bibinfo{person}{Philip Torr}, \bibinfo{person}{Shahram Izadi}, {and}
  \bibinfo{person}{Cem Keskin}.} \bibinfo{year}{2016}\natexlab{}.
\newblock \showarticletitle{Learning to Navigate the Energy Landscape}.
\newblock \bibinfo{journal}{\emph{arXiv preprint arXiv:1603.05772}}
  (\bibinfo{year}{2016}).
\newblock


\bibitem[\protect\citeauthoryear{Wang, Galliani, Vogel, Speciale, and
  Pollefeys}{Wang et~al\mbox{.}}{2021a}]%
        {wang2021patchmatchnet}
\bibfield{author}{\bibinfo{person}{Fangjinhua Wang}, \bibinfo{person}{Silvano
  Galliani}, \bibinfo{person}{Christoph Vogel}, \bibinfo{person}{Pablo
  Speciale}, {and} \bibinfo{person}{Marc Pollefeys}.}
  \bibinfo{year}{2021}\natexlab{a}.
\newblock \showarticletitle{Patchmatchnet: Learned multi-view patchmatch
  stereo}. In \bibinfo{booktitle}{\emph{Proceedings of the IEEE/CVF Conference
  on Computer Vision and Pattern Recognition}}. \bibinfo{pages}{14194--14203}.
\newblock


\bibitem[\protect\citeauthoryear{Wang, Wang, Long, Theobalt, Komura, Liu, and
  Wang}{Wang et~al\mbox{.}}{2022}]%
        {wang2022neuris}
\bibfield{author}{\bibinfo{person}{Jiepeng Wang}, \bibinfo{person}{Peng Wang},
  \bibinfo{person}{Xiaoxiao Long}, \bibinfo{person}{Christian Theobalt},
  \bibinfo{person}{Taku Komura}, \bibinfo{person}{Lingjie Liu}, {and}
  \bibinfo{person}{Wenping Wang}.} \bibinfo{year}{2022}\natexlab{}.
\newblock \showarticletitle{NeuRIS: Neural Reconstruction of Indoor Scenes
  Using Normal Priors}.
\newblock \bibinfo{journal}{\emph{arXiv preprint arXiv:2206.13597}}
  (\bibinfo{year}{2022}).
\newblock


\bibitem[\protect\citeauthoryear{Wang and Shen}{Wang and Shen}{2018}]%
        {wang2018mvdepthnet}
\bibfield{author}{\bibinfo{person}{Kaixuan Wang} {and} \bibinfo{person}{Shaojie
  Shen}.} \bibinfo{year}{2018}\natexlab{}.
\newblock \showarticletitle{Mvdepthnet: Real-time multiview depth estimation
  neural network}. In \bibinfo{booktitle}{\emph{2018 International conference
  on 3d vision (3DV)}}. IEEE, \bibinfo{pages}{248--257}.
\newblock


\bibitem[\protect\citeauthoryear{Wang, Liu, Liu, Theobalt, Komura, and
  Wang}{Wang et~al\mbox{.}}{2021b}]%
        {wang2021neus}
\bibfield{author}{\bibinfo{person}{Peng Wang}, \bibinfo{person}{Lingjie Liu},
  \bibinfo{person}{Yuan Liu}, \bibinfo{person}{Christian Theobalt},
  \bibinfo{person}{Taku Komura}, {and} \bibinfo{person}{Wenping Wang}.}
  \bibinfo{year}{2021}\natexlab{b}.
\newblock \showarticletitle{Neus: Learning neural implicit surfaces by volume
  rendering for multi-view reconstruction}.
\newblock \bibinfo{journal}{\emph{arXiv preprint arXiv:2106.10689}}
  (\bibinfo{year}{2021}).
\newblock


\bibitem[\protect\citeauthoryear{Wang, Geraghty, Matzen, Szeliski, and
  Frahm}{Wang et~al\mbox{.}}{2020}]%
        {wang2020vplnet}
\bibfield{author}{\bibinfo{person}{Rui Wang}, \bibinfo{person}{David Geraghty},
  \bibinfo{person}{Kevin Matzen}, \bibinfo{person}{Richard Szeliski}, {and}
  \bibinfo{person}{Jan-Michael Frahm}.} \bibinfo{year}{2020}\natexlab{}.
\newblock \showarticletitle{Vplnet: Deep single view normal estimation with
  vanishing points and lines}. In \bibinfo{booktitle}{\emph{Proceedings of the
  IEEE/CVF Conference on Computer Vision and Pattern Recognition}}.
  \bibinfo{pages}{689--698}.
\newblock


\bibitem[\protect\citeauthoryear{Wei, Liu, Rao, Zhao, Lu, and Zhou}{Wei
  et~al\mbox{.}}{2021a}]%
        {wei2021nerfingmvs}
\bibfield{author}{\bibinfo{person}{Yi Wei}, \bibinfo{person}{Shaohui Liu},
  \bibinfo{person}{Yongming Rao}, \bibinfo{person}{Wang Zhao},
  \bibinfo{person}{Jiwen Lu}, {and} \bibinfo{person}{Jie Zhou}.}
  \bibinfo{year}{2021}\natexlab{a}.
\newblock \showarticletitle{Nerfingmvs: Guided optimization of neural radiance
  fields for indoor multi-view stereo}. In
  \bibinfo{booktitle}{\emph{Proceedings of the IEEE/CVF International
  Conference on Computer Vision}}. \bibinfo{pages}{5610--5619}.
\newblock


\bibitem[\protect\citeauthoryear{Wei, Zhu, Min, Chen, and Wang}{Wei
  et~al\mbox{.}}{2021b}]%
        {wei2021aa}
\bibfield{author}{\bibinfo{person}{Zizhuang Wei}, \bibinfo{person}{Qingtian
  Zhu}, \bibinfo{person}{Chen Min}, \bibinfo{person}{Yisong Chen}, {and}
  \bibinfo{person}{Guoping Wang}.} \bibinfo{year}{2021}\natexlab{b}.
\newblock \showarticletitle{Aa-rmvsnet: Adaptive aggregation recurrent
  multi-view stereo network}. In \bibinfo{booktitle}{\emph{Proceedings of the
  IEEE/CVF International Conference on Computer Vision}}.
  \bibinfo{pages}{6187--6196}.
\newblock


\bibitem[\protect\citeauthoryear{Xiangli, Xu, Pan, Zhao, Rao, Theobalt, Dai,
  and Lin}{Xiangli et~al\mbox{.}}{2021}]%
        {xiangli2021citynerf}
\bibfield{author}{\bibinfo{person}{Yuanbo Xiangli}, \bibinfo{person}{Linning
  Xu}, \bibinfo{person}{Xingang Pan}, \bibinfo{person}{Nanxuan Zhao},
  \bibinfo{person}{Anyi Rao}, \bibinfo{person}{Christian Theobalt},
  \bibinfo{person}{Bo Dai}, {and} \bibinfo{person}{Dahua Lin}.}
  \bibinfo{year}{2021}\natexlab{}.
\newblock \showarticletitle{CityNeRF: Building NeRF at City Scale}.
\newblock \bibinfo{journal}{\emph{arXiv preprint arXiv:2112.05504}}
  (\bibinfo{year}{2021}).
\newblock


\bibitem[\protect\citeauthoryear{Xu, Zhou, Qiao, Kang, and Wu}{Xu
  et~al\mbox{.}}{2021}]%
        {xu2021self}
\bibfield{author}{\bibinfo{person}{Hongbin Xu}, \bibinfo{person}{Zhipeng Zhou},
  \bibinfo{person}{Yu Qiao}, \bibinfo{person}{Wenxiong Kang}, {and}
  \bibinfo{person}{Qiuxia Wu}.} \bibinfo{year}{2021}\natexlab{}.
\newblock \showarticletitle{Self-supervised multi-view stereo via effective
  co-segmentation and data-augmentation}. In
  \bibinfo{booktitle}{\emph{Proceedings of the AAAI Conference on Artificial
  Intelligence}}, Vol.~\bibinfo{volume}{2}. \bibinfo{pages}{6}.
\newblock


\bibitem[\protect\citeauthoryear{Xu, Zhu, Bao, and Xu}{Xu
  et~al\mbox{.}}{2022b}]%
        {xu2022hybrid}
\bibfield{author}{\bibinfo{person}{Jiamin Xu}, \bibinfo{person}{Zihan Zhu},
  \bibinfo{person}{Hujun Bao}, {and} \bibinfo{person}{Wewei Xu}.}
  \bibinfo{year}{2022}\natexlab{b}.
\newblock \showarticletitle{A Hybrid Mesh-neural Representation for 3D
  Transparent Object Reconstruction}.
\newblock \bibinfo{journal}{\emph{arXiv preprint arXiv:2203.12613}}
  (\bibinfo{year}{2022}).
\newblock


\bibitem[\protect\citeauthoryear{Xu and Tao}{Xu and Tao}{2019}]%
        {xu2019multi}
\bibfield{author}{\bibinfo{person}{Qingshan Xu} {and} \bibinfo{person}{Wenbing
  Tao}.} \bibinfo{year}{2019}\natexlab{}.
\newblock \showarticletitle{Multi-scale geometric consistency guided multi-view
  stereo}. In \bibinfo{booktitle}{\emph{Proceedings of the IEEE/CVF Conference
  on Computer Vision and Pattern Recognition}}. \bibinfo{pages}{5483--5492}.
\newblock


\bibitem[\protect\citeauthoryear{Xu and Tao}{Xu and Tao}{2020a}]%
        {xu2020planar}
\bibfield{author}{\bibinfo{person}{Qingshan Xu} {and} \bibinfo{person}{Wenbing
  Tao}.} \bibinfo{year}{2020}\natexlab{a}.
\newblock \showarticletitle{Planar prior assisted patchmatch multi-view
  stereo}. In \bibinfo{booktitle}{\emph{Proceedings of the AAAI Conference on
  Artificial Intelligence}}, Vol.~\bibinfo{volume}{34}.
  \bibinfo{pages}{12516--12523}.
\newblock


\bibitem[\protect\citeauthoryear{Xu and Tao}{Xu and Tao}{2020b}]%
        {Xu2020ACMP}
\bibfield{author}{\bibinfo{person}{Qingshan Xu} {and} \bibinfo{person}{Wenbing
  Tao}.} \bibinfo{year}{2020}\natexlab{b}.
\newblock \showarticletitle{Planar Prior Assisted PatchMatch Multi-View
  Stereo}.
\newblock \bibinfo{journal}{\emph{AAAI Conference on Artificial Intelligence
  (AAAI)}} (\bibinfo{year}{2020}).
\newblock


\bibitem[\protect\citeauthoryear{Xu, Xu, Philip, Bi, Shu, Sunkavalli, and
  Neumann}{Xu et~al\mbox{.}}{2022a}]%
        {xu2022point}
\bibfield{author}{\bibinfo{person}{Qiangeng Xu}, \bibinfo{person}{Zexiang Xu},
  \bibinfo{person}{Julien Philip}, \bibinfo{person}{Sai Bi},
  \bibinfo{person}{Zhixin Shu}, \bibinfo{person}{Kalyan Sunkavalli}, {and}
  \bibinfo{person}{Ulrich Neumann}.} \bibinfo{year}{2022}\natexlab{a}.
\newblock \showarticletitle{Point-NeRF: Point-based Neural Radiance Fields}.
\newblock \bibinfo{journal}{\emph{arXiv preprint arXiv:2201.08845}}
  (\bibinfo{year}{2022}).
\newblock


\bibitem[\protect\citeauthoryear{Yan, Wei, Yi, Ding, Zhang, Chen, Wang, and
  Tai}{Yan et~al\mbox{.}}{2020}]%
        {yan2020dense}
\bibfield{author}{\bibinfo{person}{Jianfeng Yan}, \bibinfo{person}{Zizhuang
  Wei}, \bibinfo{person}{Hongwei Yi}, \bibinfo{person}{Mingyu Ding},
  \bibinfo{person}{Runze Zhang}, \bibinfo{person}{Yisong Chen},
  \bibinfo{person}{Guoping Wang}, {and} \bibinfo{person}{Yu-Wing Tai}.}
  \bibinfo{year}{2020}\natexlab{}.
\newblock \showarticletitle{Dense hybrid recurrent multi-view stereo net with
  dynamic consistency checking}. In \bibinfo{booktitle}{\emph{European
  Conference on Computer Vision}}. Springer, \bibinfo{pages}{674--689}.
\newblock


\bibitem[\protect\citeauthoryear{Yang, Mao, Alvarez, and Liu}{Yang
  et~al\mbox{.}}{2020}]%
        {yang2020cost}
\bibfield{author}{\bibinfo{person}{Jiayu Yang}, \bibinfo{person}{Wei Mao},
  \bibinfo{person}{Jose~M Alvarez}, {and} \bibinfo{person}{Miaomiao Liu}.}
  \bibinfo{year}{2020}\natexlab{}.
\newblock \showarticletitle{Cost volume pyramid based depth inference for
  multi-view stereo}. In \bibinfo{booktitle}{\emph{Proceedings of the IEEE/CVF
  Conference on Computer Vision and Pattern Recognition}}.
  \bibinfo{pages}{4877--4886}.
\newblock


\bibitem[\protect\citeauthoryear{Yao, Luo, Li, Fang, and Quan}{Yao
  et~al\mbox{.}}{2018}]%
        {yao2018mvsnet}
\bibfield{author}{\bibinfo{person}{Yao Yao}, \bibinfo{person}{Zixin Luo},
  \bibinfo{person}{Shiwei Li}, \bibinfo{person}{Tian Fang}, {and}
  \bibinfo{person}{Long Quan}.} \bibinfo{year}{2018}\natexlab{}.
\newblock \showarticletitle{Mvsnet: Depth inference for unstructured multi-view
  stereo}. In \bibinfo{booktitle}{\emph{Proceedings of the European Conference
  on Computer Vision (ECCV)}}. \bibinfo{pages}{767--783}.
\newblock


\bibitem[\protect\citeauthoryear{Yao, Luo, Li, Shen, Fang, and Quan}{Yao
  et~al\mbox{.}}{2019}]%
        {yao2019recurrent}
\bibfield{author}{\bibinfo{person}{Yao Yao}, \bibinfo{person}{Zixin Luo},
  \bibinfo{person}{Shiwei Li}, \bibinfo{person}{Tianwei Shen},
  \bibinfo{person}{Tian Fang}, {and} \bibinfo{person}{Long Quan}.}
  \bibinfo{year}{2019}\natexlab{}.
\newblock \showarticletitle{Recurrent mvsnet for high-resolution multi-view
  stereo depth inference}. In \bibinfo{booktitle}{\emph{Proceedings of the
  IEEE/CVF Conference on Computer Vision and Pattern Recognition}}.
  \bibinfo{pages}{5525--5534}.
\newblock


\bibitem[\protect\citeauthoryear{Yariv, Gu, Kasten, and Lipman}{Yariv
  et~al\mbox{.}}{2021}]%
        {volsdf2021}
\bibfield{author}{\bibinfo{person}{Lior Yariv}, \bibinfo{person}{Jiatao Gu},
  \bibinfo{person}{Yoni Kasten}, {and} \bibinfo{person}{Yaron Lipman}.}
  \bibinfo{year}{2021}\natexlab{}.
\newblock \showarticletitle{Volume rendering of neural implicit surfaces}.
\newblock \bibinfo{journal}{\emph{Advances in Neural Information Processing
  Systems}}  \bibinfo{volume}{34} (\bibinfo{year}{2021}),
  \bibinfo{pages}{4805--4815}.
\newblock


\bibitem[\protect\citeauthoryear{Yariv, Kasten, Moran, Galun, Atzmon, Ronen,
  and Lipman}{Yariv et~al\mbox{.}}{2020}]%
        {yariv2020multiview}
\bibfield{author}{\bibinfo{person}{Lior Yariv}, \bibinfo{person}{Yoni Kasten},
  \bibinfo{person}{Dror Moran}, \bibinfo{person}{Meirav Galun},
  \bibinfo{person}{Matan Atzmon}, \bibinfo{person}{Basri Ronen}, {and}
  \bibinfo{person}{Yaron Lipman}.} \bibinfo{year}{2020}\natexlab{}.
\newblock \showarticletitle{Multiview neural surface reconstruction by
  disentangling geometry and appearance}.
\newblock \bibinfo{journal}{\emph{Advances in Neural Information Processing
  Systems}}  \bibinfo{volume}{33} (\bibinfo{year}{2020}),
  \bibinfo{pages}{2492--2502}.
\newblock


\bibitem[\protect\citeauthoryear{Yu, Fridovich-Keil, Tancik, Chen, Recht, and
  Kanazawa}{Yu et~al\mbox{.}}{2021}]%
        {yu2021plenoxels}
\bibfield{author}{\bibinfo{person}{Alex Yu}, \bibinfo{person}{Sara
  Fridovich-Keil}, \bibinfo{person}{Matthew Tancik}, \bibinfo{person}{Qinhong
  Chen}, \bibinfo{person}{Benjamin Recht}, {and} \bibinfo{person}{Angjoo
  Kanazawa}.} \bibinfo{year}{2021}\natexlab{}.
\newblock \showarticletitle{Plenoxels: Radiance Fields without Neural
  Networks}.
\newblock \bibinfo{journal}{\emph{arXiv preprint arXiv:2112.05131}}
  (\bibinfo{year}{2021}).
\newblock


\bibitem[\protect\citeauthoryear{Yu, Peng, Niemeyer, Sattler, and Geiger}{Yu
  et~al\mbox{.}}{2022}]%
        {monosdf}
\bibfield{author}{\bibinfo{person}{Zehao Yu}, \bibinfo{person}{Songyou Peng},
  \bibinfo{person}{Michael Niemeyer}, \bibinfo{person}{Torsten Sattler}, {and}
  \bibinfo{person}{Andreas Geiger}.} \bibinfo{year}{2022}\natexlab{}.
\newblock \showarticletitle{MonoSDF: Exploring Monocular Geometric Cues for
  Neural Implicit Surface Reconstruction}.
\newblock \bibinfo{journal}{\emph{arXiv preprint arXiv:2206.00665}}
  (\bibinfo{year}{2022}).
\newblock


\bibitem[\protect\citeauthoryear{Zhang, Yao, Li, Luo, and Fang}{Zhang
  et~al\mbox{.}}{2020b}]%
        {zhang2020visibility}
\bibfield{author}{\bibinfo{person}{Jingyang Zhang}, \bibinfo{person}{Yao Yao},
  \bibinfo{person}{Shiwei Li}, \bibinfo{person}{Zixin Luo}, {and}
  \bibinfo{person}{Tian Fang}.} \bibinfo{year}{2020}\natexlab{b}.
\newblock \showarticletitle{Visibility-aware multi-view stereo network}.
\newblock \bibinfo{journal}{\emph{arXiv preprint arXiv:2008.07928}}
  (\bibinfo{year}{2020}).
\newblock


\bibitem[\protect\citeauthoryear{Zhang, Riegler, Snavely, and Koltun}{Zhang
  et~al\mbox{.}}{2020a}]%
        {zhang2020nerf++}
\bibfield{author}{\bibinfo{person}{Kai Zhang}, \bibinfo{person}{Gernot
  Riegler}, \bibinfo{person}{Noah Snavely}, {and} \bibinfo{person}{Vladlen
  Koltun}.} \bibinfo{year}{2020}\natexlab{a}.
\newblock \showarticletitle{Nerf++: Analyzing and improving neural radiance
  fields}.
\newblock \bibinfo{journal}{\emph{arXiv preprint arXiv:2010.07492}}
  (\bibinfo{year}{2020}).
\newblock


\bibitem[\protect\citeauthoryear{Zhang, Dong, Liu, Yan, Xiao,
  et~al\mbox{.}}{Zhang et~al\mbox{.}}{2022}]%
        {zhang2022point}
\bibfield{author}{\bibinfo{person}{Wenxiao Zhang}, \bibinfo{person}{Zhen Dong},
  \bibinfo{person}{Jun Liu}, \bibinfo{person}{Qingan Yan},
  \bibinfo{person}{Chunxia Xiao}, {et~al\mbox{.}}}
  \bibinfo{year}{2022}\natexlab{}.
\newblock \showarticletitle{Point Cloud Completion Via Skeleton-Detail
  Transformer}.
\newblock \bibinfo{journal}{\emph{IEEE Transactions on Visualization and
  Computer Graphics}} (\bibinfo{year}{2022}).
\newblock


\bibitem[\protect\citeauthoryear{Zhou, Park, and Koltun}{Zhou
  et~al\mbox{.}}{2018}]%
        {Zhou2018open3d}
\bibfield{author}{\bibinfo{person}{Qian-Yi Zhou}, \bibinfo{person}{Jaesik
  Park}, {and} \bibinfo{person}{Vladlen Koltun}.}
  \bibinfo{year}{2018}\natexlab{}.
\newblock \showarticletitle{{Open3D}: {A} Modern Library for {3D} Data
  Processing}.
\newblock \bibinfo{journal}{\emph{arXiv:1801.09847}} (\bibinfo{year}{2018}).
\newblock


\end{thebibliography}

\end{document}